%% file: main.tex
\documentclass[acmlarge,screen]{acmart}

\usepackage{hyperref}
\usepackage{subfig}
\usepackage{listings}
\usepackage{amsmath}
\usepackage{color}
\usepackage{amsmath,amssymb}
\usepackage{wrapfig}
\usepackage{soul}

\DeclareMathOperator*{\argmax}{\arg\!\max}
\DeclareMathOperator*{\argmin}{\arg\!\min}

\definecolor{dkgreen}{rgb}{0,0.6,0}
\definecolor{gray}{rgb}{0.5,0.5,0.5}
\definecolor{mauve}{rgb}{0.58,0,0.82}
\definecolor{darkmagenta}{rgb}{0.55, 0.0, 0.55}
\definecolor{darkpastelpurple}{rgb}{0.59, 0.44, 0.84}
\definecolor{ao(english)}{rgb}{0.0, 0.5, 0.0}
\definecolor{darkviolet}{rgb}{0.45, 0.0, 0.93}
\definecolor{dartmouthgreen}{rgb}{0.05, 0.55, 0.06}

\def\BibTeX{{\rm B\kern-.05em{\sc i\kern-.025em b}\kern-.08emT\kern-.1667em\lower.7ex\hbox{E}\kern-.125emX}}
    
\copyrightyear{2019}
\acmYear{2019}
\setcopyright{acmlicensed}
\acmConference[Woodstock '19]{Woodstock '18: ACM Symposium on Neural Gaze Detection}{June 03--05, 2019}{Woodstock, NY}
\acmBooktitle{Woodstock '19: ACM Symposium on Neural Gaze Detection, June 03--05, 2019, Woodstock, NY}
\acmPrice{15.00}
\acmDOI{10.1145/1122445.1122456}
\acmISBN{978-1-4503-9999-9/18/06}

\setcopyright{acmcopyright}
\acmJournal{IMWUT}
\acmYear{2019} \acmVolume{3} \acmNumber{4} \acmArticle{141} \acmMonth{12} \acmPrice{15.00}\acmDOI{10.1145/3369837}

\begin{document}

%
\title{Intermittent Learning: On-Device Machine Learning on Intermittently Powered System}

%

\author{Seulki Lee}
\affiliation{%
  \institution{University of North Carolina at Chapel Hill}
  \streetaddress{201 S. Columbia St}
  \city{Chapel Hill}
  \country{USA}}
\email{seulki@cs.unc.edu}

\author{Bashima Islam}
\affiliation{%
  \institution{University of North Carolina at Chapel Hill}
  \streetaddress{201 S. Columbia St}
  \city{Chapel Hill}
  \country{USA}}
\email{bashima@cs.unc.edu}

\author{Yubo Luo}
\affiliation{%
  \institution{University of North Carolina at Chapel Hill}
  \streetaddress{201 S. Columbia St}
  \city{Chapel Hill}
  \country{USA}}
\email{yubo@cs.unc.edu}

\author{Shahriar Nirjon}
\affiliation{%
  \institution{University of North Carolina at Chapel Hill}
  \streetaddress{201 S. Columbia St}
  \city{Chapel Hill}
  \country{USA}}
\email{nirjon@cs.unc.edu}

\renewcommand{\shortauthors}{Lee et al.}

%

%
\begin{abstract}
This paper introduces \textit{intermittent learning} --- the goal of which is to enable energy harvested computing platforms capable of executing certain classes of machine learning tasks effectively and efficiently. We identify unique challenges to intermittent learning relating to the data and application semantics of machine learning tasks, and to address these challenges, we devise 1) an algorithm that determines a sequence of actions to achieve the desired learning objective under tight energy constraints, and 2) propose three heuristics that help an intermittent learner decide whether to learn or discard training examples at run-time which increases the energy efficiency of the system. We implement and evaluate three intermittent learning applications that learn the 1) air quality, 2) human presence, and 3) vibration using solar, RF, and kinetic energy harvesters, respectively. We demonstrate that the proposed framework improves the energy efficiency of a learner by up to 100\% and cuts down the number of learning examples by up to 50\% when compared to state-of-the-art intermittent computing systems that do not implement the proposed intermittent learning framework.
\end{abstract}

%
%


%

%

%
\maketitle

\input{tex/1.INTRODUCTION.tex}
\input{tex/2.CONCEPTS.tex}
\input{tex/3.FRAMEWORK.tex}
\input{tex/4.ACTION_PLANNER.tex}
\input{tex/5.SELECTION.tex}
\input{tex/6.IMPLEMENTATION.tex}
\input{tex/7.EVALUATION.tex}
\input{tex/8.LIMITATION.tex}
\input{tex/9.RELATED_WORK.tex}
\input{tex/A.CONCLUSION.tex}
\input{tex/bibliography.tex}

%

%

%

\end{document}

%% file: tex/1.INTRODUCTION.tex
\section{Introduction}

We envision a future where batteryless embedded platforms will be an effective alternative to battery-powered systems. Being batteryless will reduce environmental hazard caused by billions of batteries containing toxic and corrosive materials that are dumped in the environment every year~\cite{zeng2012prediction}. The prolonged life of batteryless systems will eliminate the cost and effort of recharging and replacing batteries and make IoT scalable~\cite{GARTNER2}. In the absence of batteries, electronic devices will be lightweight and miniature. We will be able to develop batteryless implantables and wearables that monitor and control a person's health vitals throughout their entire lifetime~\cite{mosa2017ultrathin}. With this vision in mind, batteryless computing platforms have been proposed in recent years.

With the emergence of batteryless computing platforms, we are now able to execute computer programs on embedded systems that do not require a dedicated energy source. These platforms are typically used in sensing applications~\cite{yerva2012grafting, sudevalayam2011energy, seah2009wireless, kansal2003environmental, gorlatova2010energy}, and their hardware architecture consists primarily of a sensor-enabled microcontroller that is powered by some form of harvested energy such as solar, RF or piezoelectric~\cite{priya2009energy}. Programs that run on these platforms follow the so-called \textit{intermittent computing} paradigm~\cite{maeng2017alpaca, van2016intermittent, xie2016checkpoint, lucia2017intermittent} where a system pauses and resumes its code execution based on the availability of harvested energy. Over the past decade, the efficiency of batteryless computing platforms has been improved by reducing their energy waste through hardware provisioning, through check-pointing~\cite{ransford2012mementos} to avoid restarting code execution from the beginning at each power-up~\cite{balsamo2015hibernus}, and through discarding stale sensor data~\cite{hester2017timely} which are no longer useful. Despite these advancements, the capability of batteryless computing platforms has remained limited to simple sensing applications only.

\begin{figure}[!tb]
    \includegraphics[width=0.48\textwidth]{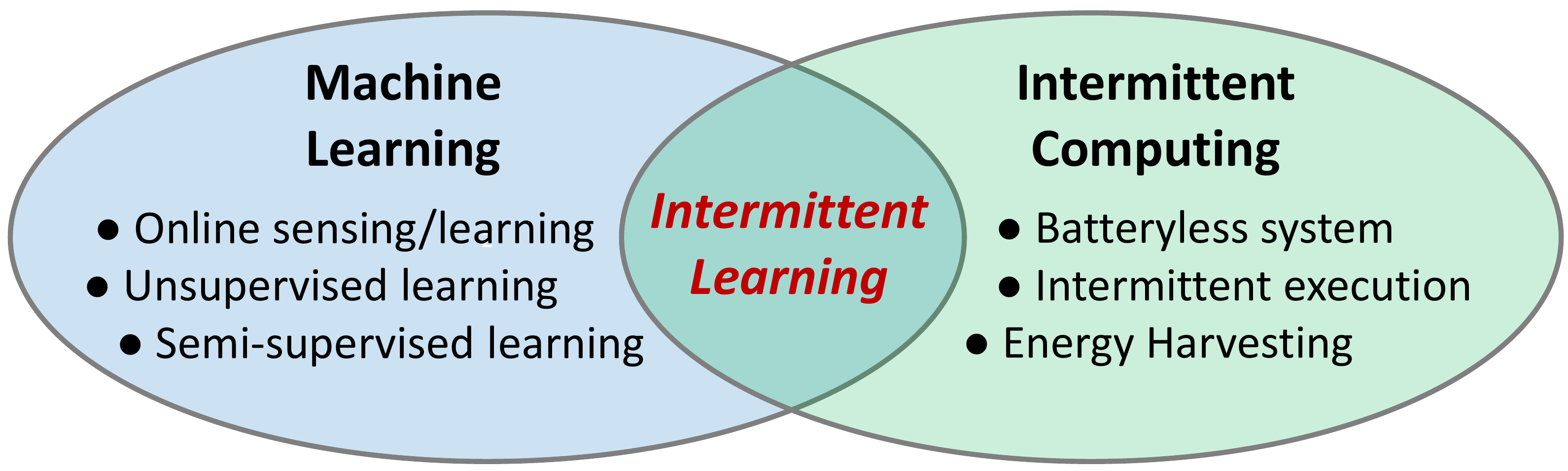}
    \hspace{2mm}
    \includegraphics[width=0.48\textwidth]{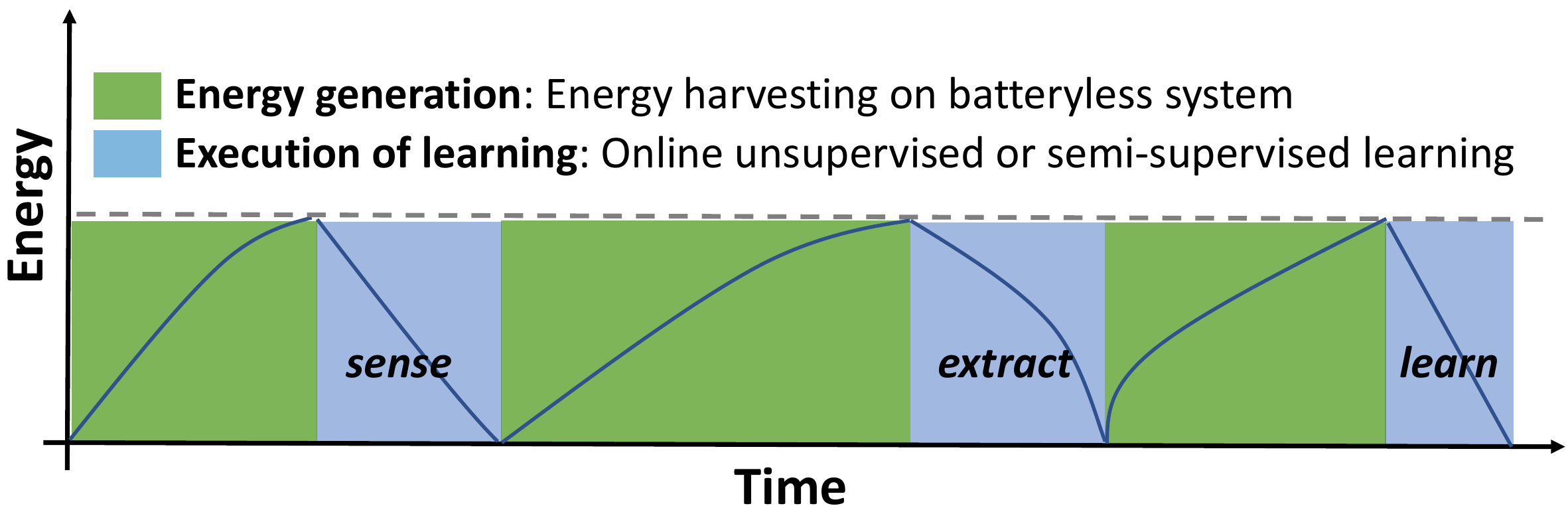}
    \caption{An \textit{intermittent learner} intermittently executes on-device online machine learning algorithms using harvested energy.}
    \label{fig:introduction}
\end{figure} 

In this paper, we introduce the concept of \textit{intermittent learning} (Figure ~\ref{fig:introduction}), which makes energy harvested embedded systems capable of executing lightweight machine learning tasks. Their ability to run machine learning tasks inside energy harvesting microcontrollers pushes the boundary of batteryless computing as these devices are able to sense, learn, infer, and evolve over a prolonged lifetime. The proposed intermittent learning paradigm enables a true lifelong learning experience in mobile and embedded systems and advances sensor systems from being smart to smarter. Once deployed in the field, an intermittent learner classifies sensor data as well as learns from them to update the classifier at run-time---without requiring any help from any external system. Such on-device learning capability makes an intermittent learner privacy-aware, secure, autonomous, untethered, responsive, adaptive, and evolving forever.

The notion of intermittent learning is similar to the intermittent computing paradigm with the primary difference that the program that runs on the microcontroller executes a machine learning task---involving both \textit{training} and \textit{inferring}. Although it may appear to be that all machine learning tasks are merely pieces of codes that could very well be run on platforms that support intermittent computing, for several reasons, a machine learning task in an intermittent computing setup is quite different. The fundamental difference between a machine learning task and a typical task on a batteryless system (e.g., sensing and executing an offline-trained classifier) lies in the data and application semantics, which requires special treatment for effective learning under an extreme energy budget. Existing works on intermittent computing address important problems, such as ensuring atomicity~\cite{maeng2017alpaca,colin2016chain}, consistency~\cite{maeng2017alpaca,colin2016chain, lucia2015simpler}, programmability~\cite{hester2017timely}, timeliness~\cite{hester2017timely}, and energy-efficiency~\cite{colin2018reconfigurable, hester2015tragedy, buettner2011dewdrop}, which enable efficient code execution of general-purpose tasks. Our work complements existing literature and specializes in a batteryless system on efficient and effective on-device learning by explicitly considering the \emph{utility of sensor data} and \emph{the execution order of different modules} of a machine learning task.

Three key properties of intermittent learning make it unique and a harder problem to solve. First, when energy is scarce, an intermittent learning system needs to decide the best action (e.g., learn vs. infer) for that moment so that its overall learning objective (e.g., the completion of learning a desired number and types of examples) is achieved. Second, since not all training examples are equally important to learning, an intermittent learning system should smartly decide to keep or discard examples at run-time, and thus be able to eliminate a large number of unnecessary and energy-wasting training actions. Third, a system that pauses and resumes its executing based on the state of its energy harvester runs a greater risk of missing real-world events that it wants to detect or learn. When both the generation of energy and the generation of training/inferable sensor data are intermittent and uncertain, the problem of learning becomes an extremely challenging feat. None of the existing intermittent computing platforms consider these issues, and thus they are not effective in learning when we execute machine learning tasks on them.

In this paper, we address these aforementioned challenges and propose the first intermittent learning framework for intermittently powered systems. The framework is targeted to a class of learning problems where the presence of energy implies the presence of data---which means either the cause of energy and data are the same, or they are highly correlated, or data is always available for best-effort sensing and inference (e.g., sporadic classification of air quality). Furthermore, we focus on \emph{long-term} and \emph{online machine learning} tasks where a batteryless system is expected to run for an extended period in time, and its learning performance is expected to improve over time. In our proposed framework, the availability of labeled data is not an absolute necessity. In other words, we study \emph{unsupervised}~\cite{russell2016artificial} and \emph{semi-supervised}~\cite{chapelle2009semi} machine learning problems in this paper, although the framework can be easily extended to incorporate supervised~\cite{russell2016artificial} and reinforcement learning~\cite{russell2016artificial} tasks by enabling real-time feedback from the environment or humans.

The intermittent learning framework comes with the necessary tools to develop on-device machine learning applications for intermittently powered systems. We provide a programming model that allows a programmer to develop an intermittent learning application that executes correctly on an intermittent system. Like many existing proposals, we adopt a task-based -- which we call action-based -- intermittent programming model~\cite{colin2018termination,yildirim2018ink,maeng2017alpaca,hester2017timely,colin2016chain,lucia2015simpler}. We provide application programmers with an energy pre-inspection tool that helps them split an existing application code into sub-modules called \emph{actions} that atomically run to completion on intermittently-powered systems. A user study on the proposed programming model shows that the concept of action-based intermittent learning is intuitive and applicable to a variety of applications.

We envision a wide variety of applications where the proposed intermittent learning paradigm applies. Three such applications are implemented and evaluated in this paper to demonstrate the efficacy of the proposed intermittent learning framework. The first one is an air quality learning system where sunlight and air-quality sensitive environmental sensors are powered by harvesting solar energy to detect an anomaly in the air quality. This batteryless learner has been monitoring, classifying, and learning air-quality indices continuously since September 2018. We have developed a webpage showing its real-time learning status\footnote{Intermittent air quality learning system: \url{https://www.cs.unc.edu/~seulki/intermittent-learning/air-quality-learning.html}\label{footnote 1}}. The second application is an RF energy-based human presence learning system that learns to detect humans passing by it in indoor spaces from the variation in RSSI patterns. The last application is a vibration monitoring scenario (applicable to human health and machine monitoring applications) where an accelerometer-based sensing system is powered by harvesting piezoelectric energy. To demonstrate that the proposed framework is portable to different platforms, we have used an AVR, a PIC, and an MSP430-based microcontroller to implement these three applications, respectively. The framework is implemented in C and has been open-sourced~\cite{CODES}.

The main contributions of this paper are the following:

$\bullet$ This is the first work that introduces the \textit{intermittent learning} concept and proposes an intermittent learning framework that enables energy harvested computing platforms to  perform on-device machine learning \textit{training}.

$\bullet$ We define a set of \textit{action primitives} for intermittent learners and devise an algorithm to determine a sequence of actions to achieve the desired learning objective while maximizing energy efficiency. 

$\bullet$ We propose three learning-example selection heuristics that enable an intermittent learner to decide whether to learn or to discard examples---which increase the efficiency in learning under tight energy constraints.

$\bullet$ We provide a programming model and development tool of intermittent learning, which allows a programmer to implement an intermittent learning application based on the action-based intermittent execution.

$\bullet$ We implement and evaluate three intermittent learning applications: an air quality, a human presence, and a vibration learning system. We demonstrate that the proposed framework improves the energy efficiency of a learning task by up to 100\% and cuts down the learning time by 50\%.  

$\bullet$ We have open-sourced the software framework to the community to facilitate the widespread use of the proposed intermittent learning framework. The anonymized code repository can be accessed here~\cite{CODES}.

%% file: tex/2.CONCEPTS.tex
\section{Intermittent Learning -- Motivation and Scope}

The goal of intermittent learning is to enable efficient and effective execution of a class of machine learning tasks on embedded systems that are powered intermittently from harvested energy. Throughout the lifetime, an intermittent learner sporadically senses, infers, learns (trains), and thus evolves its classier and model parameters over time, and get better at detecting and inferring events of interest. Like existing intermittent computing systems, an intermittent learner also pauses its execution when the system runs out of energy and resumes its execution when the system has harvested enough energy to carry out its next action. However, due to the nature of the data and application semantics of a machine learning task, an intermittent learner has to do a much better job in deciding \emph{what actions to perform} and \emph{what data to learn}---so that it can ensure its progress toward learning and inferring events of interests, while making the best use of sporadically available harvested energy.

\subsection{Motivation Behind Intermittent Learning}

On-device machine learning on embedded systems is an emerging research area~\cite{li2018learning,chauhan2018breathing,yao2017deepiot}. Batteryless systems have also joined this revolution. Recent literature on intermittent computing routinely uses \emph{on-device inference} as one of many example applications~\cite{li2018self,li2018battery,truong2018capband,gobieski2018intelligence,hester2017timely,ransford2012mementos}. For example, ~\cite{li2018self,li2018battery} harvests energy from the ambient light to power up a gesture recognition system that implements \emph{Constant False Alarm Rate} (CFAR) algorithm~\cite{scharf1991statistical}, CapBand~\cite{truong2018capband} implements a \emph{Convolutional Neural Network} (CNN) to classify hand gestures on a batteryless system that is powered by a combination of solar and RF harvesters, ~\cite{gobieski2018intelligence} implements a deep network compression algorithm~\cite{han2015deep} to fit a \emph{Deep Neural Network} (DNN) into a resource-constrained microcontroller (MSP430) which runs on energy harvested from RF sources. While these application-specific systems have inspired our work, we observe that these systems are capable of only making \emph{on-device inferences} using an \emph{offline-trained classifier}. These systems treat machine learning tasks the same way as any other computational load, and thus, they are not able to optimize the execution of machine learning-specific tasks. Furthermore, the pre-trained classifiers running on these systems are fixed and non-adaptive, which does not allow these applications to adapt automatically at run-time to improve the accuracy of the classifier.     

To complement and advance the state-of-the-art of the batteryless machine learning systems, we propose the intermittent learning framework which explicitly takes into account the dynamics of a machine learning task, in order to improve the energy and learning efficiency of an intermittent learner in a systemic fashion. The fundamental difference between the proposed framework and the existing literature is that, besides \emph{improving the efficiency of on-device inference}, the intermittent learning framework enables \emph{on-device training} to improve the effectiveness and accuracy of the learner over time. 

\subsection{Alternatives to Intermittent Learning}

An alternative to on-device learning on batteryless systems would be to sense and transmit raw or semi-processed sensor data from a batteryless system to a base station that executes the inference and/or training tasks. In fact, such \emph{offloading} solutions were popular back in the days when \emph{Wireless Sensor Networks} (WSNs) were deployed to collect data from the sensor nodes, only to be analyzed later on a remote base station~\cite{shaikh2016energy, akhtar2015energy,shaikh2016energy, lu2015wireless}. Compared to the sensor motes of those WSNs, today's microcontroller-based systems are far more advanced in terms of CPU and memory, and their energy efficiency has improved by several orders of magnitude. For instance, the latest mixed-signal microcontrollers from Texas Instruments (i.e., TI MSP430 series) comes with up to 16-bit/25 MHz CPU, 512 KB flash memory, 66 KB RAM, and 256 KB non-volatile FRAM---which are comparable to the 16-bit Intel x86 microprocessors of the early 80s which ran MS-DOS. These devices are quite capable of executing simple machine learning workloads that perform on-device classification of sensor data~\cite{ gobieskiintermittent}. In general, there are several advantages of on-device intermittent learning over relaying data to a base station:

$\bullet$ \textit{Data Transmission Cost and Latency.} Data communication between a device and a base station introduces delays and increases energy cost per bit transmission. Using back-scatter communication~\cite{lu2018ambient} apparently lower the energy cost, but the dependency on an external entity and the unpredictable delay in wireless communication still remain, which we want to avoid by design. 

$\bullet$ \textit{Privacy and Security.}
Private and confidential data, such as health vitals from a wearable device, can be safely learned on-device -- without exposing them to external entities. Security problems caused by side-channel and man-in-the-middle attacks~\cite{aziz2009detecting,kugler2003man} are avoided by design when we adopt on-device processing of sensitive data.

$\bullet$ \textit{Precision Learning and Resource Management.} Many human-in-the-loop machine learning applications running on wearable and implantable systems benefit from run-time adaptation as different persons have different preferences and different expectations from the same application. On-device learning helps a system adjust itself at run-time to satisfy each individual's needs and to optimize its own resource management.

$\bullet$ \textit{Adaptability and Lifelong Learning.} Lifelong learning~\cite{chen2016lifelong} is an emerging concept in robotics and autonomous systems where the vision is to create intelligent machines that learn and adapt throughout their lifetime. Intermittent learning enables true lifelong learning by liberating these devices from being stationary and connected to power sources, to mobile, ubiquitous, and autonomous. 

We acknowledge that some of the pitfalls of offloading machine learning tasks to base stations can be avoided via alternative methods. For instance, on-device data encryption arguably can ensure security and privacy, backscatter techniques can reduce communication energy cost, and over-the-air code updates could make the classifier adaptive. However, each of these comes with their limitations and overheads, and none are maintenance-free. Hence, considering the autonomy and maintenance-free nature of intermittent learners, combined with the full package of benefits mentioned earlier, we opt for batteryless on-device learners as our design choice.          

\subsection{The Scope of Intermittent Learning}

We limit the scope of this paper to specific types of machine learning problems and study the corresponding research challenges.

$\bullet$ \textit{Online Unsupervised and Semi-Supervised Learning.} Based on the availability and use of labeled ground-truth data, a machine learning problem can be categorized into supervised, semi-supervised, and unsupervised types~\cite{russell2016artificial}. Since batteryless computers are meant to last long and operate unattended, we exclude purely supervised learning (where labeled data is a must) from the scope of this work. Instead, we focus on the other two types, where either labeled data are unnecessary (unsupervised) or some labeled data are available for use (semi-supervised). For instance, a motion-activated intermittent learner can observe sensor readings over time and look for statistical anomalies (e.g., using an outlier detection or a cluster analysis algorithm) in its data stream. In many applications, these statistical anomalies are the ones that correspond to events of interests such as fall detection, aggressive behavior recognition, and intruder detection. Furthermore, we consider online machine learning problems where examples (i.e., a vector of sensor readings that we want to classify or learn) come one at a time, and the classifier is incrementally trained and updated as they arrive.

$\bullet$ \textit{Selection of Training Data.} In an online learning task, a learner's model parameters are updated as new training examples arrive. A typical learning algorithm takes hundreds of iterations of model updating -- one iteration for each training example -- before the learner attains a reasonable classification accuracy. However, in a real-world online learning scenario, a system might continue to receive too many similar examples and use them all to update the learner's model parameters. In such cases, the learner wastes a significant amount of energy and compute cycles in repeatedly learning the same example where learning only one representation example would have been sufficient. In summary, since not all training examples are equally important to learning, it is beneficial to discard examples that do not contribute to a learner's gain in accuracy.

$\bullet$ \textit{Choice of Actions at Run-time.} A machine learning task includes several sub-tasks (or, \emph{actions}) such as sampling the sensors, assessing the utility of sensor values in learning, saving sensor values for later use, updating the classifier model upon sensing a new data point, classifying the sensor data, and sending alerts to external systems. When the system harvests enough energy to take one or more of these actions, it must determine the best action for that moment so that its overall learning objective (e.g., the completion of its learning task and/or learning a desired number and types of examples) can be fulfilled. 

For instance, suppose, a system has harvested just enough energy to either update the current model parameters by training the learner with recently sampled data or to classify the new data using the current model. Based on the learner's performance at that moment, either action can be a valid choice. If the learner is under-performing, retraining is a more sensible action. On the other hand, if the learner is performing at its best, it makes more sense to do frequent classifications than training. Hence, dynamically choosing a proper action is an important aspect of intermittent learning, which is not considered by existing intermittent computing systems.

If we employ existing intermittent computing frameworks like MayFly~\cite{hester2017timely} to execute machine learning tasks, the system would blindly use every incoming training example to update the model parameters and thus drain the harvested energy much faster than needed. Although it considers the \textit{staleness} of data to increase the system lifetime, it does not help a learner as the data can be fresh, yet their utility toward an application's high-level goal can be null. Likewise, data can be stale, yet their utility in a learning algorithm can be high. Hence, we need to devise a mechanism to smartly choose or discard examples at run-time, and thus be able to eliminate a large number of unnecessary and energy-wasting training actions.       

$\bullet$ \textit{Occurrence of Sample Data and Energy Harvesting Cycles.} An intermittent learner learns and infers physical world events. Occurrences of these events are, in general, unpredictable. Energy harvesting cycles also depend on physical world phenomena such as motion, sunlight, or radio signals, and thus, the time and amount of harvested energy are unpredictable as well. Hence, an intermittent learner has to learn through these dual uncertainties. We identify two cases when intermittent learning is suitable.

In some applications, the physical phenomena behind the event of interest and energy harvesting are either the same or strongly correlated. For instance, piezoelectric harvesters that generate energy from motion are used in many people-centric machine learning applications, such as vibration-related health condition monitoring, sleep motion detection, and fall detection, where the core learning task is to classify human motions. In this class of applications, data and energy are available at the same time, and they are correlated. Intermittent learning framework applies to these applications with greater certainty of learning.

There is another class of sensing and inference applications where the data are either always available, or the rate of change in data is so low that an intermittently powered system can gather sufficient data during its operating cycles. Examples include environmental monitoring applications such as detecting pollutants or gaseous anomaly in the air (e.g., excessive carbon dioxide concentration), and sound pollution monitoring. An intermittent learner in these scenarios learns and infers in a best-effort manner.

%% file: tex/3.FRAMEWORK.tex
\section{Intermittent Learning Framework}

\subsection{Framework Overview}

\begin{wrapfigure}{R}{0.40\textwidth}
    \vspace{-1.5em}
    \centering
	\includegraphics[width=0.40\textwidth]{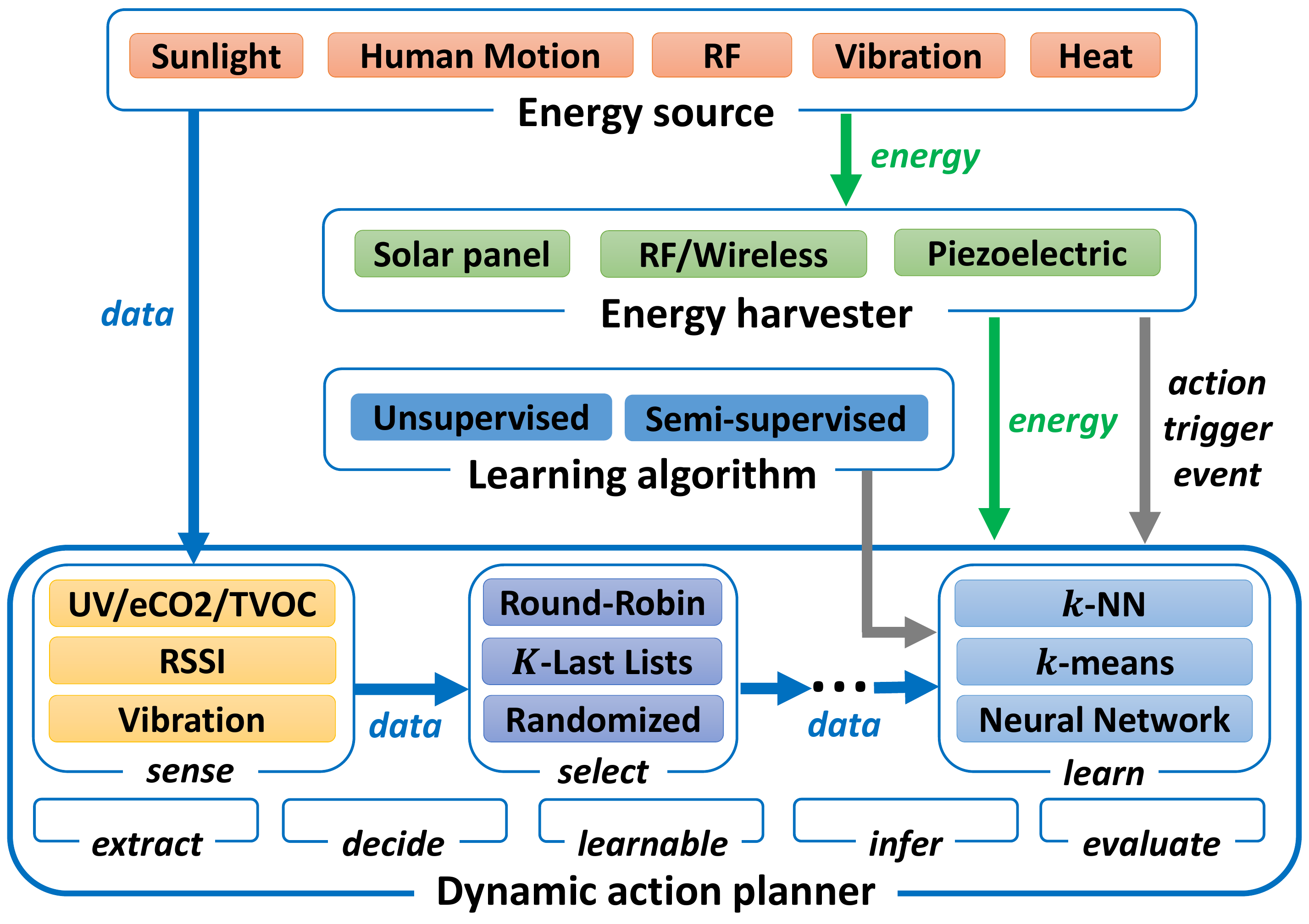}
	\caption{The intermittent learning framework showing energy sources, energy harvesters, learning algorithms, and a dynamic action planner.}
	\label{fig:architecture}
	\vspace{-1em}
\end{wrapfigure}

We propose an intermittent learning framework for intermittently powered systems that want to execute an end-to-end machine learning task which involves data acquisition, learning, and inferring. Figure~\ref{fig:architecture} shows a high-level architectural diagram of the proposed framework. The three main modules of the proposed framework corresponding to energy management, machine learning, and task planning are briefly discussed as follows:  

$\bullet$ \textit{Energy Harvester.} Batteryless computing platforms consist of one or more energy harvesters such as piezoelectric, RF, or solar panels that harvest energy from various types of sources such as sunlight, motion, RF, vibration, wind, heat, and chemical. This subsystem monitors the energy generated by the energy harvester and generates an interrupt that triggers an intermittent execution of learning tasks whenever a sufficient amount of energy is generated. In certain systems, such as ~\cite{truong2018capband}, where multiple energy harvesters are used to guarantee continuous energy supply, e.g., RF for indoors and solar for outdoors, the energy harvester subsystem takes care of selecting and switching to the preferred harvester transparently.   

$\bullet$ \textit{Library of Learning Algorithms.} We have developed a library of machine learning algorithms which contains specialized implementations of commonly used unsupervised or semi-supervised algorithms for an intermittently powered system. These algorithms are split into small pieces of code so that they are suitable for executing the intermittently powered system. Currently, the library contains the implementation of three common machine learning algorithms as templates: $k$-nearest neighbors, $k$-means, and a neural network (described later in this section). While these are able to solve many practical learning problems, if a new learning algorithm needs to be implemented for an intermittent execution, a developer can follow the modular implementation of these classifiers to get inspired on how to implement a custom algorithm in an intermittent fashion. 

$\bullet$ \textit{Dynamic Action Planner.} This module is the heart of the framework, which is responsible for selecting the right action at the right moment in order to advance the learning task toward achieving its desired learning objectives. It contains implementations of intermittently executable methods and algorithms to schedule actions, to select what to learn, and to evaluate the progress of an intermittent learner toward task completion. This module is described in detail in Section~\ref{sec:planning}.

\subsection{Action Primitives} 

\begin{figure} [t]
\begin{minipage}{0.49\textwidth}
\captionsetup{type=table}
\small
\begin{center}
\caption{List of Action Primitives.}
\label{table:actions}
\vspace{-10pt}
\begin{tabular}{r p{18em}}
    \toprule
    \textbf{Action} & \textbf{Description} \\
    \hline
    \textit{sense} & Sense and convert data to an example. \\
    \textit{extract} & Extract features from an example. \\
    \textit{decide} & Decide to \textit{learn} or \textit{infer}.\\
    \textit{select} & Determine whether a training example increases the learning performance.\\
    \textit{learnable} & Check prerequisites of a \textit{learn} action.\\
    \textit{learn} & Execute a learning algorithm intermittently.\\
    \textit{evaluate} & Evaluate the learning performance.\\
    \textit{infer} & Make an inference using the current model.\\
    \hline
\end{tabular}
\end{center}
\end{minipage}\hspace{2mm}
\begin{minipage}{0.49\textwidth}
	\centering
	\includegraphics[width=0.95\textwidth]{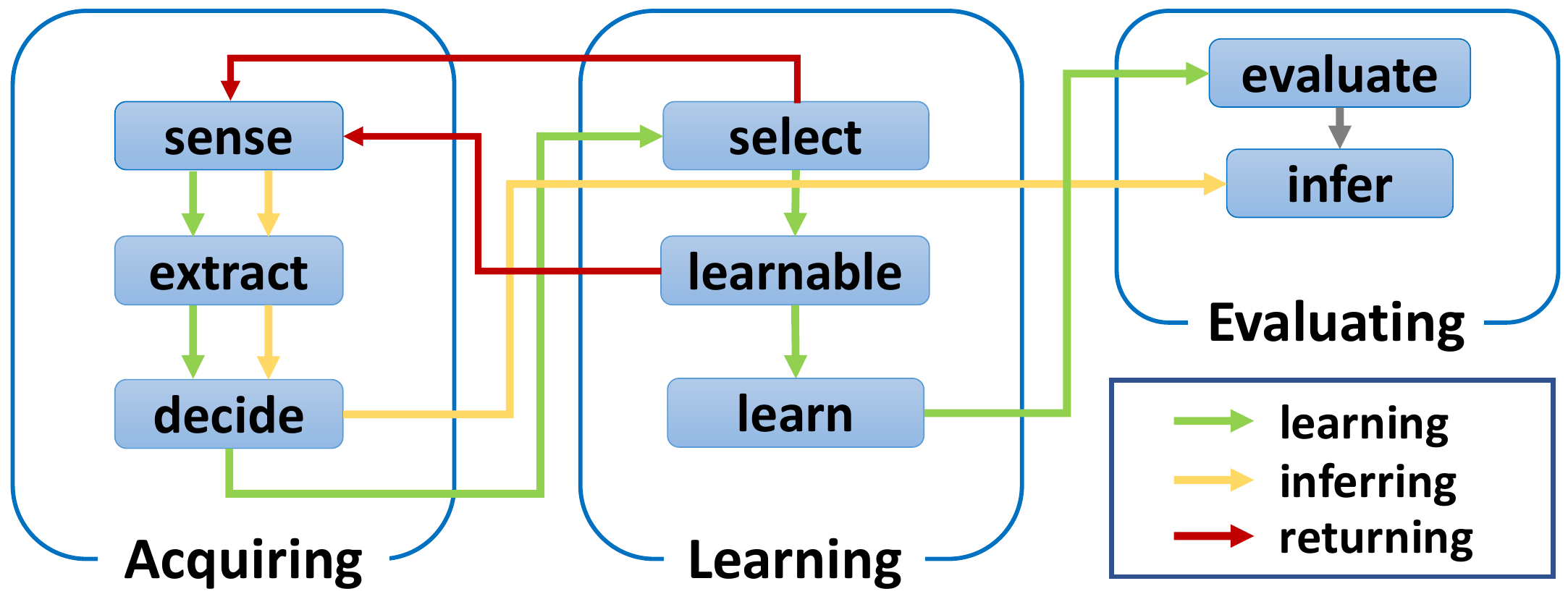}
	\caption{Action state diagram showing all actions and how they interact with each other.}
	\label{fig:action_state_diagram}
\end{minipage}
\end{figure}

We identify eight basic operations---which we refer to as \textit{actions}---that an intermittent learner \emph{may} execute in its lifetime. A complete list of actions and their brief description are presented in Table~\ref{table:actions}. Breaking a task into pieces is similar to existing task-based intermittent computing frameworks~\cite{colin2018termination,yildirim2018ink,maeng2017alpaca,hester2017timely,colin2016chain,colin2016chain,lucia2015simpler} with the difference that each action in an intermittent learning framework is associated with a semantic meaning, and the set of actions being exhaustive, we are able to optimize their execution better than a general-purpose program.

Some of these actions such as \textit{sense}, \textit{extract}, \textit{learn}, and \textit{infer} are self-explanatory. The action \textit{decide} makes a decision to execute either a \textit{learn} or an \textit{infer} action based on the learning objective (desired goal states) of a learner described in Section~\ref{sec:desired_goal_states}. \textit{Select} is related to choosing a suitable training example for learning. Heuristics for choosing training examples are described in Section~\ref{sec:selectex}. \textit{Learnable} is used to enforce preconditions of a learning algorithm, e.g., clustering algorithms require a minimum number of examples so that they can form clusters. The action \textit{evaluate} is related to the performance of the current learning model and action planning, which is described in Sections~\ref{sec:planning}.

\subsection{Action State Diagram}

A learning task involves a subset of the actions that must be executed in a certain order. An intermittent learner has to enforce this ordering of actions when executing them at run-time. For instance, \textit{sense} precedes all actions as this is where raw sensor readings are converted into an object, which we call an \textit{example}, that is processed further. Similarly, \textit{learn} or \textit{infer} cannot be executed until we execute \textit{extract} to extract features from an example to represent them in terms of feature vectors. Figure~\ref{fig:action_state_diagram} shows a state diagram consisting of all eight actions along with the direction of data flow between two consecutive actions in an execution order. For ease of understanding, we categorize them into groups of acquiring, learning, and evaluating actions.

\subsection{Intermittent Action Execution}

Several of the actions in Table~\ref{table:actions} are larger than what it takes to execute them at one shot by an intermittent learner. The limit comes from the size of the energy storage, i.e., the size of the capacitor that stores harvested energy, that can keep the system awake for a limited period in time. The size of the capacitor cannot be made arbitrarily large as that increases the charging time, and a longer charging time will result in excessive delays in sensing and processing of new data. In general, an intermittent learner sleeps and wakes up multiple times during the execution of an action. In this section, we describe how an action is implemented to make it suitable for intermittent execution by the proposed framework.  

$\bullet$ \textit{How to program actions for an intermittent execution?} An application developer implements or overrides all or a subset of the action primitives. Corresponding to each action, there is an ordered list of functions, where each function executes a part of the action that is small enough for running to completion at one shot (i.e., without interruptions). Actions can be bypassed (not programmed) if a learning algorithm does not require them. Listing~\ref{lst:user_actions} shows an example of four user-programmed actions (\textit{sense}, \textit{extract}, \textit{select}, and \textit{learn}). With the \textit{learn} action being large, it has been split into three smaller functions. An array of function pointers is implemented in the framework to facilitate an orderly execution of these parts of an action.

$\bullet$ \textit{How to determine if an action requires splitting?} Application programmers are provided with a \textit{battery-powered} development tool that guides the action splitting process. The tool checks if each action written by the programmer can be completed using a certain amount of energy, which is also specified by the programmer. We call this \textit{energy pre-inspection}-- which is an automated tool that identifies and warns if an action requires more energy than the target. This tool helps a programmer interactively split implemented modules until they fit into the target energy. The details of action decomposition are described in Section~\ref{subsec:programming_model}. 

\begin{minipage}{.48\textwidth}
\begin{lstlisting}[language=C, xleftmargin=0, caption={User-programmed action example.}, label={lst:user_actions},basicstyle=\small,xleftmargin=-0.04\textwidth] 
/* actions.c */
/* learning actions programmed by user */
int sense() { /* user-defined code of sense */ }
int extract() { /* user-defined code of extract */ }
int select() { /* user-defined code of select */ }
int learn_1() { /* user-defined 1st part of learn */ }
int learn_2() { /* user-defined 2nd part of learn */ }
int learn_3() { /* user-defined 3rd part of learn */ }

/* list of each action */
int (*sense_[])() = { sense };
int (*extract_[])() = { extract };
int (*select_[])() = { select };
int (*learn_[])() = { learn_1, learn_2, learn_3 };
\end{lstlisting}
\end{minipage}
\hspace{2mm}
\begin{minipage}{.48\textwidth}
\begin{lstlisting}[language=C, caption={Brief workflow of intermittent learning.}, label={lst:intermittent_learning},basicstyle=\small] 
/* intermittent_learning.c */
int (*dynamic_action_planner())() {
    // code for selecting next action
    return next_action; 
}
void action_trigger() { // action-trigger event ISR
    action = dynamic_action_planner(); // next action
    action(); // execute selected next action
}
void main() {
    init_actions(); // executed only once
    set_interrupt(); // setup action-trigger event
    sleep(); // enter low-power mode
}
\end{lstlisting}
\end{minipage}

$\bullet$ \textit{Who invokes these actions?} At each wake-up, the dynamic action planner routine is called upon by the framework to select an action to execute. Listing~\ref{lst:intermittent_learning} shows a code snippet showing three functions, including the \texttt{main()}. The function \texttt{action\_trigger()} is executed at each wake up and it calls \texttt{dynamic\_action\_planner()} to get a pointer to an action to execute.

\subsection{Intermittent Learning Programming Model}
\label{subsec:programming_model}

We provide a programming interface that allows a programmer to develop an intermittent learning application that executes correctly when a system is intermittently powered.

$\bullet$ \textit{Action-based Programming.} 
Similar to the task-based intermittent computing platforms~\cite{colin2018termination,yildirim2018ink,maeng2017alpaca,hester2017timely,colin2016chain,lucia2015simpler}, an action in the proposed intermittent learning framework is a user-defined block of code. An action, given sufficient energy to execute to completion, is guaranteed to have memory-consistency and control-flow that can be equivalently achieved with a continuously-powered execution. If power fails during an action's execution, the intermittent learning framework discards the intermediate results, and the action starts over from the beginning when power becomes available again by keeping track of the completion status of each action. Actions that consume more energy than the maximum energy budget that the hardware can support need to be decomposed into smaller actions.

\begin{figure} [!tb]
\begin{minipage}{0.49\textwidth}
	\centering
	\includegraphics[width=0.95\textwidth]{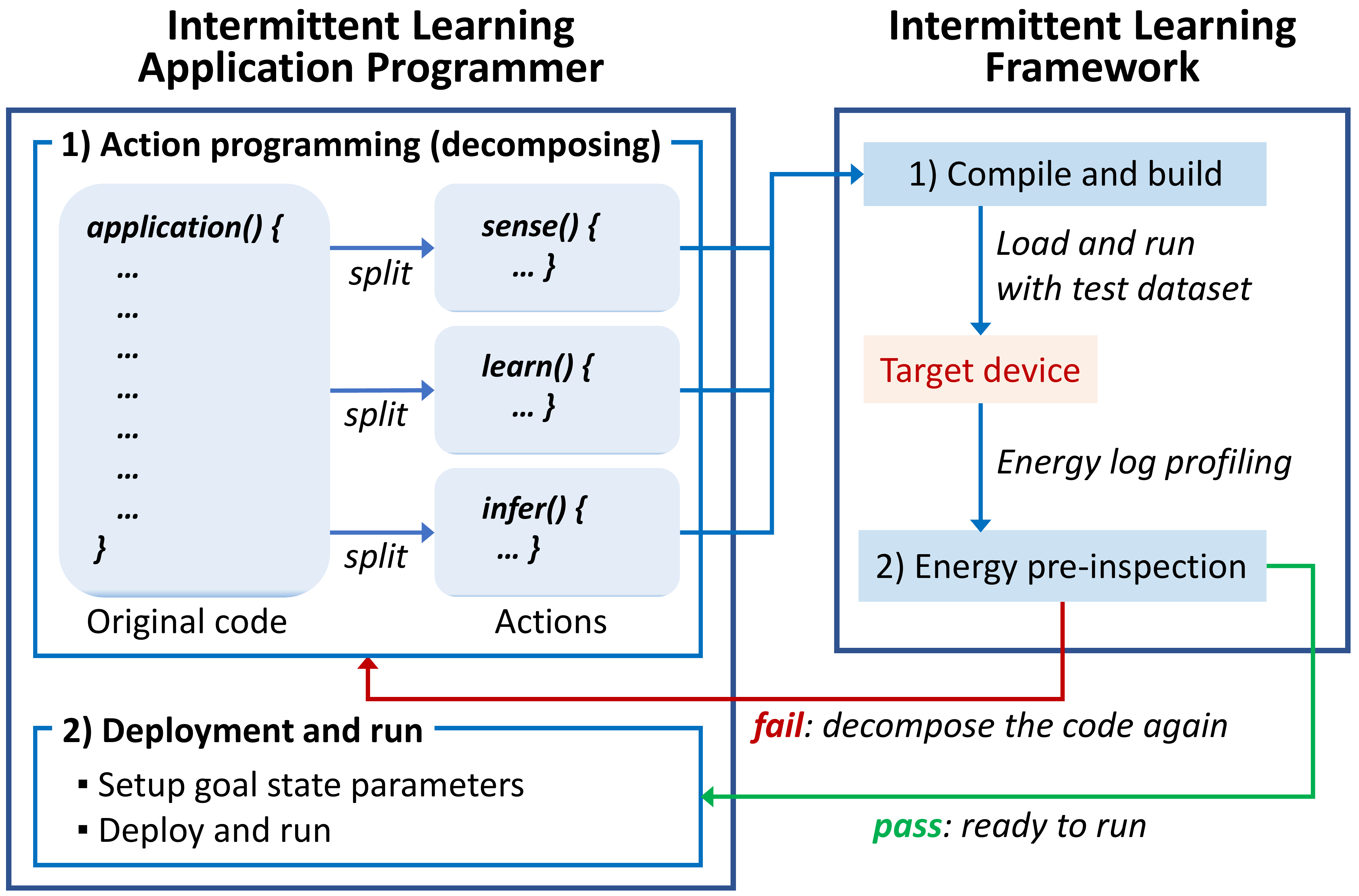}
	\caption{Illustration of the programming model.}
	\label{fig:development_procedure}
\end{minipage}\hspace{2mm}
\begin{minipage}{0.49\textwidth}
\captionsetup{type=table}
\small
\begin{center}
\caption{The result of the user study. The scale for questions Q1-Q3: 1 = the easiest, and 10 = the most difficult.}
\vspace{-10pt}
\label{table:user_study}
\resizebox{\textwidth}{!}{
\begin{tabular}{l|c|c|c}
& \textbf{Avg} & \textbf{Min} & \textbf{Max} \\
\hline
\textbf{Q1. In a scale 1-to-10, how easy} & 5.4  & 2 & 10 \\
\textbf{did you find to understand the concept} & & & \\
\textbf{of action-based intermittent learning?} & & & \\
\hline
\textbf{Q2. In a scale 1-to-10, how easy} & 5.6  & 2  & 10 \\
\textbf{did you find to split the code?} & & & \\
\hline
\textbf{Q3. In a scale 1-to-10, how easy} & 3.7  & 1  & 10 \\
\textbf{did you find to calculate the total} &  &   &  \\
\textbf{energy consumption of the code?} & & &  \\
\hline
\textbf{Q4. How much time did you spend} & 14  & 3  & 30 \\
\textbf{to split the code (in minutes)?} & & & \\
\end{tabular}
}
\vspace{.3cm}
\end{center}
\end{minipage}
\end{figure}

$\bullet$ \textit{Memory Model.} Similar to task-based intermittent computing platforms~\cite{maeng2017alpaca}, the atomicity of actions is guaranteed by maintaining two types of data --- \emph{global} data that are shared between actions and \emph{local} data that reside in a single action. Different actions can share global data by using \emph{action-shared variables}, which are named in the global scope and allocated in the non-volatile memory. Once an action completes writing a value to an action-shared variable, the value can be read by any action by referencing the variable name. Local data are scoped only to a single action like ordinary local variables in a function and are allocated in the volatile memory.

$\bullet$ \textit{Application Development.} Figure~\ref{fig:development_procedure} depicts the development process of an intermittent learning application. To develop a new application, the programmer decomposes the application code into actions by implementing or overriding all or a subset of action primitives which are executed in the order defined by the state diagram. Once actions are implemented, \emph{energy pre-inspection} is performed to make sure that no action consumes more energy than the hardware can support. The energy pre-inspection is performed by a custom tool that we developed by extending TI's EnergyTrace++~\cite{EnergyTrace}, which comes with the intermittent learning framework. The tool first loads and runs the compiled binary on the battery-powered target device and measures the energy consumption of each action using EnergyTrace. In order to obtain the worst-case energy consumption of an action at reasonably high confidence, the target device runs all test cases from all datasets as the input. This is done to maximize the chances of the system to execute different control flows and data-based branches. The tool analyzes the log file of energy measurements and lists all actions that consumed more energy than the maximum allowed and prompts the programmer to split those actions further until all actions pass the test. Finally, the binary that passes the energy pre-inspection is pushed to the target batteryless device.

$\bullet$ \textit{User Study.} We conduct a user study to understand 1) whether the concept of action-based intermittent learning is intuitive and applicable to applications, and 2) the intermittent learning framework provides the necessary components to write on-device machine learning programs on intermittently powered systems.

\begin{wrapfigure}{R}{0.40\textwidth}
	\centering
	\includegraphics[width=0.40\textwidth]{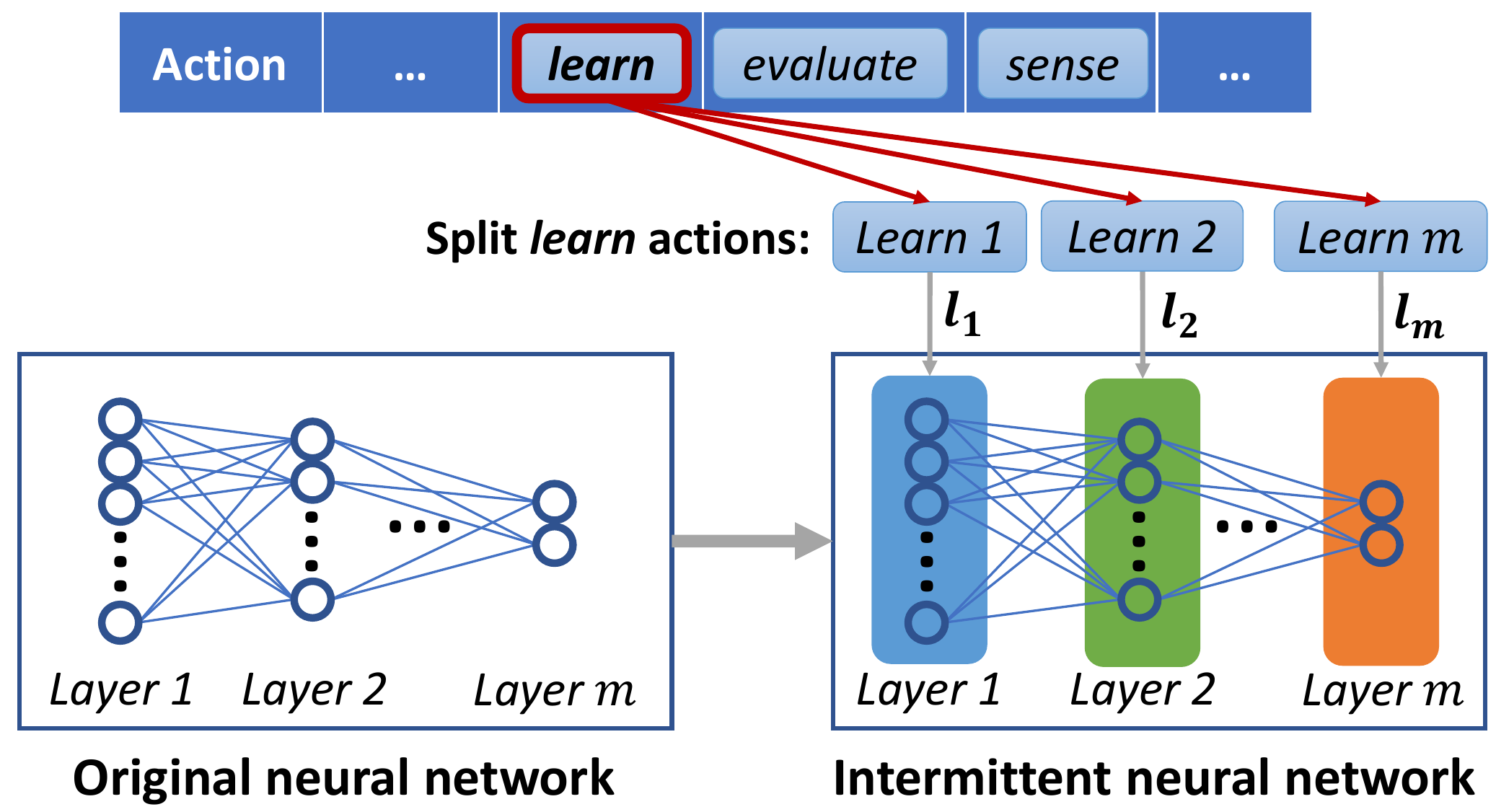}
	\caption{An example of intermittent execution of a back-propagation algorithm to train a neural network. The original network is segmented into layers, and each layer is intermittently executed. Both feed-forward and back-propagation are performed layer by layer in an intermittent manner.}
	\label{fig:intNN}
\end{wrapfigure}

The study involved 35 undergrad computer science students (15 female and 20 male) who were provided with an application code having three large functions (actions), and were asked to decompose and reprogram it into actions having a certain energy budget. Prior to the study, a short introduction to the concept of intermittently-powered systems and the energy constraints associated with programming such systems was provided. After the 30 minute experiment, the participants assessed the difficulty and intuitiveness of the programming model by answering the questions shown in Table~\ref{table:user_study}. On average, the participants assessed that the difficulty level of decomposition is moderate (5.4 and 5.6), and they spent 14 minutes to split the code. We acknowledge that the study is limited due to the small sample size and difficulty in testing multiple applications. Nevertheless, the user study shows that the developers with basic programming knowledge can easily program an intermittent learning application without any significant trouble.

\subsection{Example: An Intermittent Neural Network}

Among all the actions in Table~\ref{table:actions}, in general, the \textit{learn} action has a higher complexity than most others. Hence, we discuss an intermittent execution of it as an illustration. In particular, we illustrate how a feed-forward neural network learner is executed intermittently (Figure~\ref{fig:intNN}). We choose an execution strategy where each layer of the neural network is processed at a time. This is the same network which is later used in the neural network-based $k$-means algorithm in the vibration learning application in Section~\ref{sec:vibration_learning}. 

Figure~\ref{fig:intNN} shows that when the dynamic action planner decides to launch a \emph{learn} action, each of the $m$ layers of the original neural network $\{l_1, l_2, \dots, l_m\}$ gets executed sequentially in the forward direction (feed-forward) and then in the backward direction (back-propagation) to complete one cycle of learning. The system continues to execute each layer $l_i$ as long as the current energy level is higher than required. Once a cycle is completed, the dynamic action planner gets back the control and chooses the next action.

%% file: tex/4.ACTION_PLANNER.tex
\section{Dynamic Action Planner}
\label{sec:planning}

In this section, we describe the \emph{dynamic action planner} which determines a sequence of actions in an online manner. Whenever a sufficient amount of energy is harvested to execute at least one action, the planner dynamically selects the best action that should be performed next, considering the current energy level and the performance of the learner over a short time horizon in the future.

\subsection{System State and Transitions}

We define the state of the proposed system in terms of the examples that are currently in the system and their execution status. Note that the \emph{state of the system} is different from the \emph{action state diagram} (Figure~\ref{fig:action_state_diagram}) which does not involve the execution status of the examples.

For instance, at the beginning of the system, there is no example inside the system. The first time the system harvests enough energy to act, it \textit{senses} new data $x_i$ and then waits for the next action. We denote this state as $\{(x_i, sense)\}$. The next time the system harvests energy, it has more options depending upon the amount of harvested energy, e.g., it can either \textit{sense} a new data $x_{i+1}$, or execute the next action \textit{extract} on $x_i$. This results in two possible next states: $\{(x_i, sense), (x_{i+1}, sense)\}$ and $\{(x_i, extract)\}$. In general, given a set of examples in the system, $X = \{x_1, \dots, x_N\}$ and the supported actions by the system, $A = \{a_1, \dots, a_K\}$, the state of the system, $S$ is defined by a set of two-tuples $\{(x_i,a_j)\} \subset X\times A$, which denotes that the most recent action performed on $x_i$ is $a_j$. 


A transition from one state $S$ to another state $S'$ happens in one of the following two ways:

$\bullet$ The dynamic action planner may choose to sense new data. In this case, a new example $x_{N+1}$ enters the system, resulting in an addition of a new tuple of the form $(x_{N+1}, sense)$ to the system state. Hence,  $S' = S \cup \{(x_{N+1}, sense)\}$.

$\bullet$ A tuple $(u,v) \in S$ is chosen by the dynamic action planner. The system determines the next action $v'$, for example, $u$ in accordance with the action state diagram of Figure~\ref{fig:action_state_diagram}, and either takes action $v'$ on $u$, or $u$ leaves the system if there are no next actions. Hence, the new state $S'$ is either $\{S - (u,v)\} \cup \{(u,v')\}$, or just $S - \{(u,v)\}$.

\subsection{Desirable Goal States}
\label{sec:desired_goal_states}

The goal of the dynamic action planner is to advance the current system state toward a desirable goal state via a series of state transition decisions. The goal state of an online learning system, especially in the absence of labeled ground truth data, is defined in terms of \emph{the rate of examples learned}, \emph{the rate of inferences performed}, or \emph{a combination of these two rates}. For instance, a common strategy is to maintain a desirable learning rate, $\rho_l$ (i.e., learned examples in $L$ energy harvesting cycles) in the beginning, and once the system has learned a desirable number of example $n_l$, the goal is reset to maintaining a desirable inference rate, $\rho_c$ (i.e., inferring the desired number of examples in $L$ energy harvesting cycles). Parameters such as $\rho_l, n_l, \rho_c, L$ are application dependent and are determined via empirical studies and from domain expertise.

However, for some applications, the empirical parameters may not bring the desired behavior as the learning environment (e.g., distribution of input examples) changes over time. To overcome this, intermittent learning systems should learn and update the goal state parameters. For example, by evaluating the need for further learning (e.g., via human feedback or obtaining inference results from more capable externals systems) the parameters can be readjusted at run-time. The system can also continue to build statistics on the frequency of learning based on the utility of learning examples obtained from the example selection methods discussed in Section~\ref{sec:selectex}. In our current implementation of the framework, we use empirically determined parameters. We leave the research on automatic parameter adaptation strategy as future work.

\subsection{Selecting an Action}

$\bullet$ \textit{Action Selection.} For a learner that learns and evolves throughout its lifetime, the process of selecting the best action at every decision point is a never-ending search process as the decision horizon consisting of all future steps is open-ended and infinite. Furthermore, since each state has more than one possible next states, the state-space of the system grows exponentially. Hence, if we aim at selecting a globally best sequence of decisions, depending on the nature of the desired goal state, the optimization algorithm may take forever to find a solution. 

To handle this state explosion problem, we consider a \textit{finite decision horizon} on which we search for a locally best solution. In other words, at each decision point, the action planner looks ahead at all possible resultant states due to the next $L$ transitions to find a sequence of state transitions that take the system closest to a goal state. From our experience, $L$ should be in the order of the longest path on the action state diagram. Once the sequence is obtained, only the first action corresponding to the first state transition is selected for execution.

$\bullet$ \textit{Increasing Planning Efficiency.}
Even within a finite horizon of length $L$, the planner has to consider a large number of states. For instance, assuming $N$ examples currently in the system and a horizon of length $L$, there are $\mathcal{O}(N^L)$ states for the planner to explore. To improve the efficiency of the search, we take additional measures during state-space unfolding, i.e., limiting the number of admitted examples, limiting the value of $L$, bypassing some boolean actions like \textit{select} and \textit{learnable} at random (with a low probability) and using their default return value instead, and combining lightweight actions with succeeding actions. The last two refinements reduce the dwell time of an example in the system, and thus reduces the average number of active examples within the decision horizon.

%% file: tex/5.SELECTION.tex
\section{Selecting Examples to Learn} \label{sec:selectex}

An intermittent computing system must be very keen on exploiting every opportunity to save energy. In an intermittent learning scenario, a substantial amount of energy is saved when a learner selects a minimal subset of training examples that yield a comparable learning performance to using the full training set. This section describes how the framework decides whether an example should be used to retrain the current classifier. At first, we describe four well-known example selection criteria in machine learning~\cite{kabkab2016dcnns, brown2012experimental}. Then we describe three heuristics that meet one or more of these criteria and are currently implemented in the proposed framework.   

\subsection{Desired Criteria for Selecting Examples}

Before proposing metrics to quantify the utility of an example toward learning performance, we list a set of desired criteria for the chosen subset, $B$ of a given training set, $T$.

$\bullet$ \textit{Uncertainty.} The current learning model, $\theta$ should be less certain about an example $x \in B$ belonging to any class, $y$. Otherwise, $x$ does not bring new information to the current learner. This can be expressed as: 
\begin{equation}
x = \argmax_{x \in B} \Big(-\sum_y P(y|x,\theta) \log P(y|x,\theta) \Big)
\end{equation}

$\bullet$ \textit{Balance.} The set of chosen examples $B$ should have a balanced selection from all classes. Otherwise, the learner will be biased toward the class that has more training examples. 

$\bullet$ \textit{Diversity.} The chosen examples $x \in B$ should be diverse within themselves. Otherwise, the set of chosen examples will have redundancy. Therefore, given a dissimilarity metric $d(x_i,x_j)$, we maximize the mean distance between all pairs of selected examples:
\begin{equation}
\argmax_{B \subset T} \frac{1}{|B|^2} \sum_{x_i \in B} \sum_{x_j \in B} d(x_i, x_j)
\end{equation}

$\bullet$ \textit{Representation.} The left-out examples should have representatives in the chosen set, $B$. Otherwise, a learner will miss important information that may be left out in the non-selected set. Therefore, we should minimize the average distance between selected and non-selected examples:
\begin{equation}
\argmin_{B \subset T} \frac{1}{|B|\times |T-B|} \sum_{x_i \in B} \sum_{x_j \in T-B} d(x_i, x_j)
\end{equation}

The balance criterion has been analytically proven by the machine learning community to increase the convergence rate of gradient-based iterative learning algorithms~\cite{brown2012experimental}. Likewise, the other three criteria, i.e., uncertainty, diversity, and representation have been also proven to increase learning performance~\cite{kabkab2016dcnns}.

\subsection{Proposed Online Example Selection Heuristics}

Selecting a subset of the training set that satisfies all or most of the above criteria are computationally expensive. Furthermore, in an online learning scenario, the full training set is not readily available as the learner observes examples one at a time over its lifetime. Hence, in order to determine if an example should be learned by an intermittent learner, we devise three simple yet effective heuristics that are incorporated into the framework:

$\bullet$ \textit{Round-Robin}. To ensure \textit{balance}, selected examples fall into $k$ clusters in a round-robin fashion. Assuming $n$ examples have so far been used to obtain clusters with centroids $\mu_1, \dots, \mu_k$, example $x_{n+1}$ is selected if the following condition is true:  
\begin{equation}
    1 + n\bmod k = \argmin\limits_{1 \le j \le k} d(x_{n+1}, \mu_j)
\end{equation}

$\bullet$ \textit{k-Last Lists.} To ensure \textit{diversity} and \textit{representation}, we maintain two $k$-element lists $B$ and $B'$ that keep track of the last $2k$ examples that were selected and not selected, respectively. The \textit{diversity} and \textit{representation} scores (as described in the previous subsection) are calculated using the lists $B$ and $B'$. A new example $x_i$ is selected if both of the following conditions are met:
\begin{equation}
\begin{split}
     diversity~(B\cup\{x_i\}) &> diversity~(B) \\
    representation~(B\cup\{x_i\}, B') &< representation~(B, B')   
\end{split}
\end{equation}

$\bullet$ \textit{Randomized Choice.} To ensure \textit{uncertainty}, we select an example $x_i$ with a probability of $p_i$. Here, the value of $p_i$ can be used as a threshold for entropy to meet the uncertainty criterion (mentioned in the previous subsection) or can simply be a value to control the selection rate of examples. 

Note that none of these above heuristics require the knowledge of the complete training set. These are applicable to unsupervised and semi-supervised learners as they do not require the class labels. The effectiveness of these heuristics largely depends on the nature of the online learning problem. A comparison of these is presented in the evaluation section.   

%% file: tex/6.IMPLEMENTATION.tex
\section{Application Implementation}

We implement three intermittent learning applications that monitor, learn, and classify air quality indices, human presence, and a vibration pattern. These systems are powered by solar, RF, and piezoelectric harvesters, respectively. To demonstrate the portability of the proposed framework, these systems are implemented on three different microcontroller platforms, i.e., an AVR, a PIC, and an MSP430-based microcontroller, respectively. This section describes the implementation of these systems along with their end-to-end classification performance, deferring the in-depth evaluation to Section~\ref{sec:evaluation}.

\subsection{Air Quality Learning (Solar)}
\begin{figure} [tb]
    \centering
    \subfloat[\textit{Custom learning platform PCB}]
    {
        \includegraphics[width=0.30\textwidth]{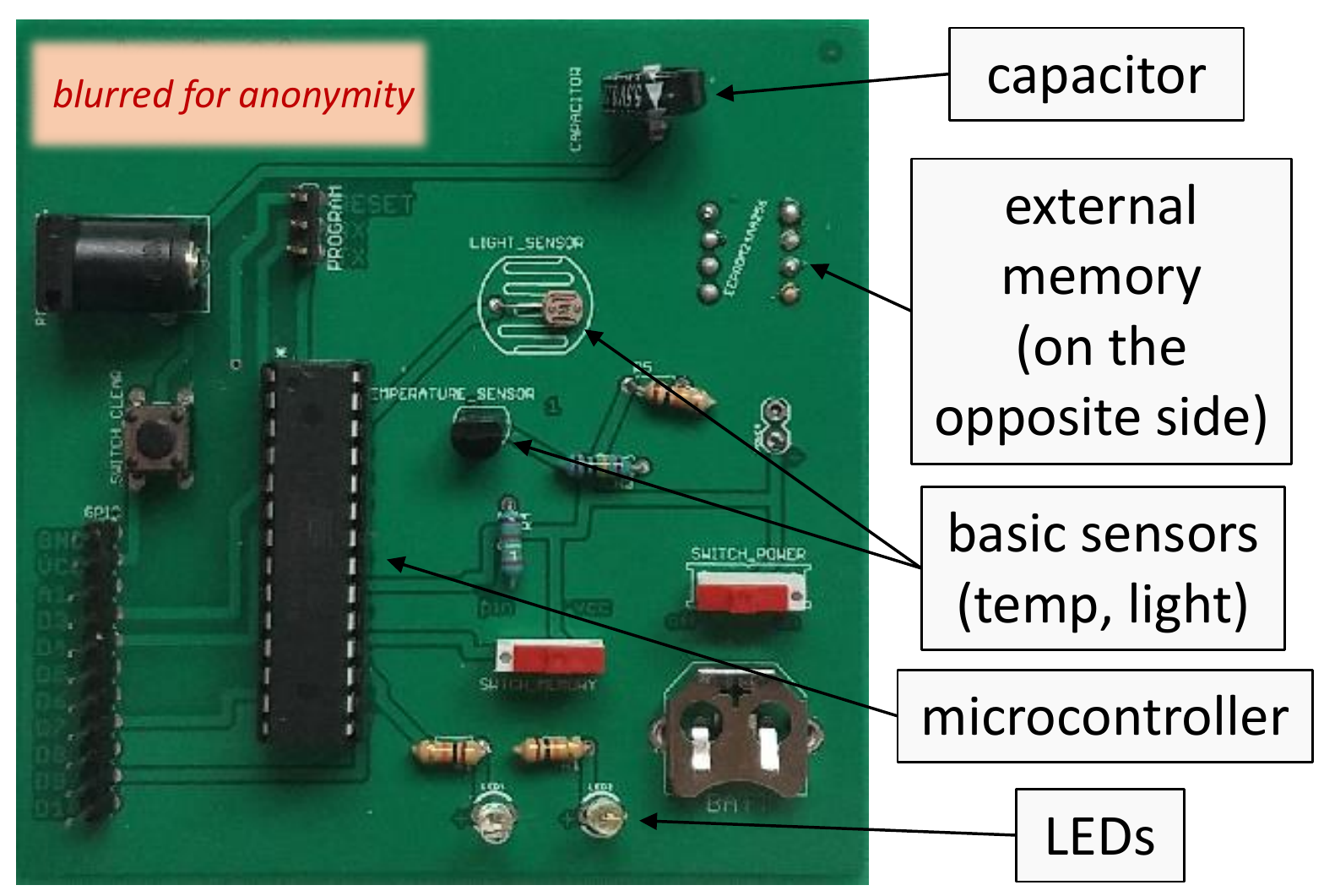}
        \label{fig:lesularduino}
    }
    \subfloat[Air quality learning system] {
        \includegraphics[width=0.27\textwidth]{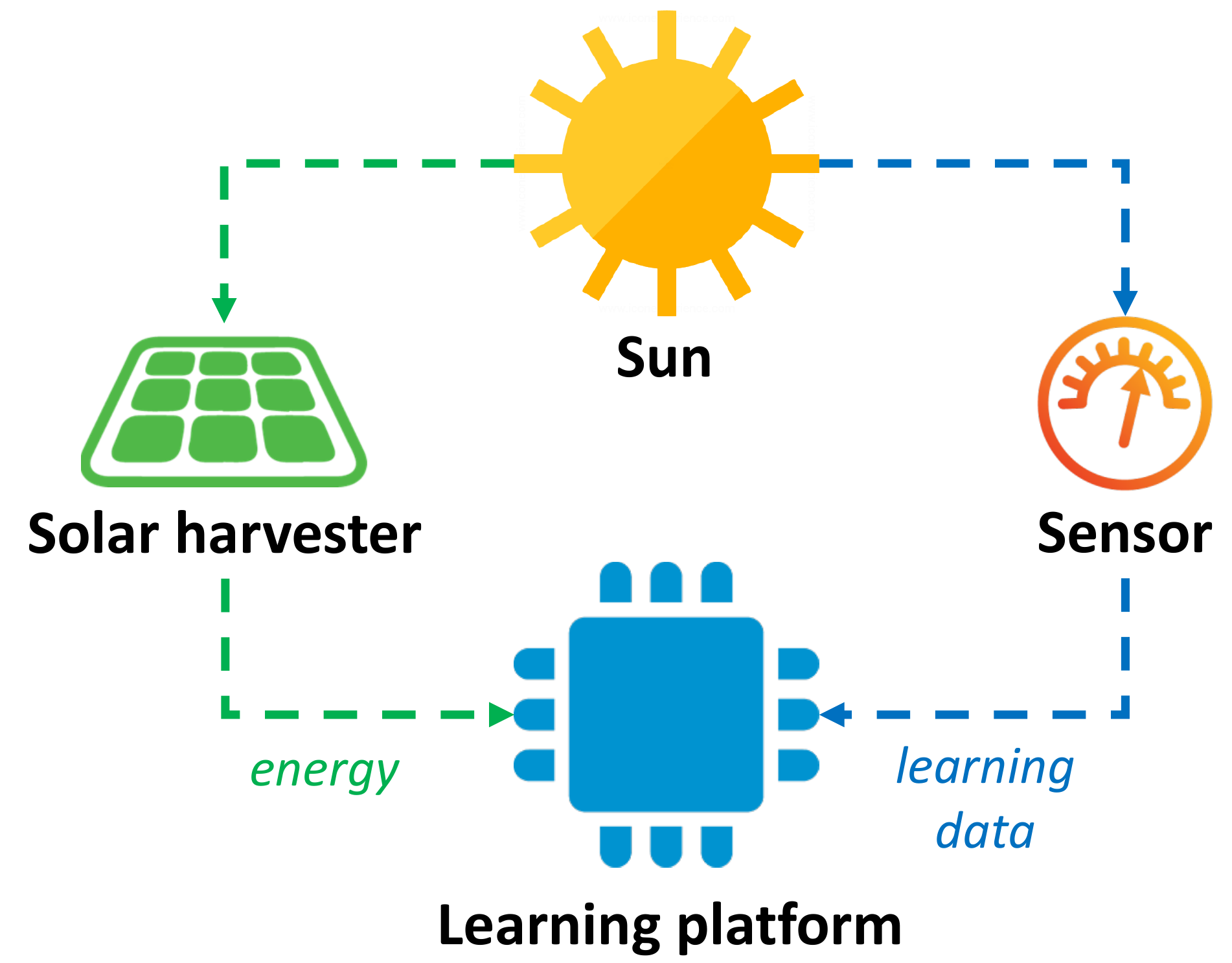}
        \label{fig:air_quality_learner}
    }\hspace{3mm}%
    \subfloat[Detection accuracy]
    {
        \includegraphics[width=0.28\textwidth]{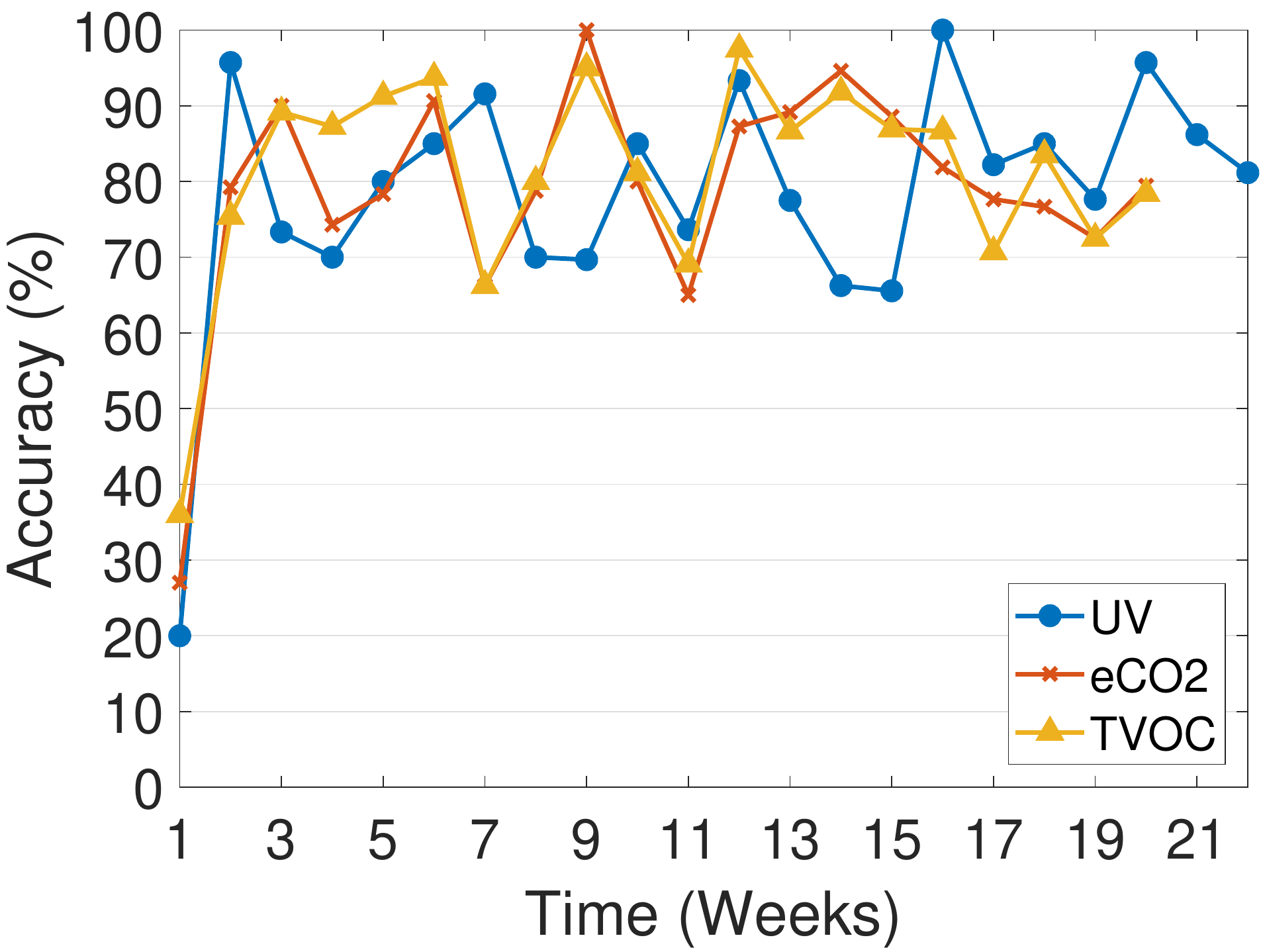}
        \label{fig:air_quality_performance}
    }
    \caption{Air quality learning system uses a custom-built platform and is powered by solar energy.}
    \label{fig:air_system}
\end{figure}

$\bullet$ \textit{Overview.}
The air quality learning system detects and notifies abnormalities in air quality indices such as the \emph{ultraviolet radiation} (UV), \emph{equivalent carbon dioxide} (eCO2), and \emph{total volatile organic compound} (TVOC) by learning their normal levels on harvested solar energy. Unlike sensing systems that just report the absolute sensing values, it learns the evolving status of air quality and provides environmental context-based notifications, which is smarter than reporting simple index values.

The system has been deployed in the real-world (near a window of an apartment), and it is active since September 21, 2018. We have an anonymous website showing the real-time status of the learner, which is updated every 10 minutes\textsuperscript{\ref{footnote 1}}. For the demonstration purpose, we use an additional gateway device that reads the classification results from the batteryless learner and sends them to the web.

$\bullet$ \textit{System.} As the experimental platform, we develop a custom printed circuit board (PCB) which is shown in Figure~\ref{fig:air_system}(a). The board consists of an ATmega328p microcontroller having a 1KB internal EEPROM, light and temperature sensors, a 32KB external non-volatile EEPROM, a 0.2F supercapacitor as the energy reservoir, output indicator LEDs, and energy harvester circuitry. Although more advanced energy management hardware such as multiple capacitors~\cite{colin2018reconfigurable, hester2017flicker, hester2015hardware} can be used for more efficient use of harvested energy, we keep our hardware design simple to focus on the feasibility, behavior, and performance of the learning framework. The air-quality sensors measuring UV, eCO2, and TVOC are externally connected to the PCB (not shown in the figure). The board harvests solar energy and executes machine learning algorithms following the proposed intermittent learning framework. As shown in Figure~\ref{fig:air_system}(b), the air quality learning system utilizes the custom PCB as the learning platform and a small solar panel for energy harvesting. When the sunlight is available, the solar panel charges the supercapacitor and powers up the circuitry to wake up the learner. Upon wake up, the system collects data from sensors and executes the learning actions. Note that although the sunlight is present for the most of the day, as the system is powered through a limited sized capacitor that drains quickly when the system runs, the input power to the system is intermittent, and thus requiring the framework to save/restore the intermediate system states into/from the non-volatile memory.

$\bullet$ \textit{Learning Algorithm.} 
The $k$-nearest neighbor algorithm is used to learn and detect an anomaly in the ambient air quality. We choose the $k$-nearest neighbor algorithm for clustering among other alternatives such as \emph{autoencoders} since the application does not deal with high dimensional data, and the carefully-designed features (described next) are more compute- and energy-efficient than autoencoders. Following the proposed framework, we implement the \textit{sense} action that reads three sensor values (UV, eCO2, and TVOC) every 32 seconds. For every 60 sensor readings, the \textit{extract} action generates five features-- mean, standard deviation, median, root mean square (RMS), and peak-to-peak amplitude (P2P). The five features generated by the \textit{extract} action constitute an example which is used for learning (i.e., the \textit{learn} action) or detecting an anomaly (i.e., the \textit{infer} action).

Prior to learning, the \textit{select} action determines whether the newly-obtained example should be learned or discarded using the example selection heuristic. If the example is selected for learning, the \textit{learn} action updates the threshold score for anomaly detection by learning the latest set of examples, including the newly-obtained one. The anomaly score $AS_i$ for the $i^{th}$ example $e_i$ in an example set is calculated as $AS_i=\sum_{j=1}^{k}d(e_i,e_j)$, where $e_j$ is the $j^{th}$ nearest neighbor example of $e_i$, $k$ is the number of nearest neighbors in the set, and $d(\cdot)$ is the feature distance function~\cite{cola2015node}. The feature distance between two examples $e_i$ and $e_j$ is defined as $d(e_i,e_j)=\sqrt{\sum_{m=1}^{n}(f^{e_i}_{m}-f^{e_j}_{m})^2}$, where $f^{e_i}_{m}$ is the $m^{th}$ feature of the example $e_i$, $f^{e_j}_{m}$ is the $m^{th}$ feature of the example $e_j$, and $n$ is dimension of the feature vector. After computing the anomaly score for all examples in the set, an anomaly threshold $AS_{TH}$ is determined by taking the 90th percentile of the anomaly score.

To detect an anomaly (i.e., the \textit{infer} action), the system calculates the anomaly score $AS_{new}$ for the newly-obtained example. It is classified as abnormal, if $AS_{new} > AS_{TH}$, and normal, otherwise. Note that the anomaly threshold $AS_{TH}$ evolves over time as new examples are learned at run-time.

Figure \ref{fig:air_system}(c) shows the anomaly detection accuracy of the system for the three indicators, i.e., UV, eCO2, TVOC for 20 weeks. The anomalies are detected with 81\%--83\% average accuracy for the air quality indicators. To calculate the accuracy of the learners, we download the classification results as well as the raw data from the device once every week. The raw data is visualized and inspected by human experts to obtain the ground truth labeling, which is compared with the classification results of the learner to calculate the accuracy.  

\subsection{Mobile Human Presence Learning (RF)}
\begin{figure} [tb]
    \centering
    \subfloat[RF energy learning platform]
    {
        \includegraphics[width=0.22\textwidth]{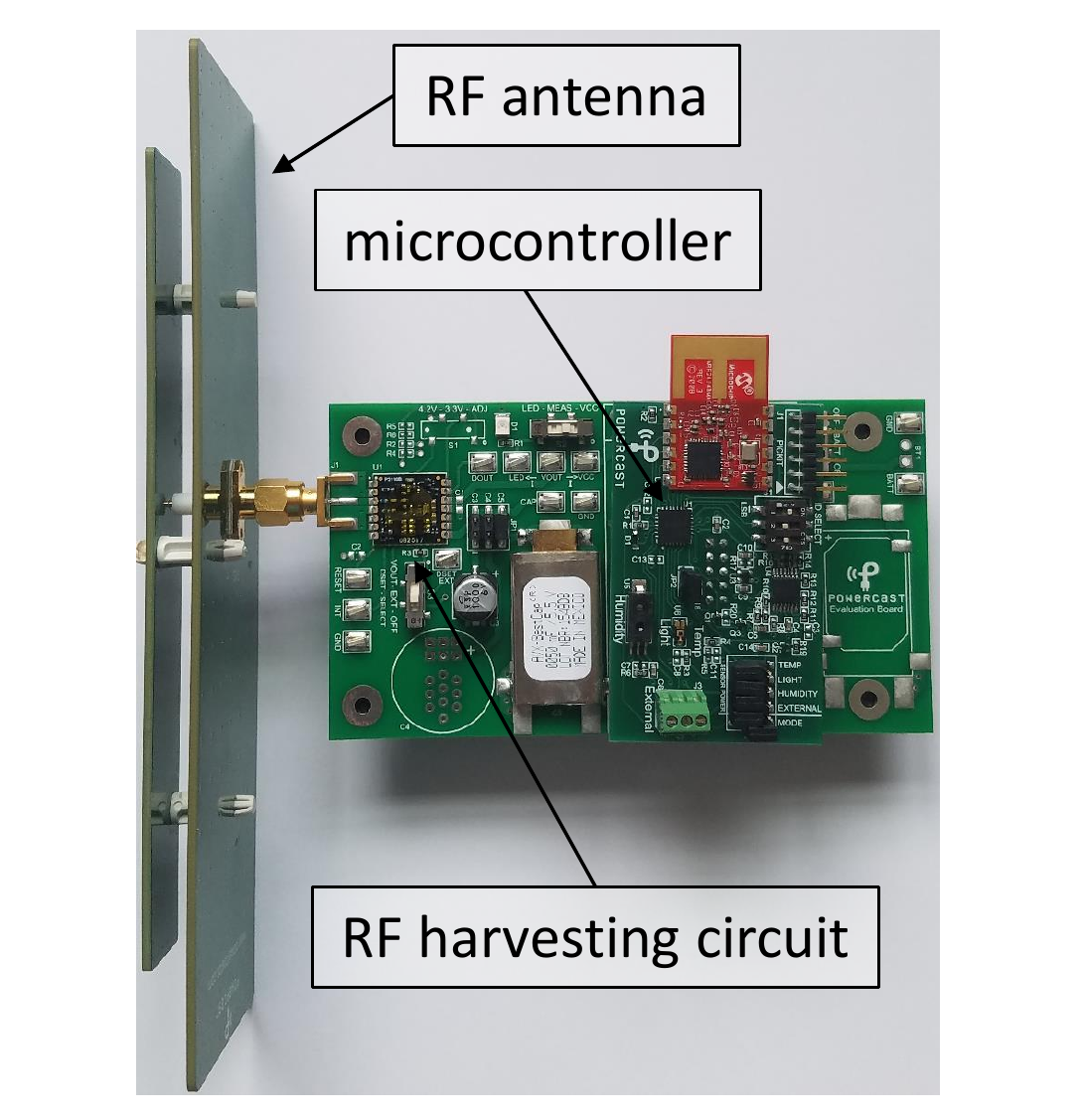}
        \label{fig:rf_kit}
    }
    \subfloat[Human presence learning system] 
    {
        \includegraphics[width=0.33\textwidth]{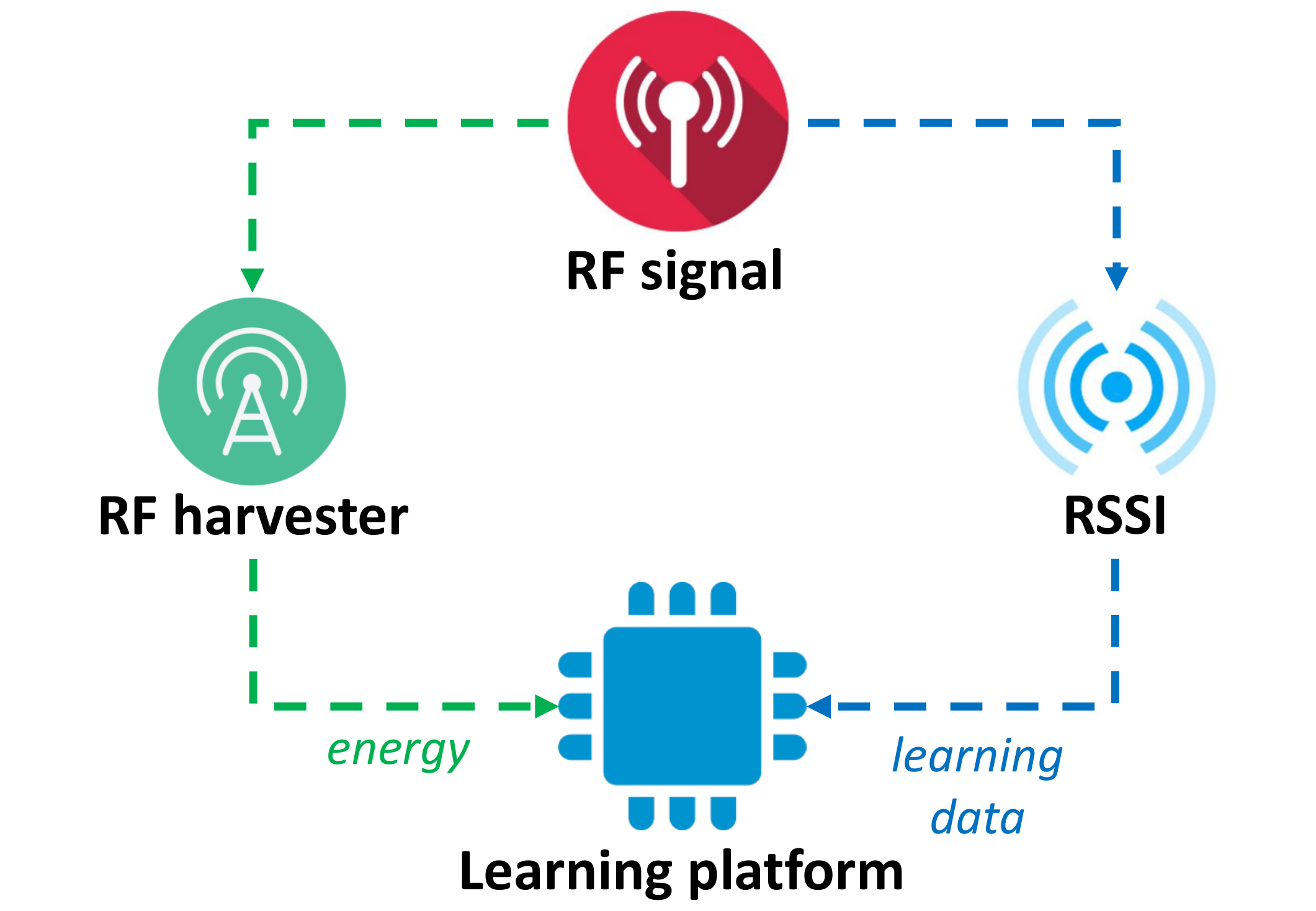}
        \label{fig:human_presence_learner}
    }
    \subfloat[Detection accuracy] 
    {
        \includegraphics[width=0.31\textwidth]{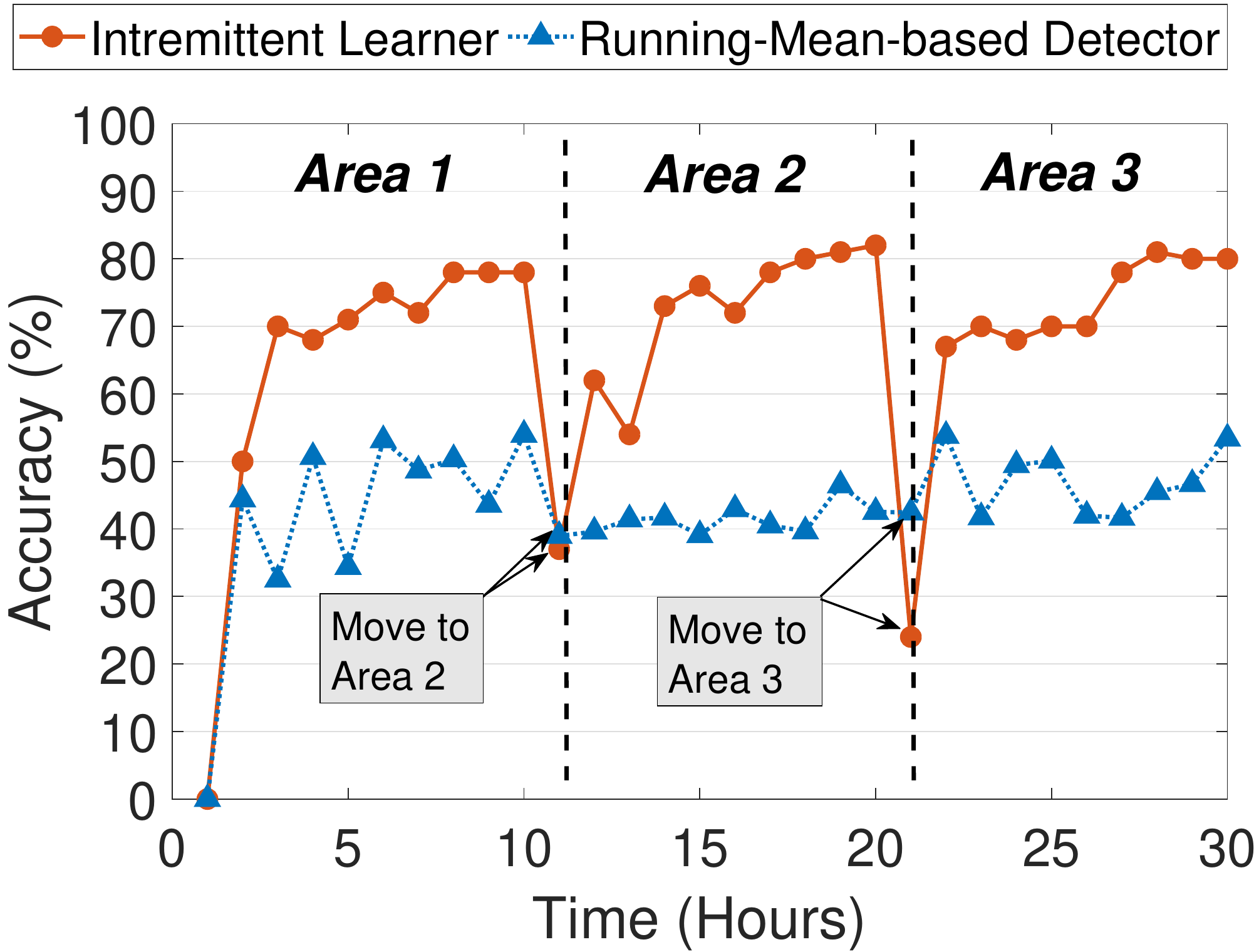}
        \label{fig:human_presence_performance}
    }
    \caption{Mobile human presence learning on RF energy.}
    \label{fig:human_system}
\end{figure}

$\bullet$ \textit{Overview.} We implement a mobile human presence learning system that is powered by harvesting RF energy. It detects the presence of a person in indoor space by observing the short-term variation in the received signal strength indicator (RSSI) values and by learning a dynamic threshold that helps it determine if a person is present or not. This is different from an RSSI threshold-based human presence detection system that does not generalize across different physical world environments or when the RF properties of the same environment change. Using the proposed intermittent learning framework, the human presence learner continuously learns the RF pattern, and thus it is able to learn and adapt its model parameters to accurately detect the presence of humans---even when the system is moved from one place to another. Using this learner, a mobile social robot~\cite{lemaignan2017artificial} can perceive the presence of humans when other types of sensors are ineffective (e.g., cameras in the dark).   

$\bullet$ \textit{System.} The system consists of three major parts that are shown in Figure~\ref{fig:human_system}(a) -- an RF antenna (850-950 MHz)~\cite{powercast1}, an RF harvesting circuit (P2110)~\cite{powercast2} and a PIC24F16KA102 microcontroller. Additionally, a 50mF capacitor and a 512-byte EEPROM (built-in the microcontroller) are used as the energy reservoir and non-volatile data storage, respectively. Figure~\ref{fig:human_system}(b) shows that both energy and data come from the RF signal. When the capacitor is charged by harvesting energy from the RF power source, the system starts to measure RSSI and learns to detect human presence or absence. The learning examples consist of the features obtained from RSSI values, and the learning model is saved in the non-volatile memory so that when the power goes off, the system does not lose its state. 

$\bullet$ \textit{Learning Algorithm.}
Similar to the air quality learning system, a $k$-nearest neighbor learner is used for anomaly detection. First, the RSSI power levels received at the antenna (ranging from 0.04mW to 50mW) are measured and calculated by the \textit{sense} action to collect a set of 10 to 30 values. The number and rate of RSSI readings constituting the set depends on the strength and the power of the signal. Four features (i.e., mean, standard deviation, median, and root mean square (RMS) of RSSI values) are extracted by the \textit{extract} action from a set of RSSI values. The extracted features constitute an example that is used either for learning (\textit{learn}) or for human presence detection (\textit{infer}) as dictated by the dynamic action planner. Since the learning and inferring algorithms in this application are similar to the air quality learning system, their details are omitted. The main difference between these two systems is that the human presence learner learns and updates its model more frequently and more intermittently (between tens of milliseconds and seconds) than the air quality learner (between minutes and hours) since RF signals change much faster than air quality sensor values.

In order to evaluate the performance of the system and its ability to adapt to a new environment, we deploy and measure its accuracy at three different areas by moving it from one place to another. The accuracy is compared against a baseline system that uses a threshold changing over time based on the run-time mean of the RSSI values to detect human presence. Figure~\ref{fig:human_system}(c) shows the accuracy of the system at three different locations as the system is moved. The accuracy is tested every hour using 30 test cases of human presence and absence. As shown in the figure, when the intermittent learning system is moved to a new area, it recovers its detection accuracy within a few hours by adapting its model parameters to the new RF environment which is very different from the previous one. For instance, the accuracy drops to 38\% at hour 11 after moving to area 2, but it rises back to 76\% at hour 15 and increases to 82\% at hour 20. The baseline system's accuracy stays below 50\% for all areas.

$\bullet$ \textit{Overview.} The vibration of machines such as industrial machinery, HVAC equipment, vehicles, and household appliances carries the signature of their state of operation and health status. By observing and learning their regular vibration pattern, we can predict their impending failure when there is a deviation or irregularity in their vibration pattern. Vibration anomaly detection systems can also be used in human health and wellness applications. For example, a gait anomaly detector can give a warning sign of walking abnormalities such as the freezing of gait~\cite{giladi1998freezing} or a sudden fall by learning and classifying a user's walking pattern. Early detection of Parkinson's disease is possible by noticing tremors (hand or foot shaking)~\cite{zimmermann1994tremors}, and detecting leg shaking (SPINDLES)~\cite{xia2017spindles} are examples of people-centric vibration sensing and inference application. 

We develop a vibration learning system that is powered by harvesting piezoelectric energy. It detects a potential malfunction of a vibrating object or a human limb by monitoring and learning the regular vibration pattern using an accelerometer sensor, and then detects and reports anomalies. The system is shown in Figure~\ref{fig:vibration_system}(b). The system can be attached to a target to learn the level of vibration that may relate to an impending breakdown or an anomaly.

$\bullet$ \textit{System.} As shown in Figure~\ref{fig:vibration_system}(a), a piezoelectric harvester (PPA-2014)~\cite{PPA}, generating power between 1.8mW and 36.5mW, is connected to an MSP430FR5994 microcontroller via a piezoelectric harvesting circuit (LTC3588). A 6mF capacitor stores the harvested energy. We use the microcontroller's built-in 256KB FRAM as the non-volatile storage to save the system state. A low-power accelerometer sensor (LIS3DH) attached to the tip of the piezoelectric harvester senses the three-dimensional vibration at the sampling rate of 50Hz. 

\subsection{Vibration Learning (Piezoelectric)}
\label{sec:vibration_learning}
\begin{figure} [tb]
    \centering
    \subfloat[Piezoelectric energy learning platform] 
    {
        \includegraphics[width=0.29\textwidth]{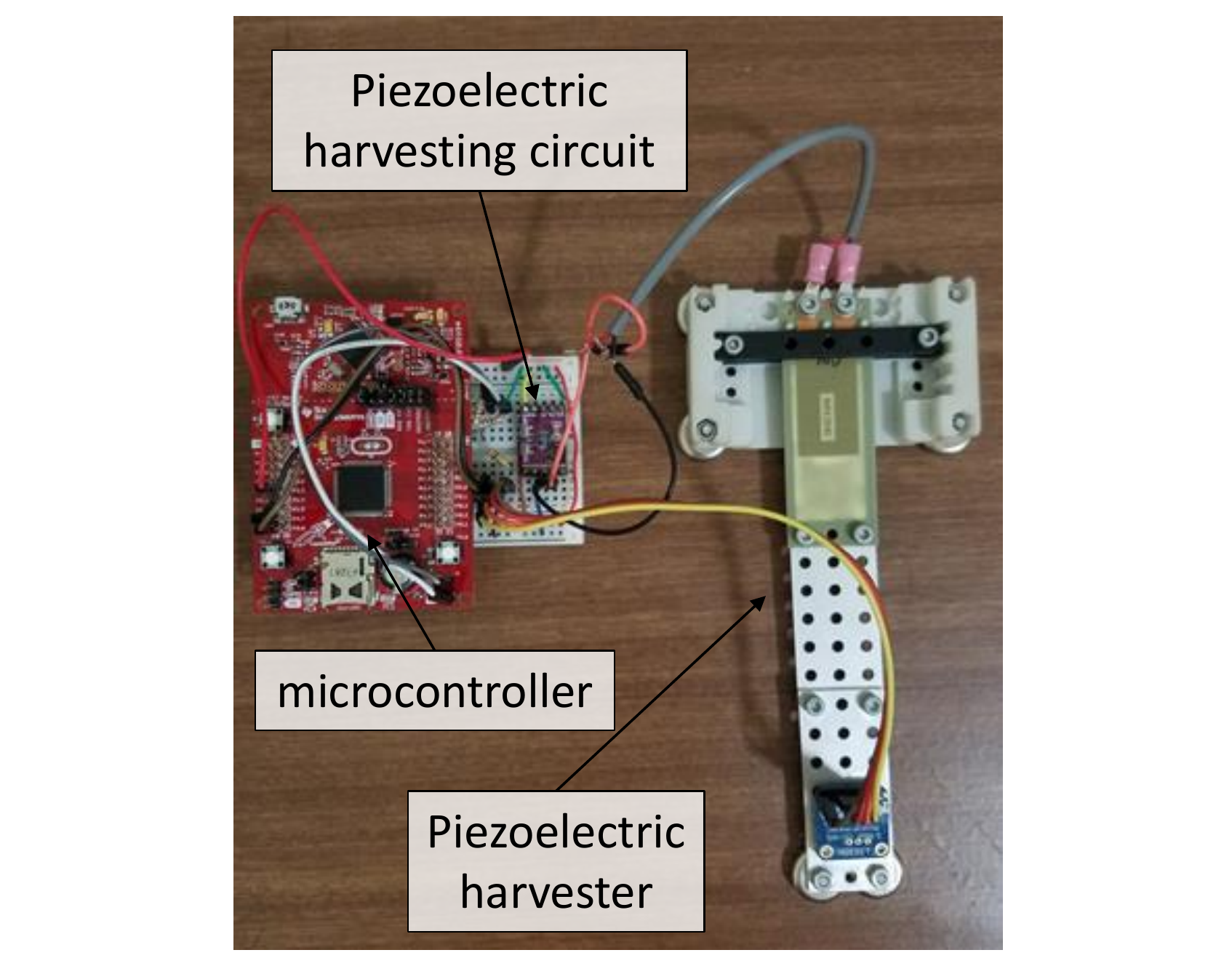}
        \label{fig:piezo_kit}
    } \hspace{2mm}%
    \subfloat[Vibration learning system] 
    {
        \includegraphics[width=0.23\textwidth]{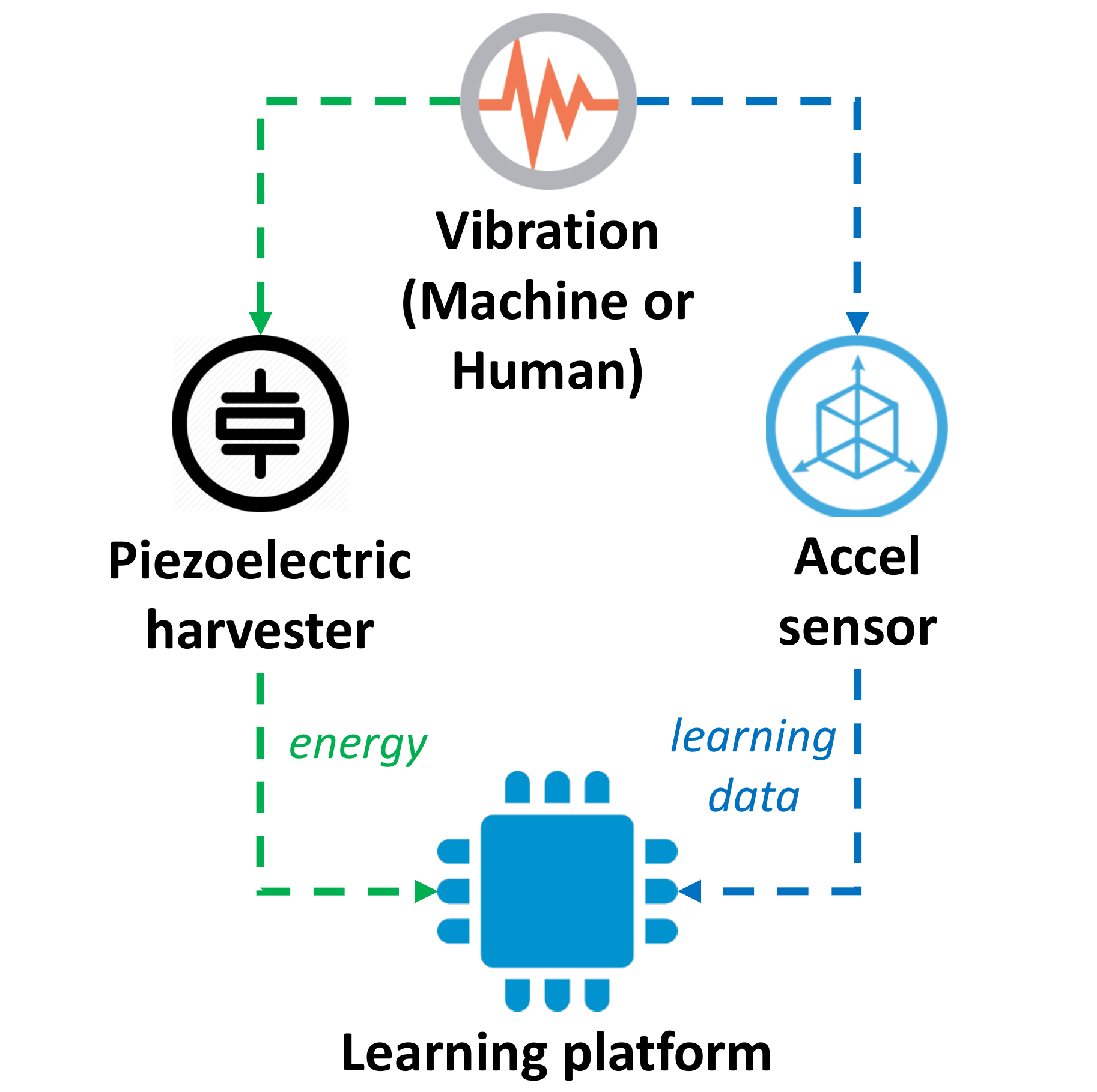}
        \label{fig:vibration_learner}
    } \hspace{2mm}%
    \subfloat[Detection accuracy] {
        \includegraphics[width=0.31\textwidth]{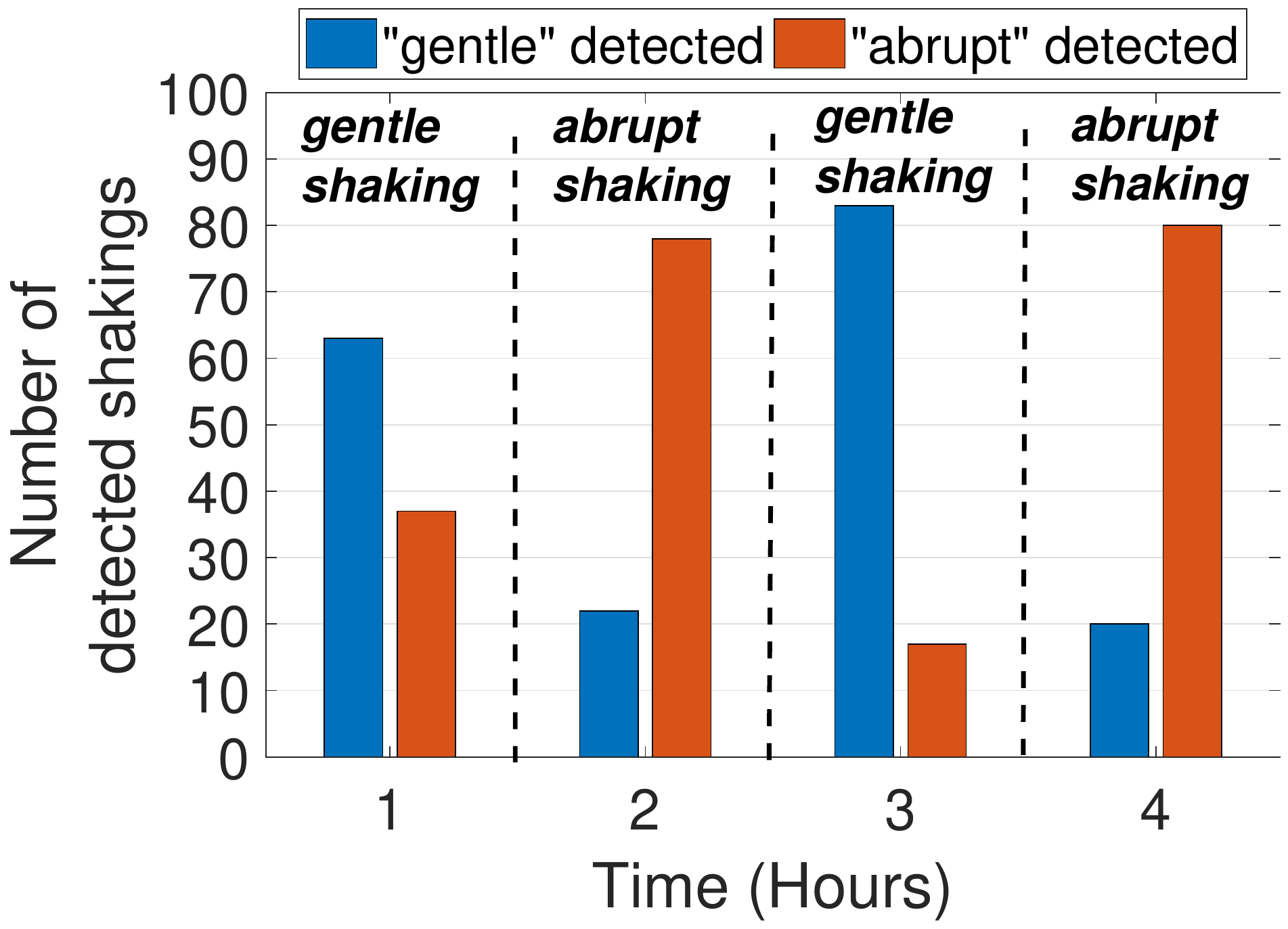}
        \label{fig:performance_vibration}
    }
    \caption{Vibration learning on piezoelectric energy.}
    \label{fig:vibration_system}
\end{figure}

$\bullet$ \textit{Learning Algorithm.} We implement a \emph{cluster-then-label}~\cite{goldberg2010new,zhu2005semi} learner that utilizes both labeled and unlabeled data where the training examples first go through a clustering step, and then the clusters are labeled. The learner classifies new examples by finding the cluster it belongs to and then uses the label of the cluster to classify the example. This approach falls under the general category of semi-supervised learning but is different from alternatives such as label propagation~\cite{xiaojin2002learning}. 

For clustering, we implement a two-layer neural network-based $k$-means algorithm~\cite{marsland2015machine} where the input and output layers correspond to the feature vector of an example and the two clusters (normal and abnormal vibration), respectively. 
Unlike typical $k$-means algorithms that have all examples in its batch learning setup, only one example (at a time) is available to our online learner, and the cluster means are unknown. Hence, we feed one example to the neural network at a time and approximate the cluster means by moving the neuron closer to the current input example---making that center even more likely to be the best match next time that input is seen.

The \textit{learn} action implements the clustering algorithm which uses feature vectors extracted by the \textit{extract} action consisting of the mean, standard deviation, median, root mean square (RMS), peak-to-peak amplitude (P2P), zero-crossing rate (ZCR), and average absolute acceleration variations (AAV). Two output neurons corresponding to the two clusters (normal and abnormal vibration) are fully connected to the input layer neurons. An activation value, $a_{j}$ for each neuron is calculated by $a_{j}=\sum_{i=1}^{n}w_{ij}x_i$, where $w_{ij}$ is the weight between the $i$th element of the input vector and the $j$th neuron, $x_{i}$ is the $i$th element of the input vector, and $n$ is the length of the input vector, $\mathbf{x}$. We implement competitive learning where only the neuron with the largest activation value wins and only the weights connected to the winner are updated at each iteration since the winner neuron corresponds to the cluster that is the closest to the current input. The weights of the winner neuron, $w_{ij}$ are updated by $\Delta w_{ij}=\eta(x_i - w_{ij})$, where $\eta$ is the learning rate. To classify new data (i.e., \textit{infer} action), features of new example are extracted and fed into the neural network as the input. The output neuron with the highest activation value is chosen as the predicted class. 

We conduct a set of controlled experiments with the vibration anomaly detector. We attach the system to an arm of a person and let the system learn to cluster the arm shaking into two categories: gentle vs. abrupt shaking. Gentle and abrupt arm movements are performed by shaking the arm less than five times and more than ten times in five seconds, respectively. Figure~\ref{fig:vibration_system}(c) shows the classification accuracy for four hours of the experiment. 100 gentle shaking gestures are performed during the first and the third hour, while 100 abrupt shaking gestures are performed during the second and the fourth hour. As shown in the figure, the system learns and classifies the two movements with 76\% average accuracy using the kinetic energy generated by the arm shaking gestures.

%% file: tex/7.EVALUATION.tex
\section{Evaluation}
\label{sec:evaluation}

We conduct in-depth experiments to evaluate various aspects of the three applications described in the previous section. First, their performance is compared with 1) state-of-the-art intermittent computing systems that execute learning and inference steps periodically, and implements neither the dynamic action planner nor the example selection heuristics (Section~\ref{subsec:intermittent}), and 2) three popular offline machine learning algorithms for anomaly detection (Section~\ref{subsec:offline}). Second, we evaluate the effect of example selection heuristics (Section~\ref{subsec:heuristics}) and energy harvesting patterns (Section~\ref{subsec:harvesting_pattern}) on the performance of the learner. Third, we measure the energy consumption and execution time of each action and quantify the overhead of the system (Section~\ref{subsec:energy_consumption_and_time}).

\subsection{Comparison with the State-of-the-Art Intermittent Computing Systems}
\label{subsec:intermittent}

\begin{figure} [t]
    \centering
    \subfloat[Air quality 1 (UV)] 
    {
        \includegraphics[width=0.32\textwidth]{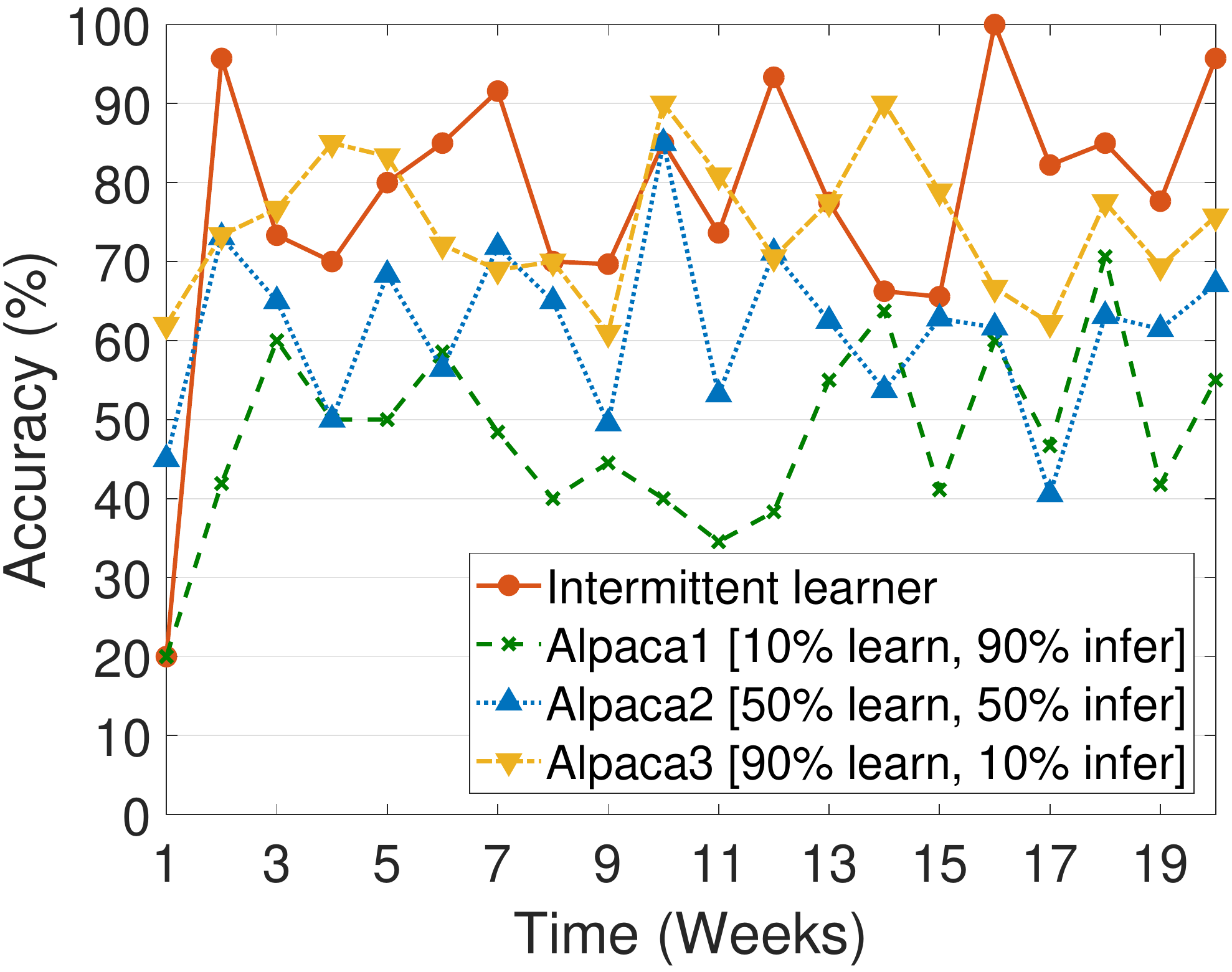}
        \label{fig:uv_accuracy_duty_alpaca}
    }
    \subfloat[Air quality 2 (eCO2)] 
    {
        \includegraphics[width=0.32\textwidth]{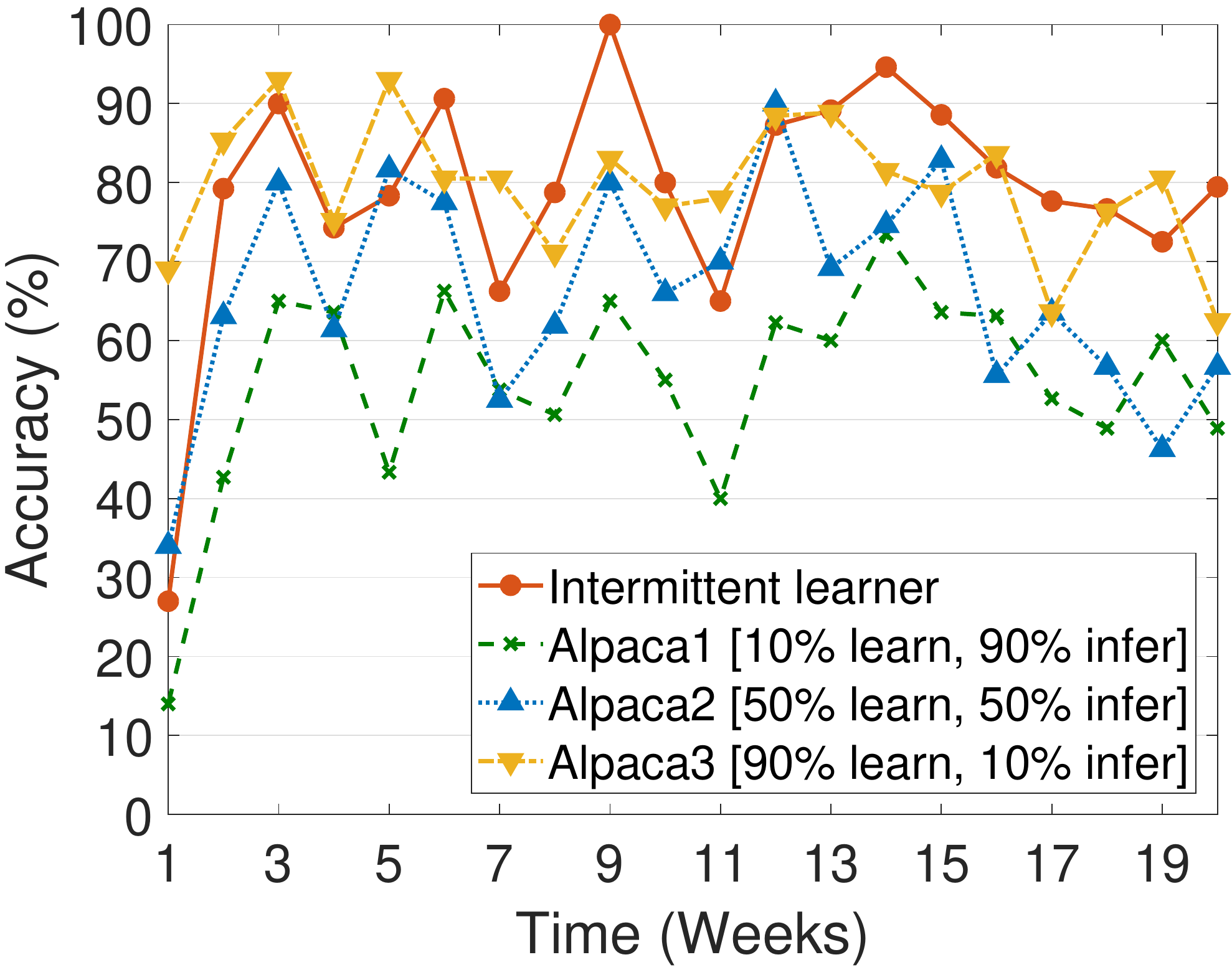}
        \label{fig:eco2_accuracy_duty_alpaca}
    }
    \subfloat[Air quality 3 (TVOC)] 
    {
        \includegraphics[width=0.32\textwidth]{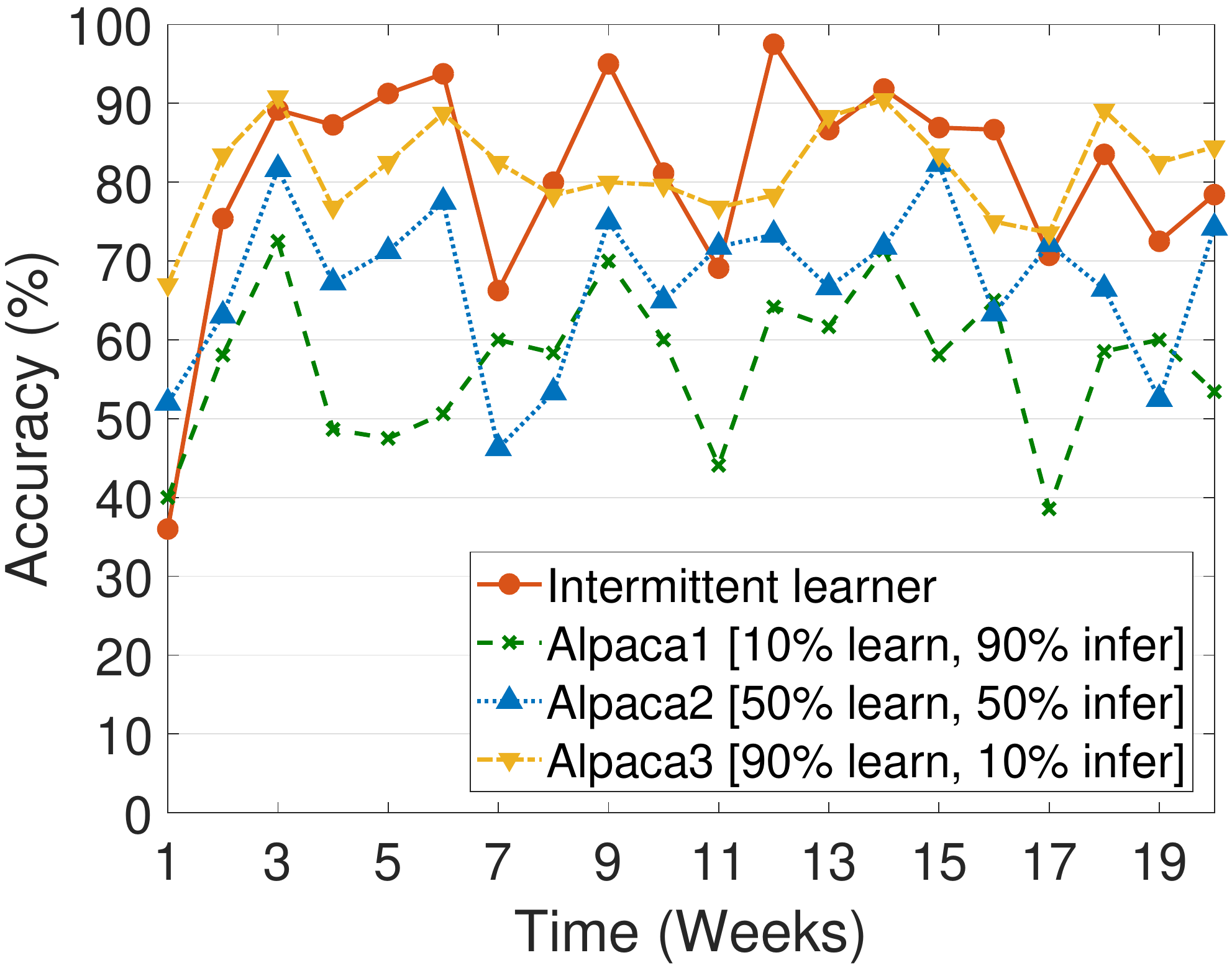}
        \label{fig:tvoc_accuracy_duty_alpaca}
    }
    \\
\begin{minipage}{0.64\textwidth}
    \subfloat[Human presence] 
    {
        \includegraphics[width=0.49\textwidth]{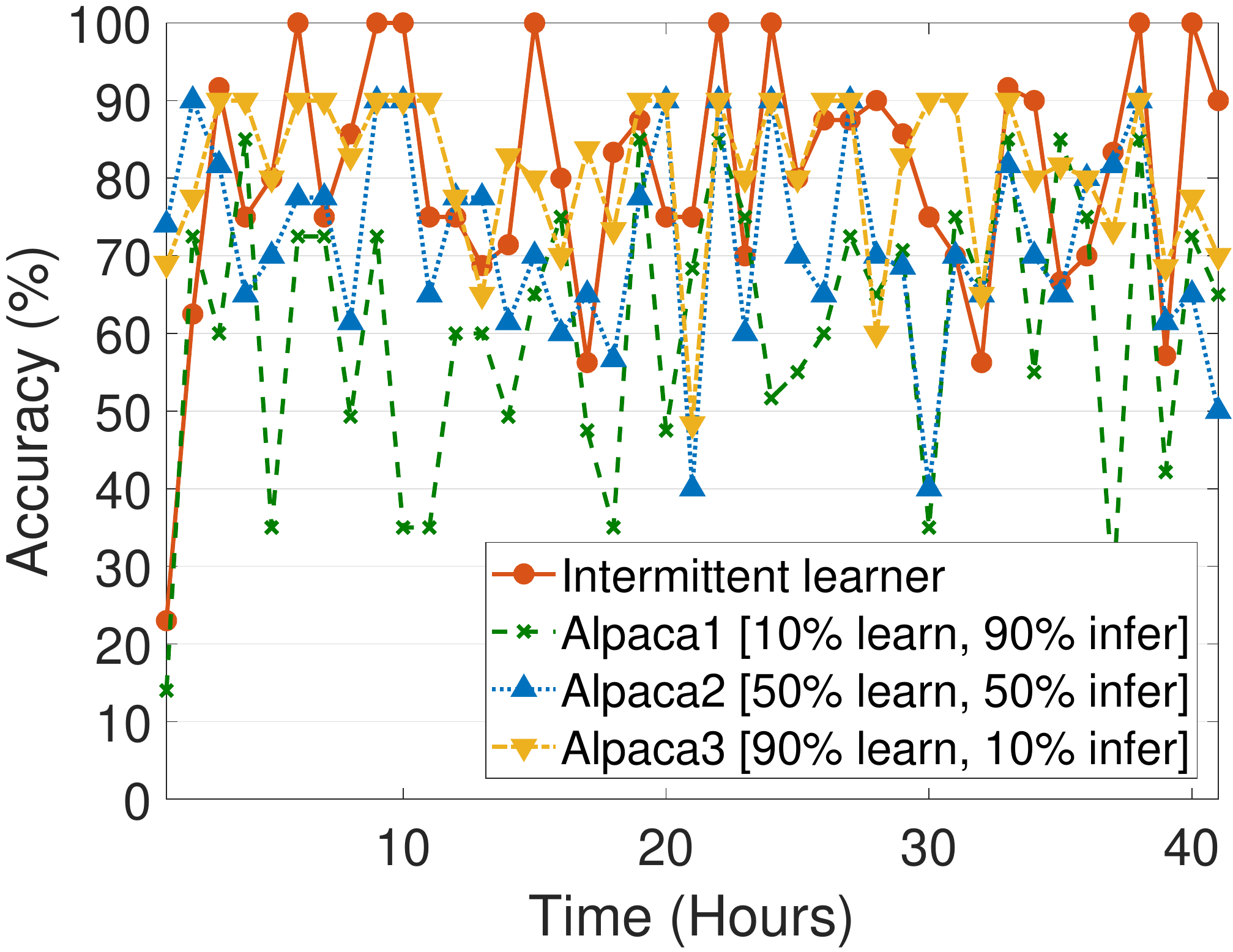}
        \label{fig:rssi_accuracy_duty_alpaca}
    }
    \subfloat[Vibration] 
    {
        \includegraphics[width=0.49\textwidth]{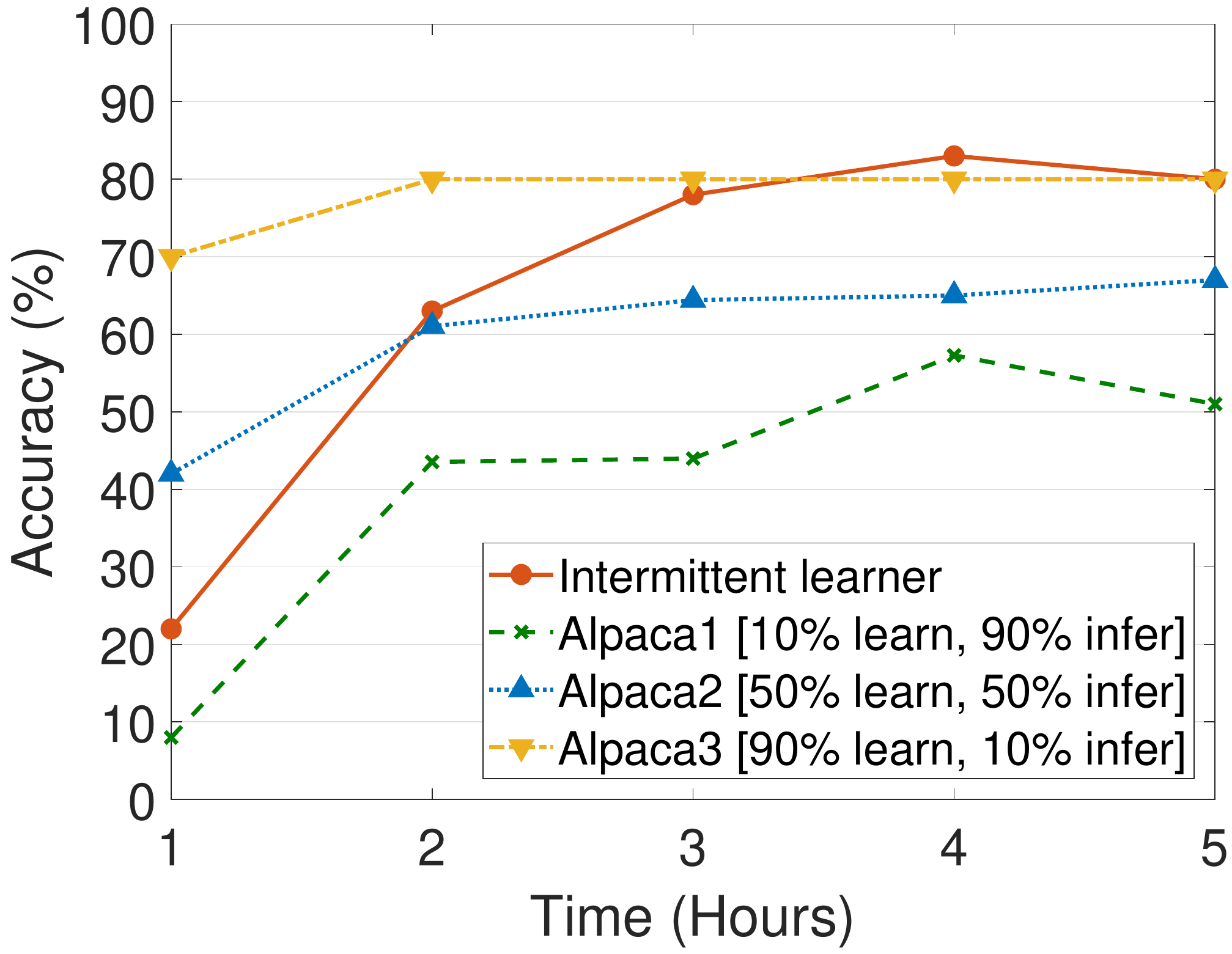}
        \label{fig:vibration_accuracy_alpaca}
    }
\end{minipage} \hspace{1mm}
\begin{minipage}{0.34\textwidth}
\captionsetup{type=table} 
\tiny
\vspace{-12pt}
\caption{Average detection accuracy (\%): Intermittent learner vs. Alpaca.}
\label{table:alpaca}
\begin{center}
\vspace{-10pt}
\begin{tabular}{rcccc}
    \toprule
    \textbf{} & \textbf{Inter.} & \textbf{Alpaca} & \textbf{Alpaca} & \textbf{Alpaca}\\
    \textbf{} & \textbf{Lean} & \textbf{Duty} & \textbf{Duty} & \textbf{Duty}\\
    \textbf{} & & \textbf{10/90} & \textbf{50/50} & \textbf{90/10} \\
    \toprule
    \textbf{\textit{UV}} & \textbf{81\%} & 48\% & 61\% & 74\% \\
    \textbf{\textit{eCO2}} & \textbf{81\%} & 54\% & 66\% & 79\% \\
    \textbf{\textit{TVOC}} & \textbf{83\%} & 57\% & 61\% & 81\% \\
    \hline
    \textbf{\textit{Human}} & \textbf{82\%} & 60\% & 71\% & 81\% \\
    \textbf{\textit{Presence}} & & & & \\
    \hline
    \textbf{\textit{Vibration}} & \textbf{76\%} & 40\% & 59\% & 78\% \\
    \toprule
\end{tabular}
\end{center}
\end{minipage}
\caption{Accuracy comparison with Alpaca (no dynamic action planner and example selection)}
\label{fig:alpaca}
\end{figure}

\begin{figure} [tb]
    \centering
    \subfloat[Air quality 1 (UV)] 
    {
        \includegraphics[width=0.32\textwidth]{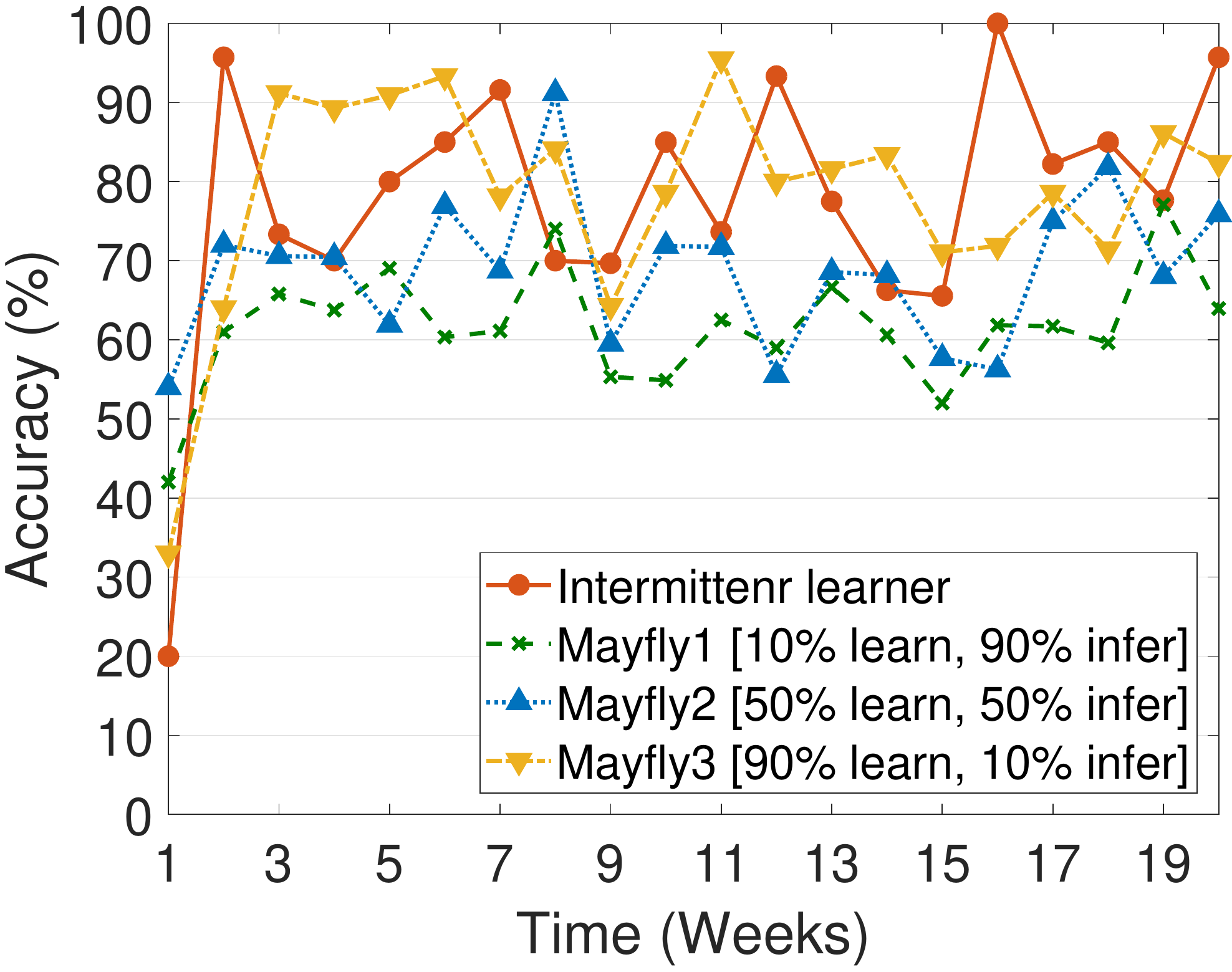}
        \label{fig:uv_accuracy_duty_mayfly}
    }
    \subfloat[Air quality 2 (eCO2)] 
    {
        \includegraphics[width=0.32\textwidth]{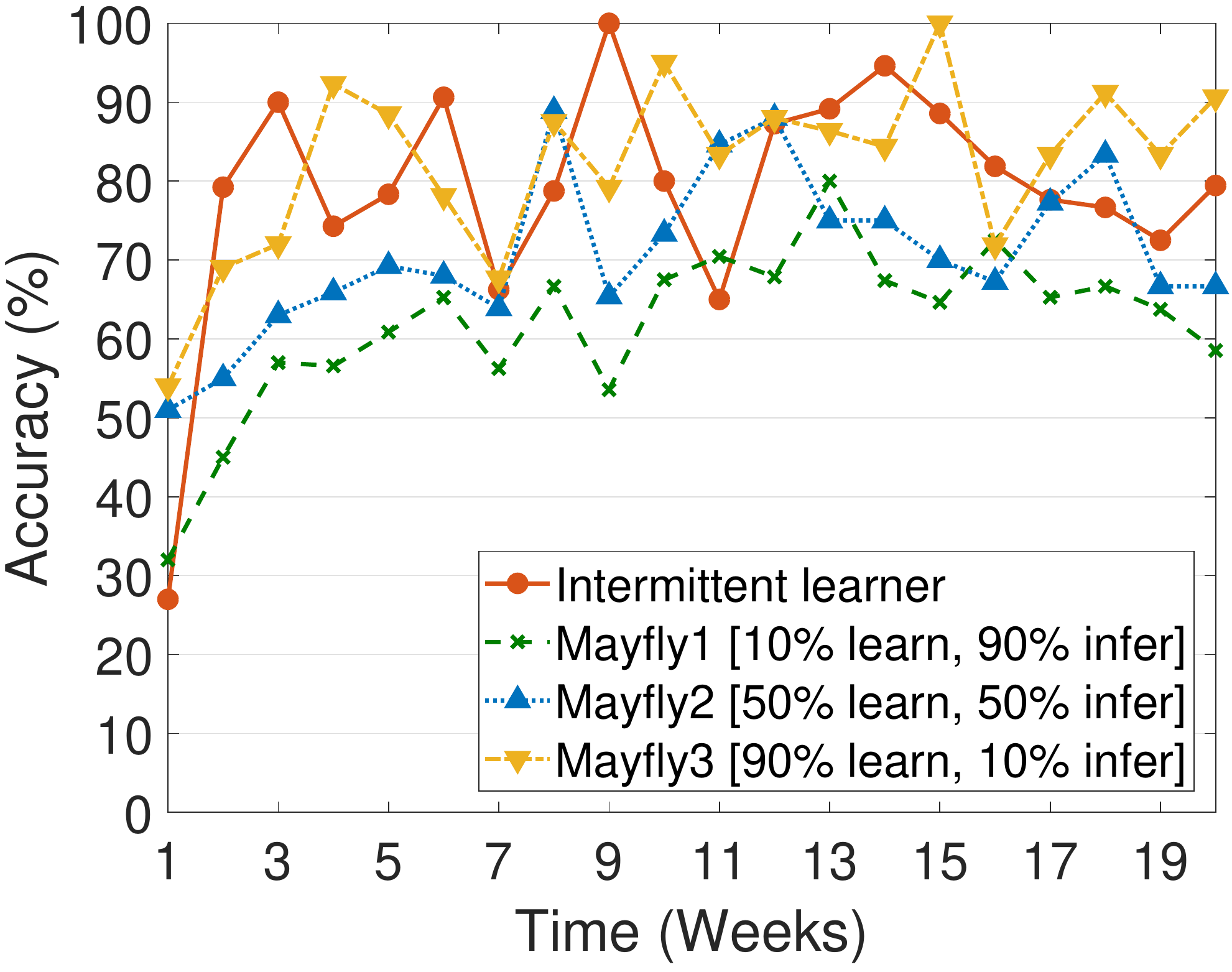}
        \label{fig:eco2_accuracy_duty_mayfly}
    }
    \subfloat[Air quality 3 (TVOC)] 
    {
        \includegraphics[width=0.32\textwidth]{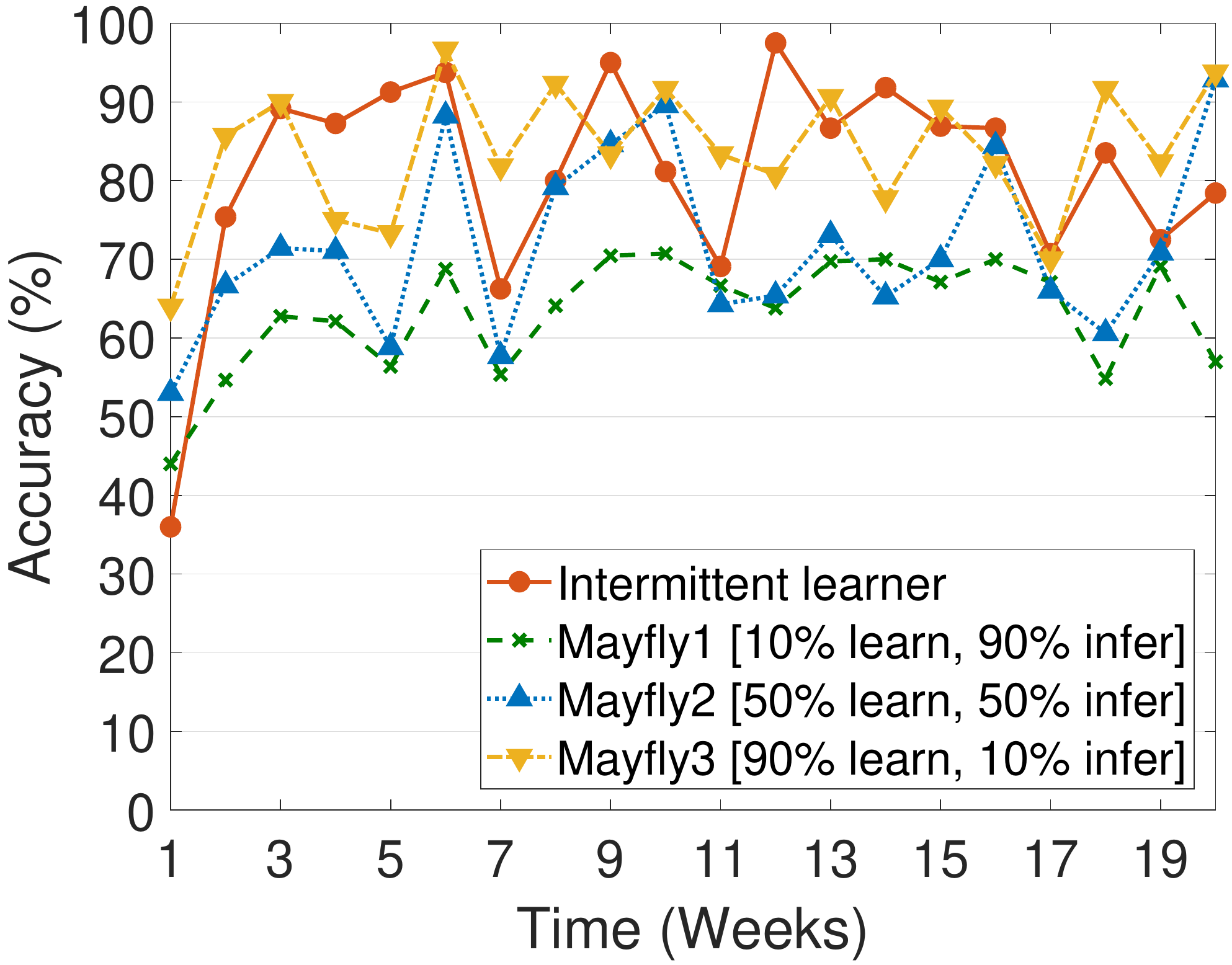}
        \label{fig:tvoc_accuracy_duty_mayfly}
    }
    \\
\begin{minipage}{0.64\textwidth}
    \subfloat[Human presence] 
    {
        \includegraphics[width=0.49\textwidth]{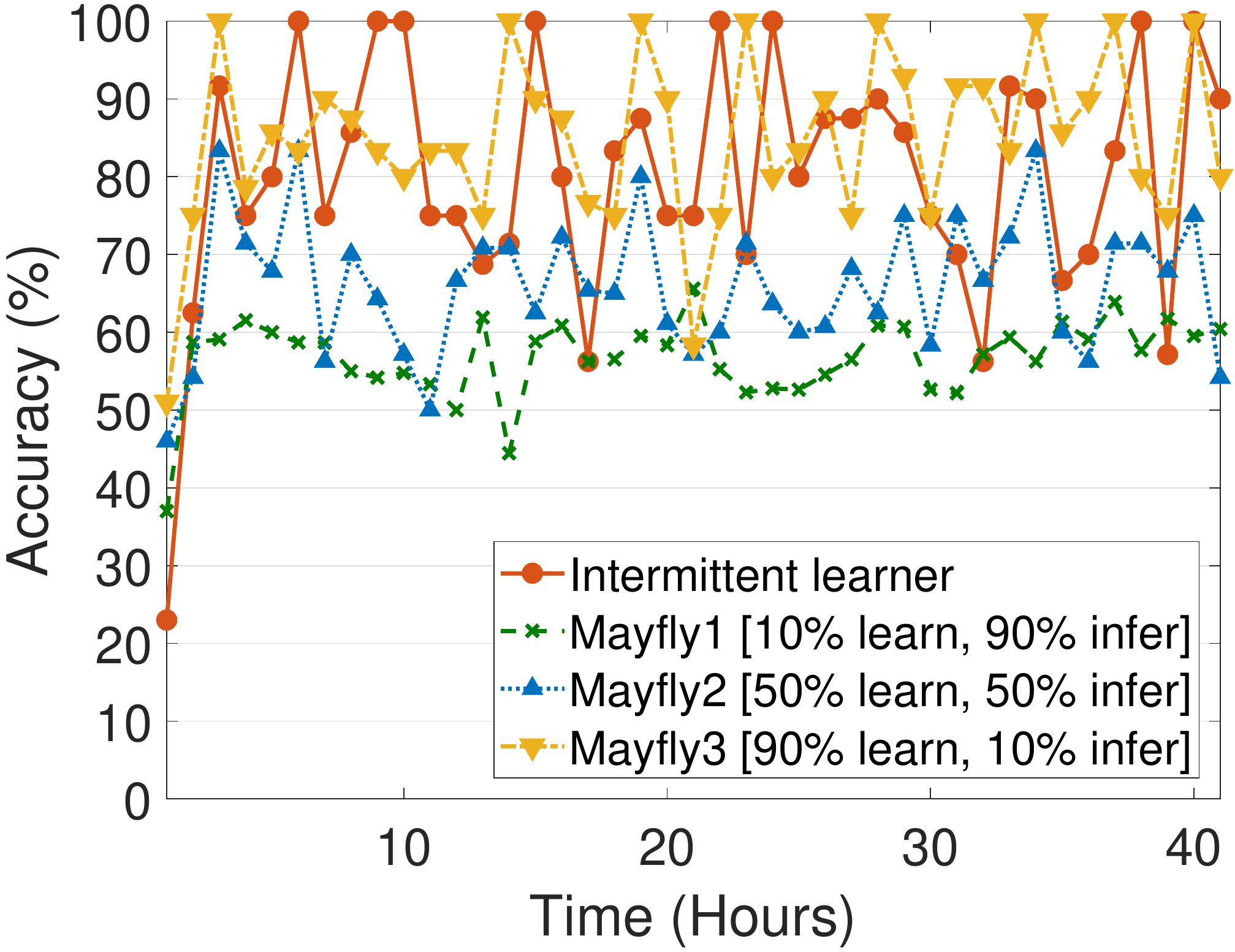}
        \label{fig:rssi_accuracy_duty_mayfly}
    }
    \subfloat[Vibration] 
    {
        \includegraphics[width=0.49\textwidth]{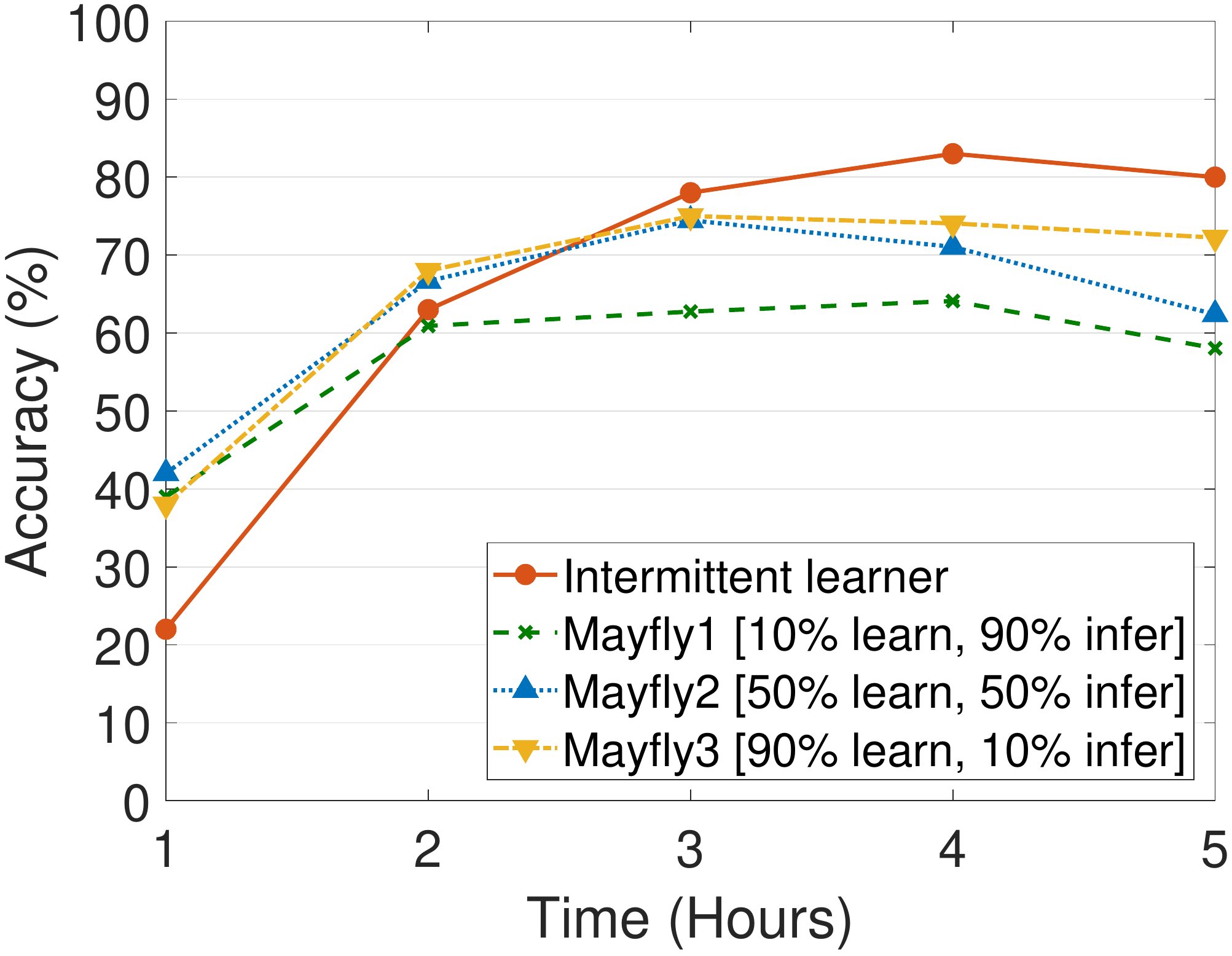}
        \label{fig:vibration_accuracy_dut_mayfly}
    }
\end{minipage} \hspace{1mm}
\begin{minipage}{0.34\textwidth}
\captionsetup{type=table} 
\tiny
\vspace{-12pt}
\caption{Average detection accuracy (\%): Intermittent learner vs. Mayfly.}
\label{table:mayfly}
\begin{center}
\vspace{-10pt}
\begin{tabular}{rcccc}
    \toprule
    \textbf{} & \textbf{Inter.} & \textbf{Mayfly} & \textbf{Mayfly} & \textbf{Mayfly}\\
    \textbf{} & \textbf{Lean} & \textbf{Duty} & \textbf{Duty} & \textbf{Duty}\\
    \textbf{} & & \textbf{10/90} & \textbf{50/50} & \textbf{90/10} \\
    \toprule
    \textbf{\textit{UV}} & \textbf{81\%} & 61\% & 69\% & 79\% \\
    \textbf{\textit{eCO2}} & \textbf{81\%} & 61\% & 71\% & 81\% \\
    \textbf{\textit{TVOC}} & \textbf{83\%} & 63\% & 71\% & 83\% \\
    \hline
    \textbf{\textit{Human}} & \textbf{82\%} & 56\% & 66\% & 84\% \\
    \textbf{\textit{Presence}} & & & & \\
    \hline
    \textbf{\textit{Vibration}} & \textbf{76\%} & 56\% & 63\% & 65\% \\
    \toprule
\end{tabular}
\end{center}
\end{minipage}
\caption{Accuracy comparison with Mayfly (no dynamic action planner and example selection)}
\label{fig:mayfly}
\end{figure}

\begin{figure} [tb]
    \centering
    \subfloat[Air quality] 
    {
        \includegraphics[width=0.32\textwidth]{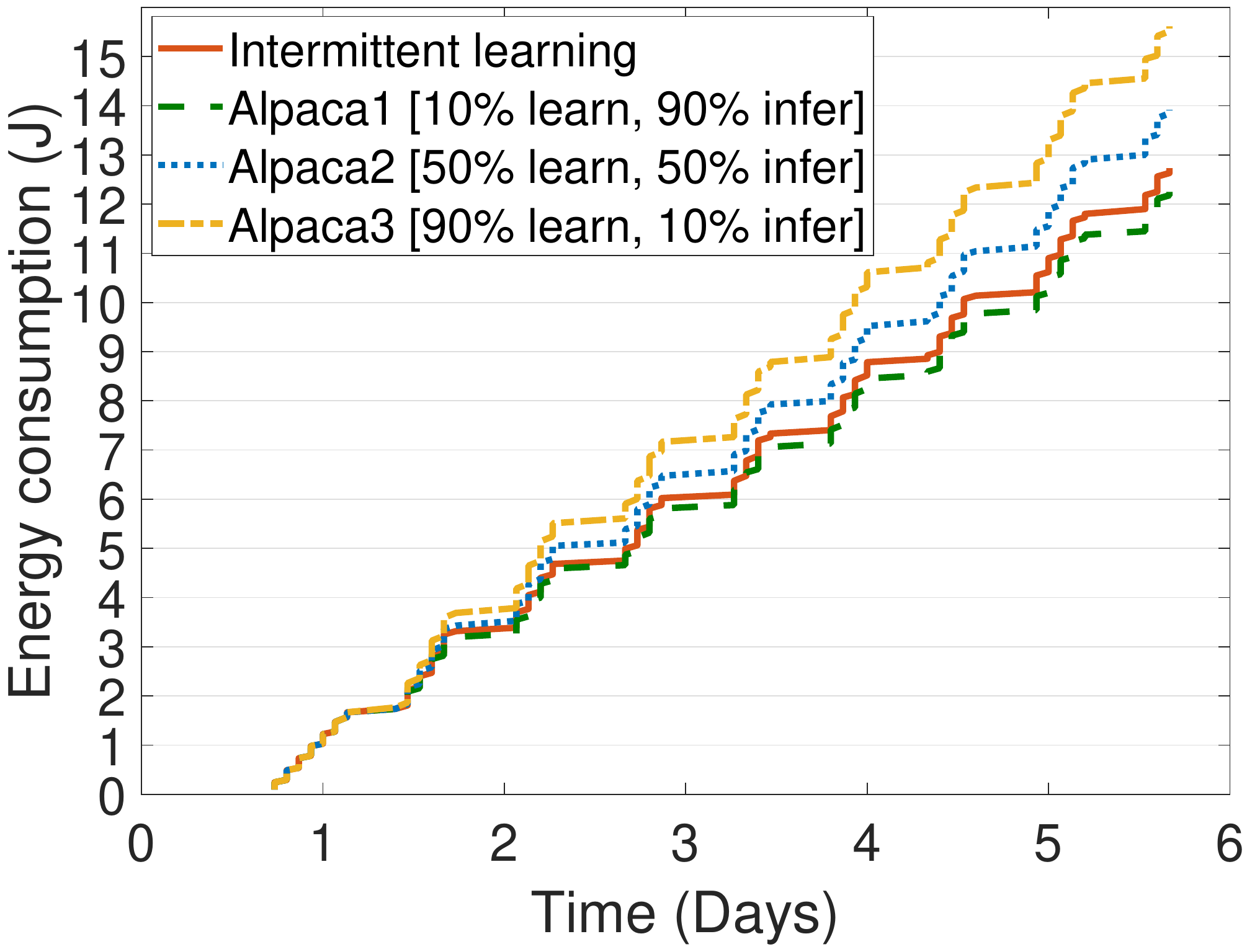}
        \label{fig:air_energy_consumption_alpaca}
    }
    \subfloat[Human presence] 
    {
        \includegraphics[width=0.32\textwidth]{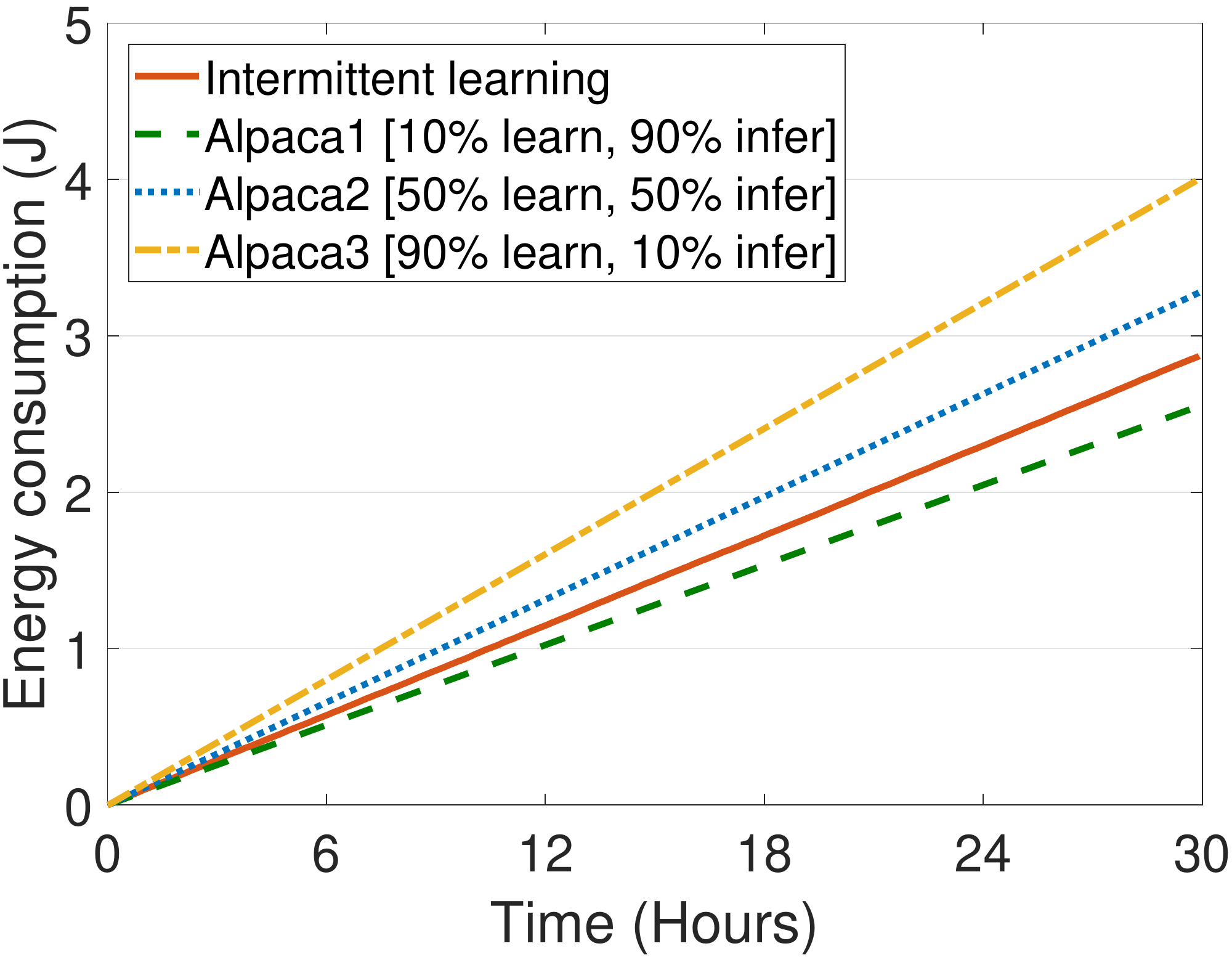}
        \label{fig:rssi_energy_consumption_alpaca}
    }
    \subfloat[Vibration] 
    {
        \includegraphics[width=0.32\textwidth]{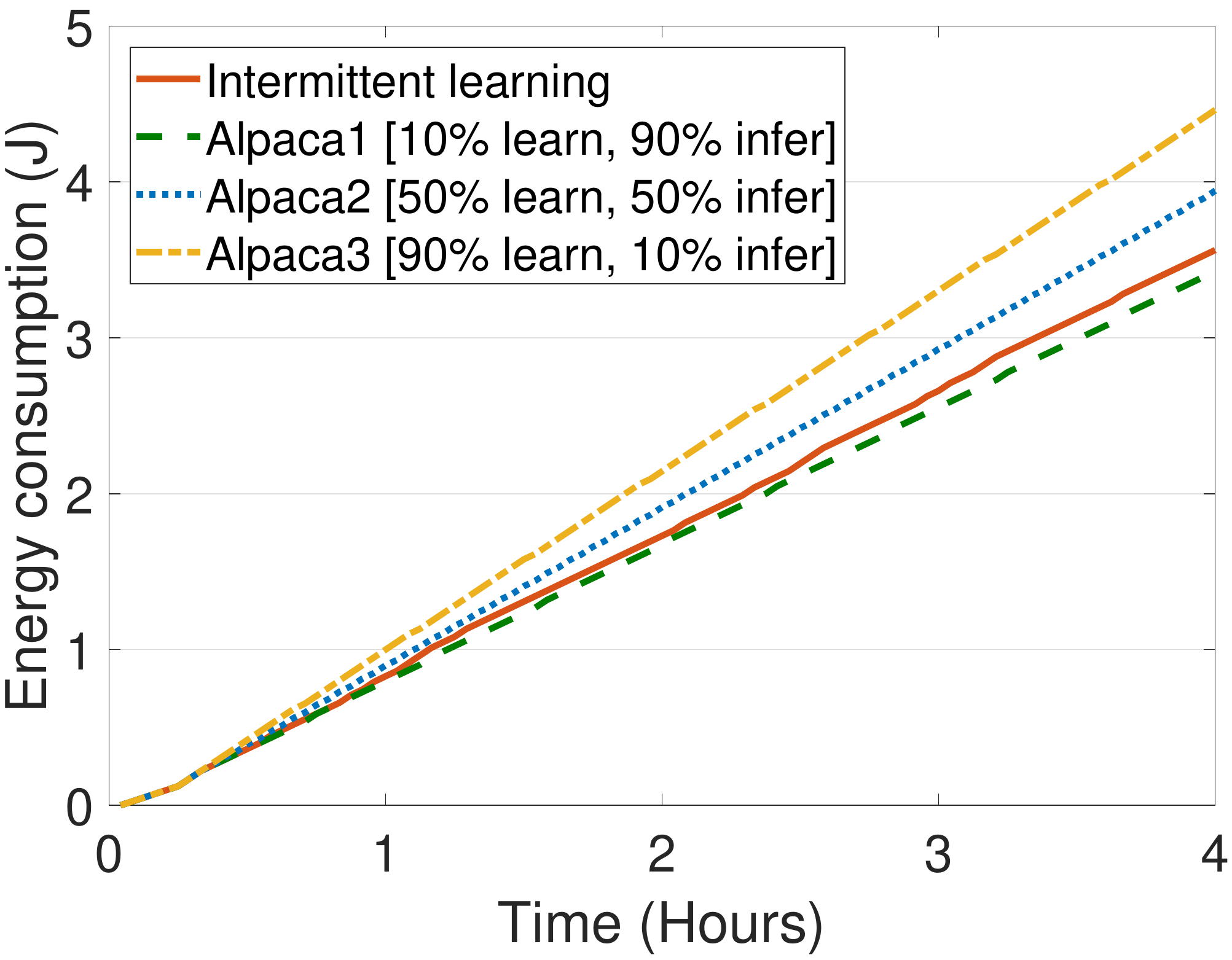}
        \label{fig:vibration_energy_consumption_alpaca}
    }
    \caption{Energy consumption comparison with Alpaca (no dynamic action planner and example selection)}
    \label{fig:energy_consumption_alpaca}
\end{figure}

We compare the accuracy of the three intermittent learners (air-quality, human presence, and vibration learning) against two state-of-the-art task-based intermittent computing systems: Alpaca~\cite{maeng2017alpaca} and Mayfly~\cite{hester2017timely}. Both of the baseline systems execute the same learning algorithm as ours, but they do not implement the proposed framework. Instead, the two baseline systems repeat a fixed sequence of actions periodically, and they \emph{duty-cycle} the execution of \textit{learn} and \textit{infer} actions according to a predefined schedule. For example, Alpaca with a duty-cycle parameter of [90\% \textit{learn}, 10\% \textit{infer}] executes the \textit{learn} action 90\% of the time and the \textit{infer} action 10\% of the time, after executing the \textit{sense} and \textit{extract} actions. Mayfly works the same way as Alpaca with the exception that it discards stale examples by setting a data expiration interval. None of the baseline solutions implement example selection heuristics.

Figures~\ref{fig:alpaca}(a)-(e) and~\ref{fig:mayfly}(a)-(e) compare the accuracy of the intermittent learners against Alpaca and Mayfly-based implementation of the same applications. We use three duty-cycle parameters for the baseline solutions: [10\% learn, 90\% infer], [50\% learn, 50\% infer], and [90\% learn, 10\% infer]. Table~\ref{table:alpaca} and \ref{table:mayfly} summarize the results. Overall, the intermittent learning systems achieve 80\% average accuracy while Alpaca and Mayfly-based implementations achieve 54\%--79\% and 59\%--78\% average accuracy, respectively, depending on the duty-cycle parameters. For both Alpaca and Mayfly, as the amount of \textit{learn} action increases from 10\% to 90\%, the accuracy increases, and finally, it becomes comparable to the accuracy of the intermittent learning systems when the duty-cycle has 90\% \textit{learn} actions. However, the intermittent learning systems achieve 80\% accuracy by executing 50\% less number of \textit{learn} actions compared to Alpaca and Mayfly for [90\% learn, 10\% infer] duty-cycle. As a result, the intermittent learners increase the inference throughput by performing more \textit{infer} actions than the baseline intermittent computing systems that waste time and energy in performing unproductive \textit{learn} actions. We also observe that different actions are chosen by the dynamic action planner at run-time based on the state of the system, while the baseline systems follow a repeated fixed-sequence of actions, e.g., 90\% of the time [\textit{sense, extract, learn}] and 10\% of the time [\textit{sense, extract, infer}] sequence without caring for the learning performance.

\begin{figure} [tb]
    \subfloat[Air quality 1 (UV)] 
    {
        \includegraphics[width=0.32\textwidth]{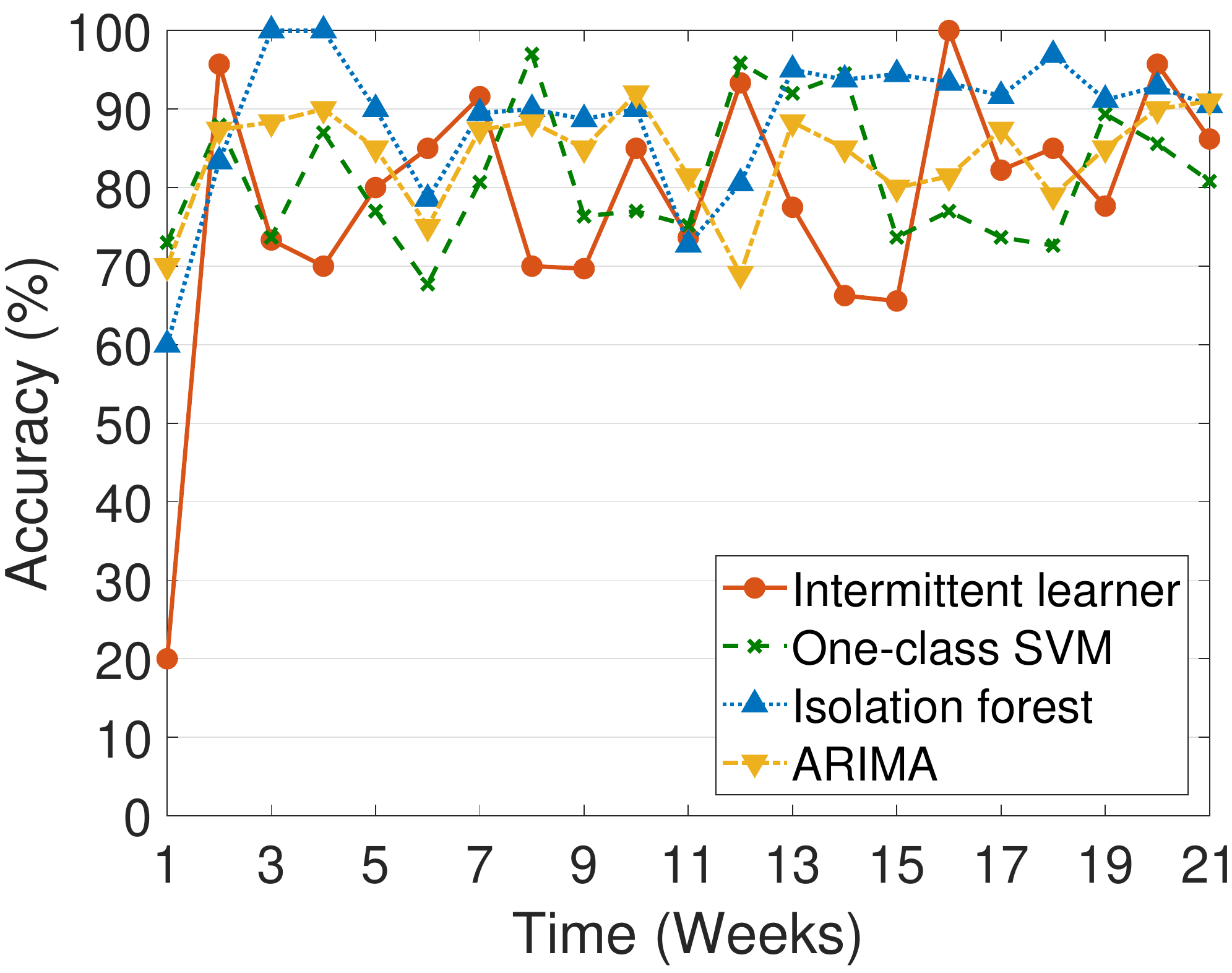}
        \label{fig:uv_accuracyy_offline}
    }
    \subfloat[Air quality 2 (eCO2)] 
    {
        \includegraphics[width=0.32\textwidth]{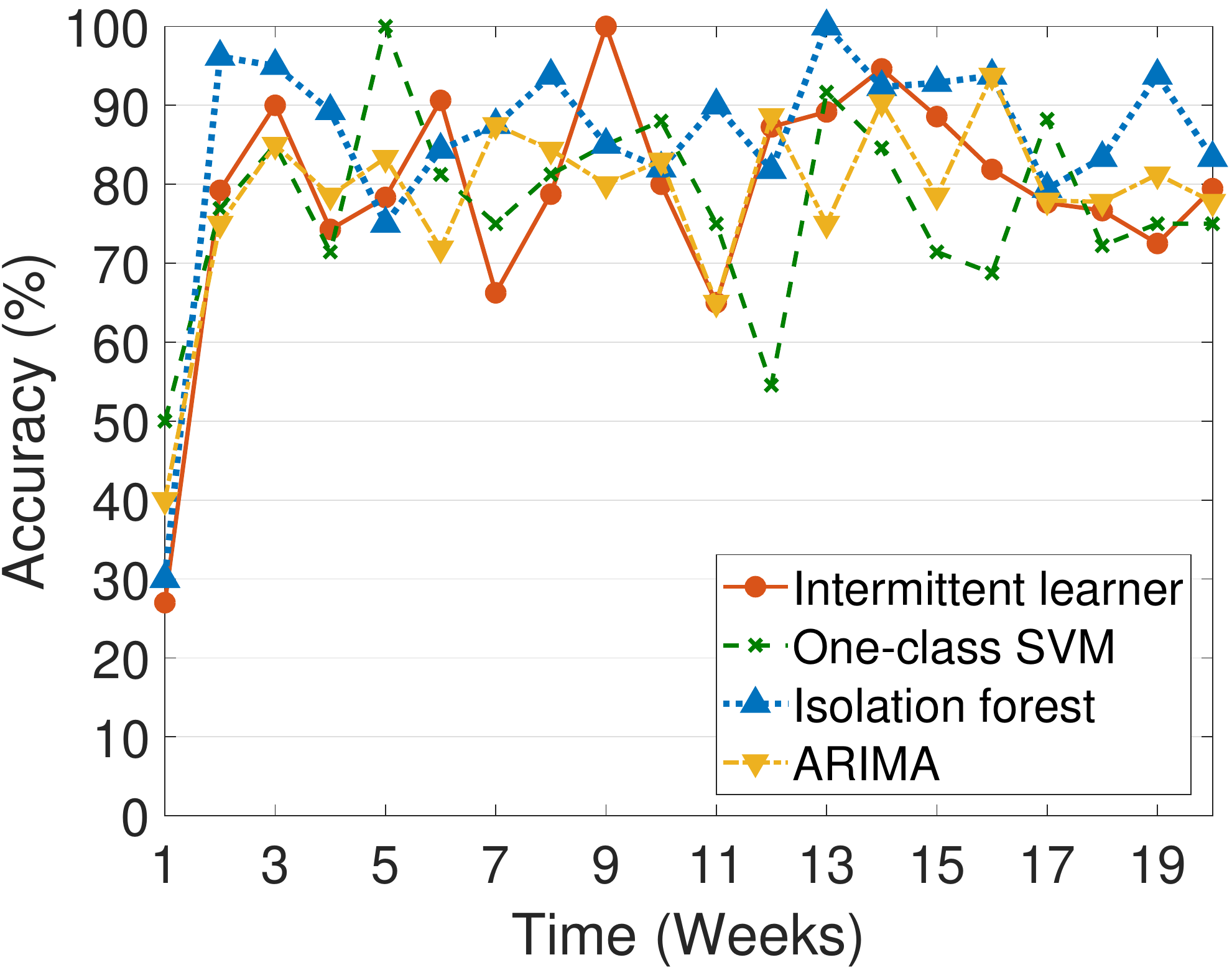}
        \label{fig:eco2_accuracyy_offline}
    }
    \subfloat[Air quality 3 (TVOC)] 
    {
        \includegraphics[width=0.32\textwidth]{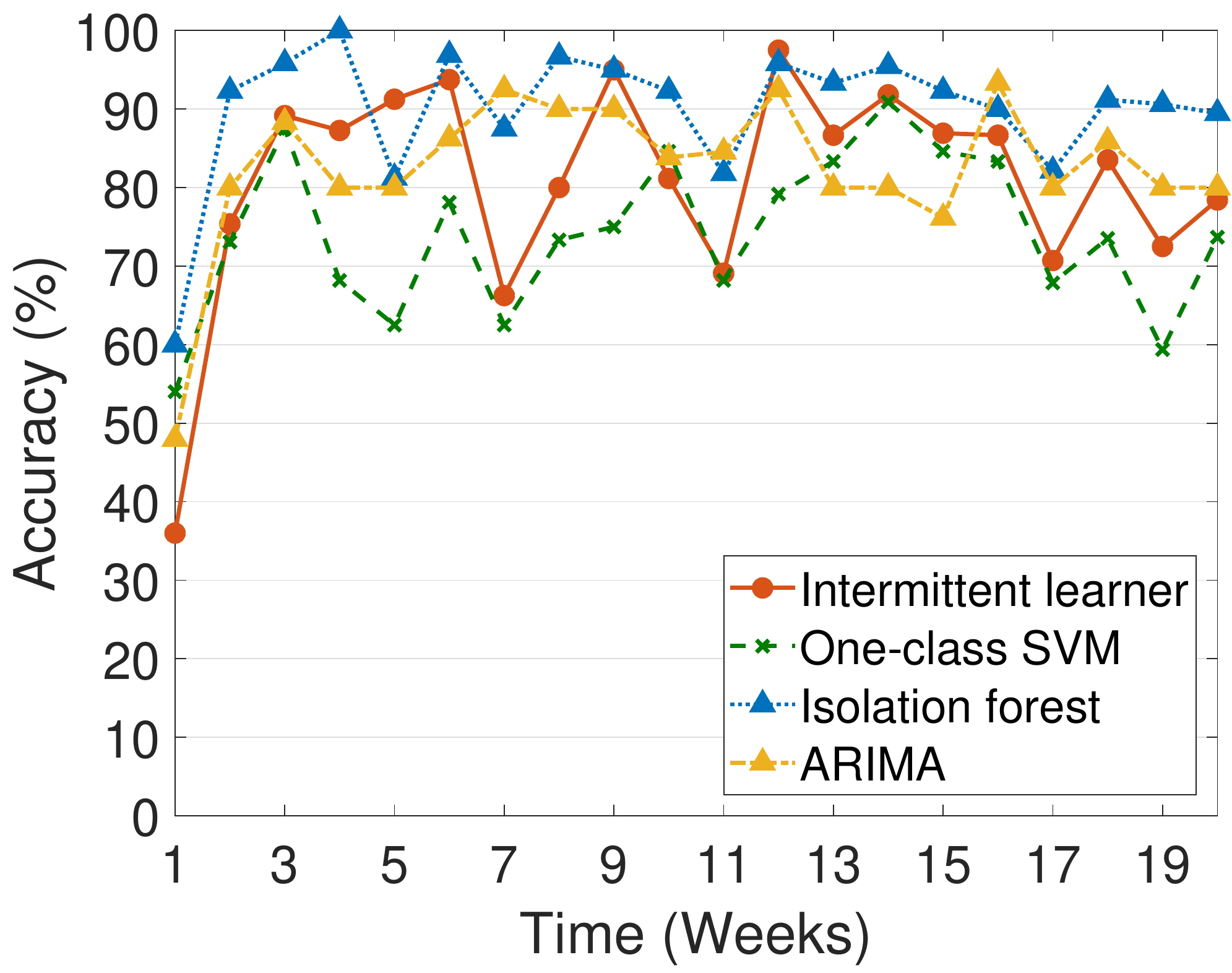}
        \label{fig:tvoc_accuracyy_offline}
    }
    \\
\begin{minipage}{0.64\textwidth}
    \subfloat[Human Presence] 
    {
        \includegraphics[width=0.49\textwidth]{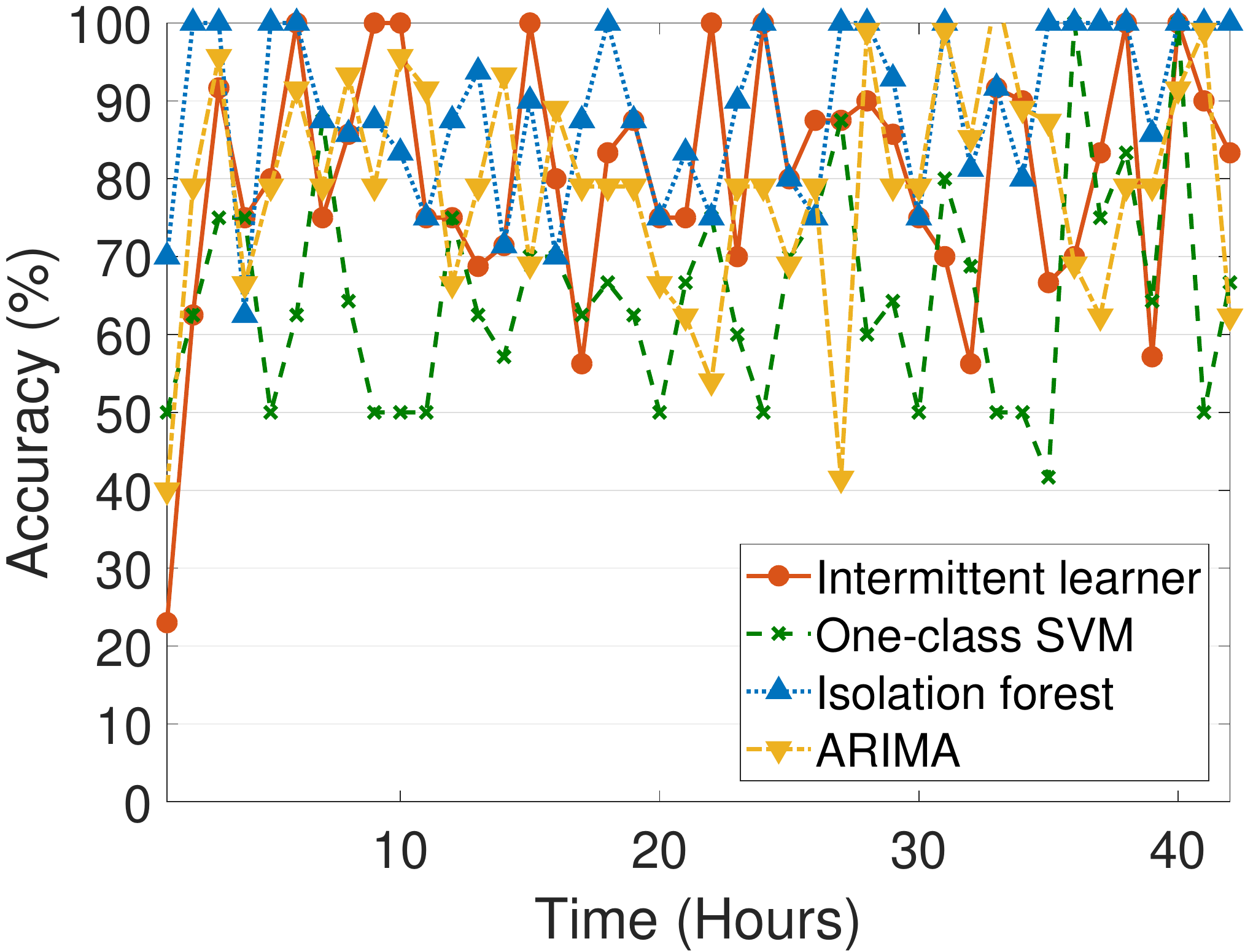}
        \label{fig:rssi_accuracyy_offline}
    }
    \subfloat[Vibration] 
    {
        \includegraphics[width=0.49\textwidth]{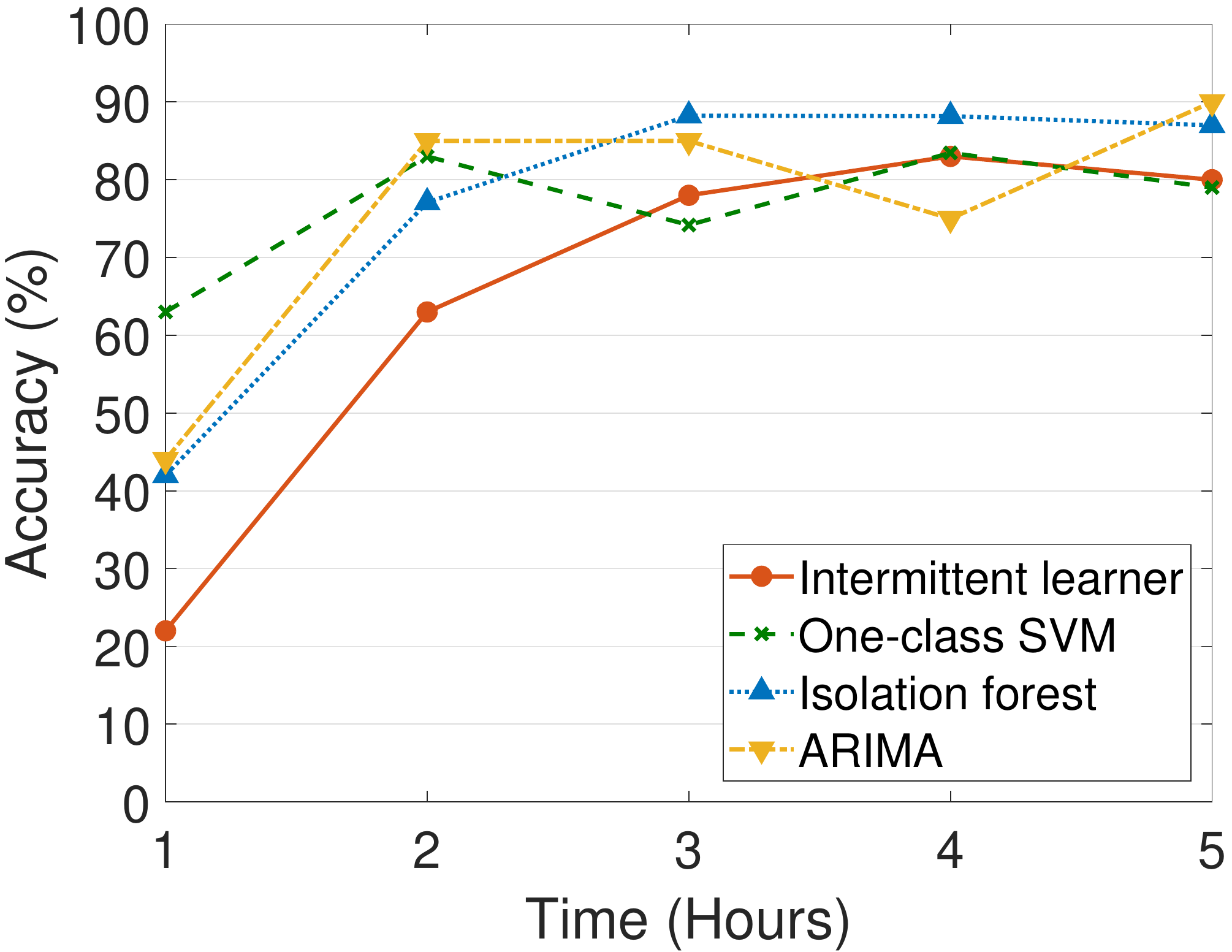}
        \label{fig:vibration_accuracyy_offline}
    }
\end{minipage} \hspace{1mm}
\begin{minipage}{0.34\textwidth}
\captionsetup{type=table} 
\vspace{-2pt}
\tiny
\caption{Average detection accuracy (\%): Intermittent learner vs. offline machine learning anomaly detectors.}
\vspace{-10pt}
\label{table:offline}
\begin{center}
\begin{tabular}{rcccc}
    \toprule
    \textbf{} & \textbf{Inter.} & \textbf{One-} & \textbf{Isolation} & \textbf{ARIMA}\\
    \textbf{} & \textbf{Learn} & \textbf{class} & \textbf{Forest} & \\
    &  & \textbf{SVM} &  & \\
    \toprule
    \textbf{\textit{UV}} & \textbf{81\%} & 81\% & 88\% & 84\% \\
    \textbf{\textit{eCO2}} & \textbf{81\%} & 78\% & 88\% & 80\% \\
    \textbf{\textit{TVOC}} & \textbf{83\%} & 75\% & 89\% & 80\% \\
    \hline
    \textbf{\textit{Human}} & \textbf{82\%} & 70\% & 85\% & 79\% \\
    \textbf{\textit{Presence}} & & & & \\
    \hline
    \textbf{\textit{Vibration}} & \textbf{76\%} & 79\% & 85\% & 83\% \\
    \toprule
\end{tabular}
\end{center}
\end{minipage}
\caption{Accuracy comparison with offline machine learning anomaly detectors (one-class SVM, isolation forest and ARIMA).}
\label{fig:offline}
\end{figure}

Figures~\ref{fig:energy_consumption_alpaca}(a)-(c) compare the total energy consumption of the intermittent leaning framework and Alpaca-based implementation of the three applications over time. For all three applications, the intermittent learning system consumes less energy than Alpaca with [90\% learn, 10\% infer] and [50\% learn, 50\% infer] duty-cycle parameters, but consume slightly more energy than Alpaca with [10\% learn, 90\% infer] duty-cycle parameters. For instance, the proposed system consumes 37\% less energy than Alpaca with [90\% learn, 10\% infer] duty-cycle at hour 30 for the human presence learning experiment in Figure~\ref{fig:energy_consumption_alpaca}(b), but still achieves similar average accuracy to Alpaca with [90\% learn, 10\% infer] duty-cycle. In other words, the intermittent learning system achieves at least 1.6$\times$ higher accuracy than Alpaca when both the systems consume the same amount of energy. This is because the dynamic action planner intelligently selects actions at run-time, which leads the system to spend less energy and time. Furthermore, the data selection module trains the system with examples that are likely to improve its learning performance and prevents the system from wasting energy in learning examples that do not.

\subsection{Comparison with Offline Machine Learning Algorithms}
\label{subsec:offline}

We compare the accuracy of anomaly detection of the three intermittent learners against three widely used offline anomaly detectors that are based on: 1) one-class SVM (Support Vector Machine)~\cite{manevitz2001one} with RBF (Radial Basis Function) kernel, 2) isolation forest~\cite{liu2012isolation,liu2008isolation}, and 3) Auto-Regressive Integrated Moving Average (ARIMA)-based clustering. Unlike the proposed framework which selects examples to learn at run-time, these offline detectors use all the examples for anomaly detection at once. Figures~\ref{fig:offline}(a)-(e) compare the accuracy of the intermittent learners against the offline anomaly detectors. The average accuracy of these detectors are summarized in Table~\ref{table:offline}. We observe that the intermittent learners achieve a comparable accuracy (80\%) to the three offline detectors (78\%, 86\% and 83\% for the one-class SVM, isolation forest and ARIMA, respectively) while selecting and learning only 44\% of the input examples and judiciously discarding 56\% of the examples that are unlikely to increase the accuracy of the learner by using the round-robin selection method.

\subsection{Effect of Example Selection Heuristics}
\label{subsec:heuristics}

\begin{figure} [tb]
    \subfloat[Air quality (UV)] 
    {
        \includegraphics[width=0.32\textwidth]{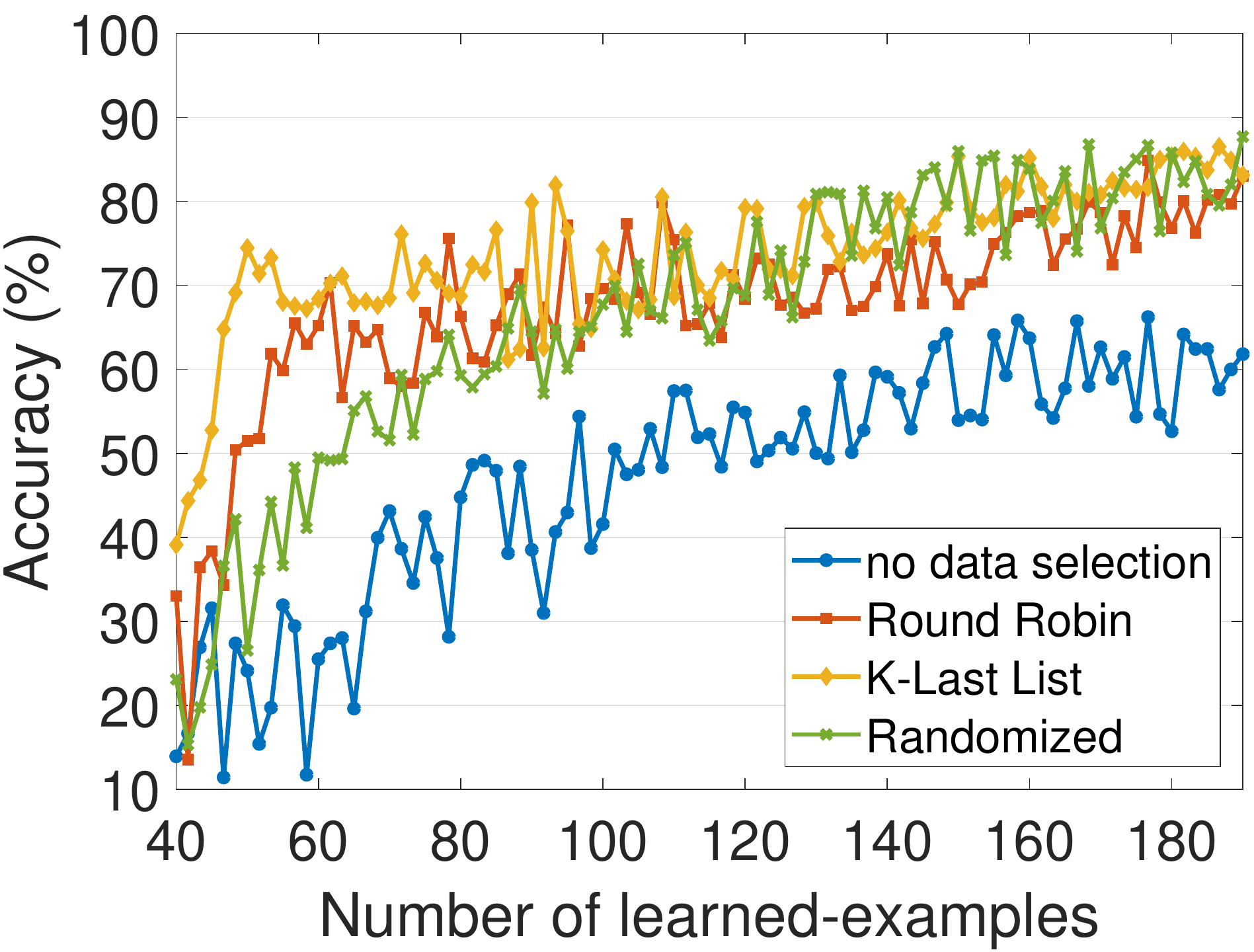}
        \label{fig:air_heuristic}
    }
    \subfloat[Human presence] 
    {
        \includegraphics[width=0.32\textwidth]{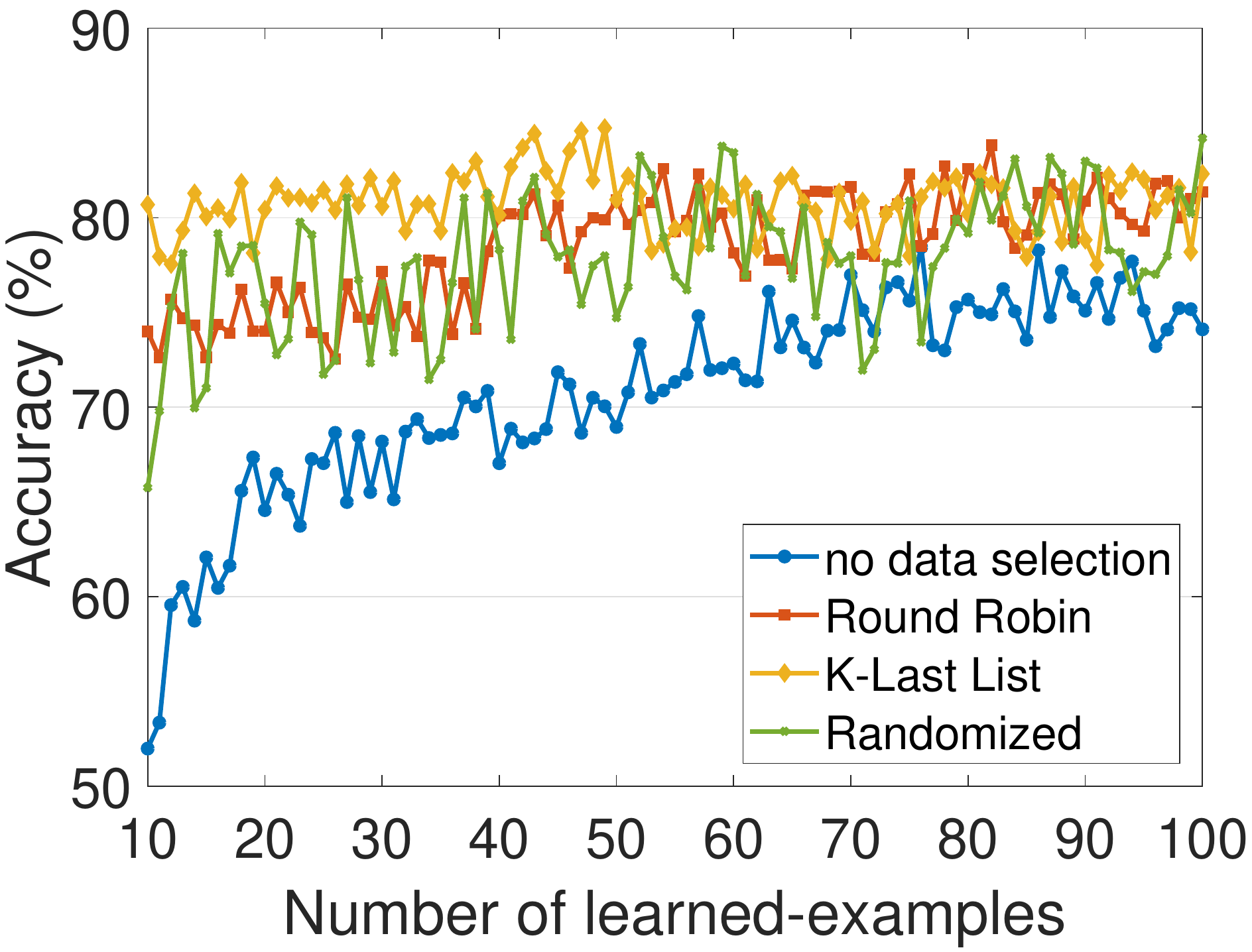}
        \label{fig:rssi_heuristic}
    }
    \subfloat[Vibration] 
    {
        \includegraphics[width=0.32\textwidth]{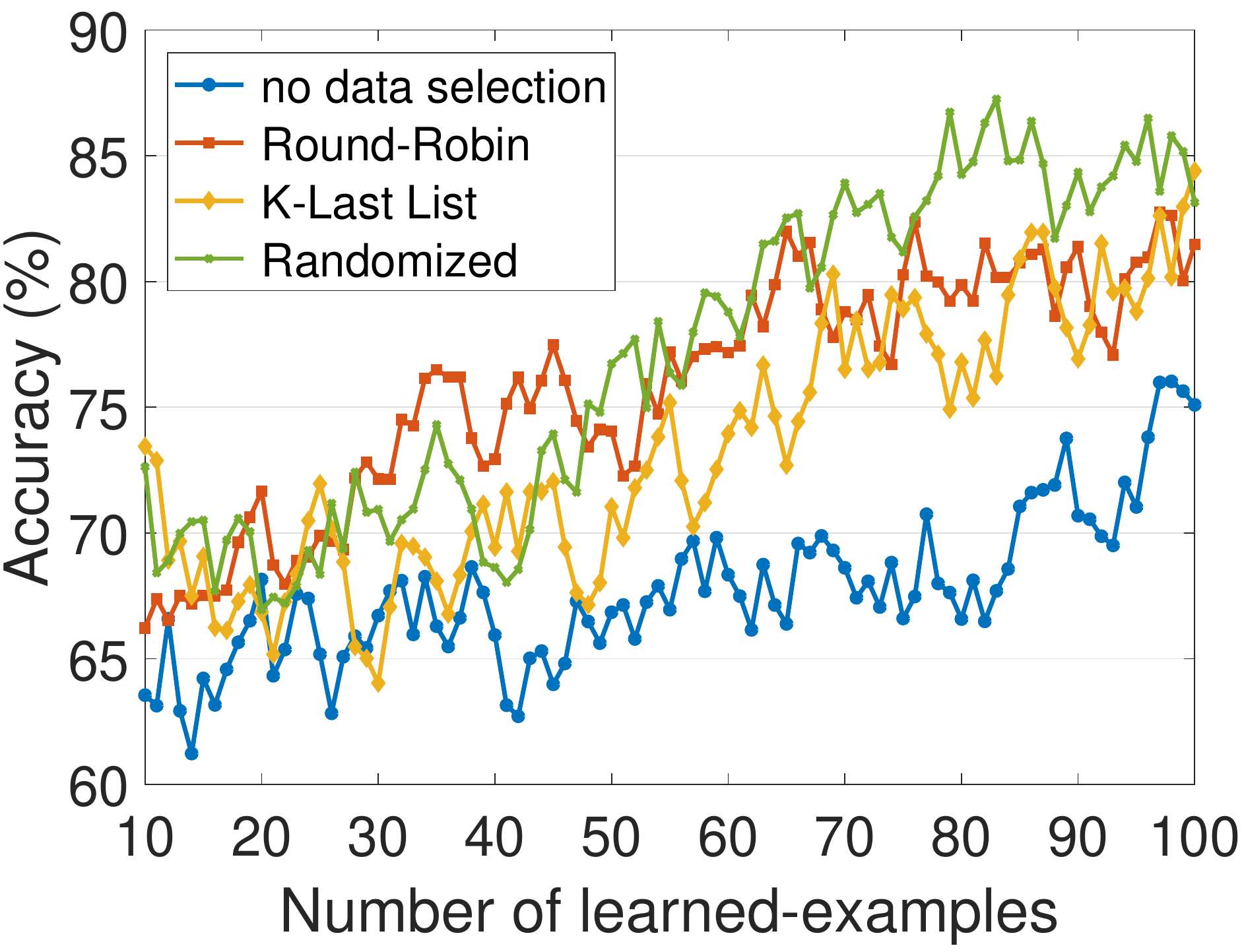}
        \label{fig:vibration_heuristic}
    }
    \caption{Effect of example selection heuristics: accuracy vs. number of learned-examples}
    \label{fig:heuristic}
    \subfloat[Air quality (UV)] 
    {
        \includegraphics[width=0.32\textwidth]{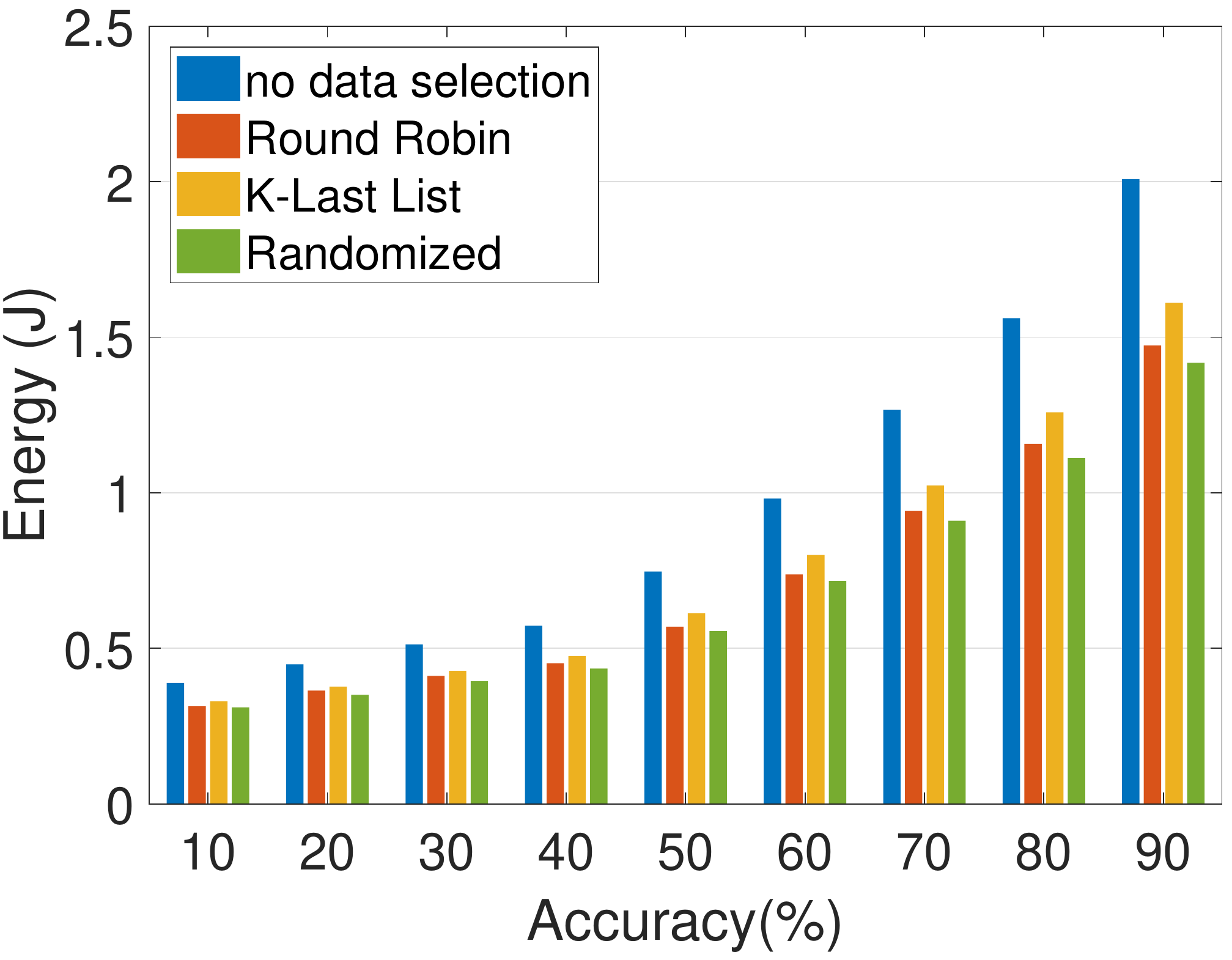}
        \label{fig:air_heuristic_energy}
    }
    \subfloat[Human presence] 
    {
        \includegraphics[width=0.32\textwidth]{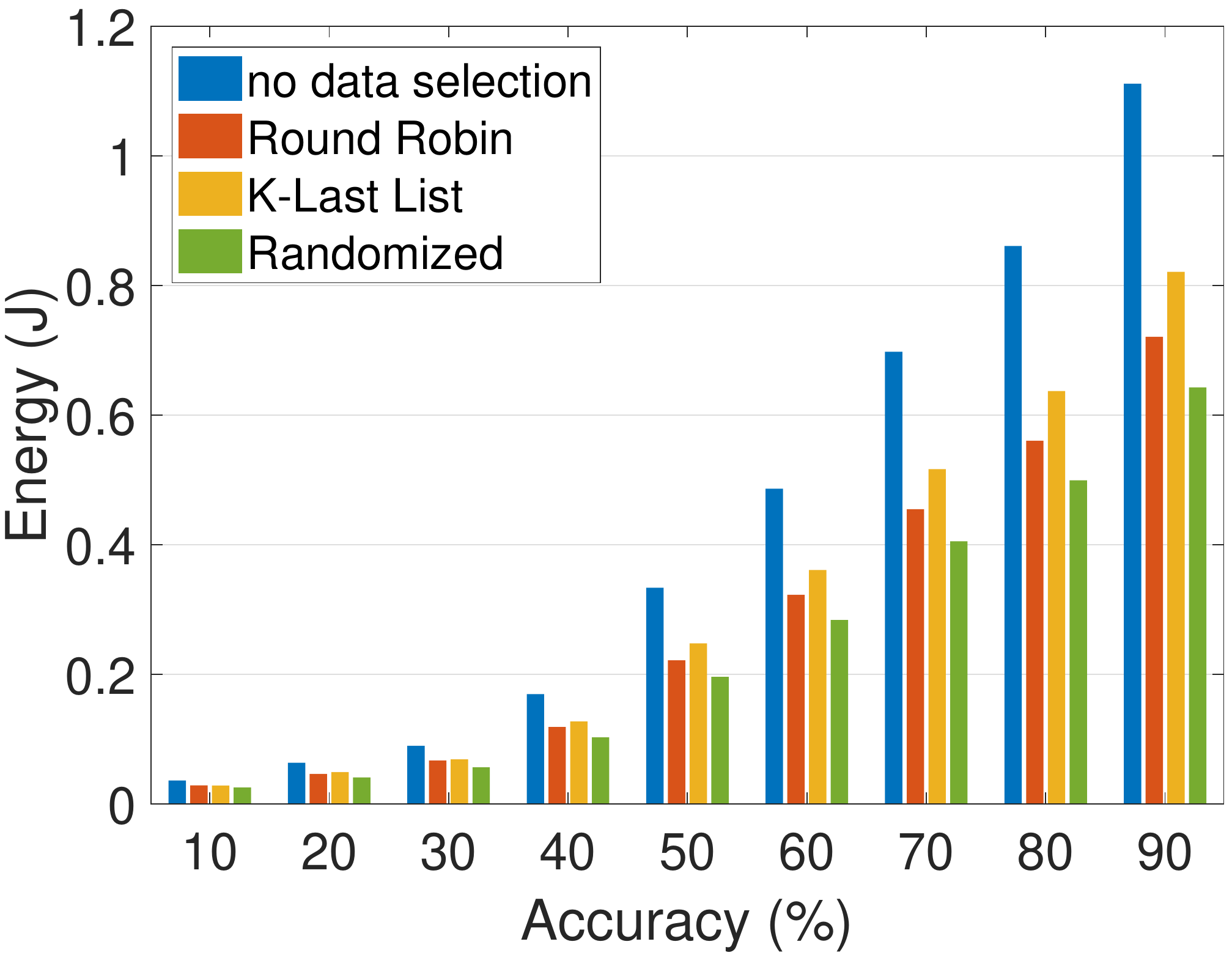}
        \label{fig:rssi_heuristic_energy}
    }
    \subfloat[Vibration] 
    {
        \includegraphics[width=0.32\textwidth]{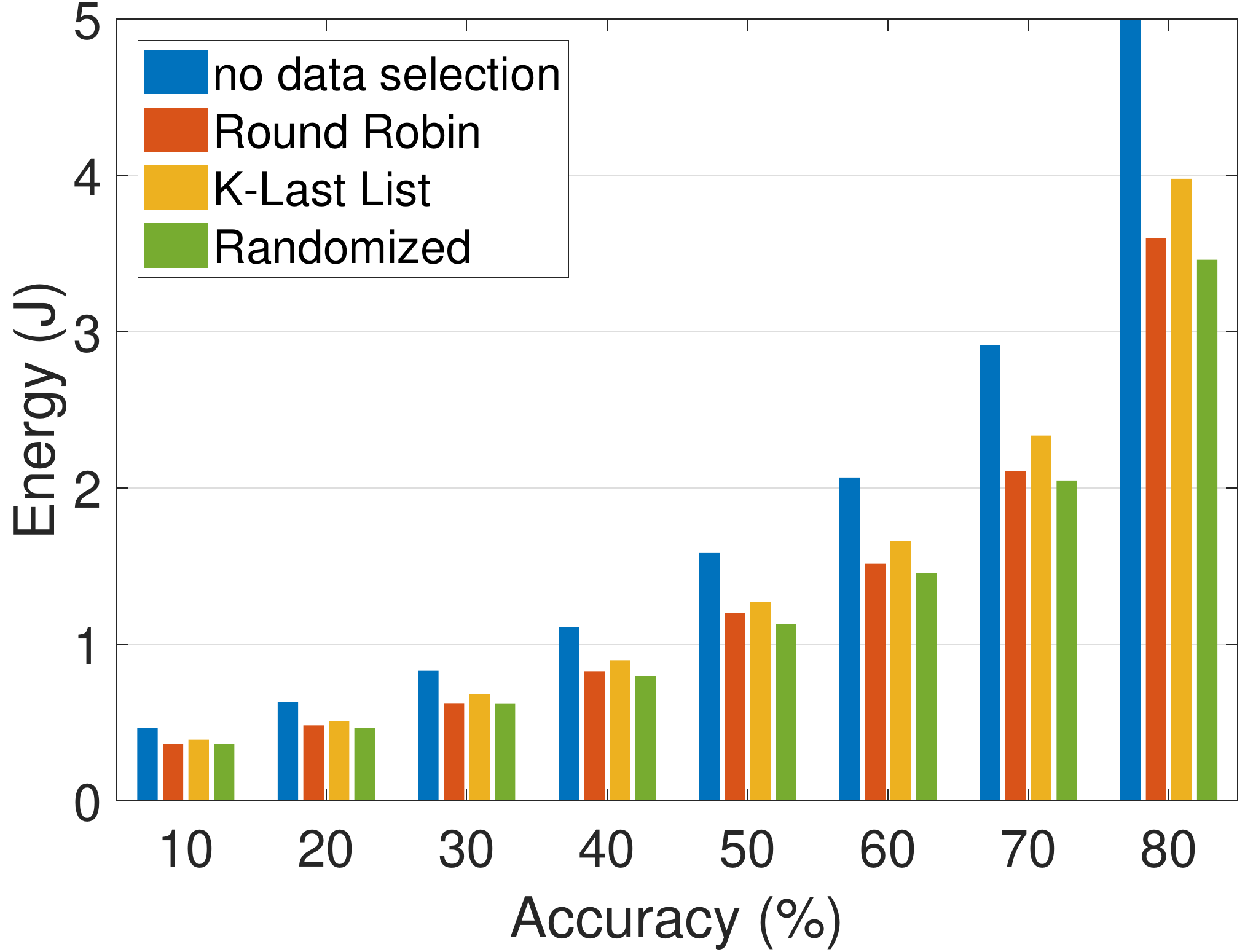}
        \label{fig:vibration_heuristic_energy}
    }
    \caption{Effect of example selection heuristics: accuracy vs. energy}
    \label{fig:heuristic_energy}
\end{figure}

To evaluate the effect of example selection heuristics, we compare the three proposed training example selection heuristics, i.e., round-robin, $k$-last lists, and randomized selection against no data selection strategy, i.e., every example is used for training. Figures~\ref{fig:heuristic}(a)-(c) plot the detection accuracy over the number of learned-examples for each heuristic. We observe that all three heuristics consistently demonstrate higher accuracy than the no data selection policy. This may seem counter-intuitive at first, but the reason for a higher accuracy by any of the heuristics than the no data selection policy is that in Figures~\ref{fig:heuristic}(a)-(c), we report the actual number of examples learned by the four strategies, which is not generally the same as the number of examples entering the system. For instance, the no data selection policy learns all of the 180 examples it encounters and achieves 60\%; whereas the round-robin heuristic achieves 80\% accuracy after learning 180 examples, but it has encountered much more than 180 examples and chose to learn only the best 180 ones. By skipping examples that are unlikely to improve the accuracy, the intermittent learning systems with these selection heuristics achieve the same level of accuracy with less energy, which is evident from Figures~\ref{fig:heuristic_energy}(a)-(c).

In both air quality and human presence learning systems, the $k$-last lists selection increases the accuracy rapidly in the beginning as shown in Figure~\ref{fig:heuristic}(a) and \ref{fig:heuristic}(b). The round-robin and randomized heuristic catch up with the accuracy of $k$-last lists as more examples are seen, and finally, the accuracy converges to 82\% and 80\% for air quality and human presence learning systems, respectively. For the vibration learning system in  Figure~\ref{fig:heuristic}(c), the $k$-last lists and the randomized selection reach a similar level of accuracy (83--84\%) after learning about 100 examples, but the randomized selection heuristic reaches the highest accuracy (87\%) earlier than the $k$-last lists. The round-robin heuristic shows a better performance for a smaller number of examples (20--45 examples) in the beginning, but gets caught up by the other two after learning 50--70 examples. 
Considering the high computational complexity and energy cost of $k$-last lists, we conclude that randomization or round-robin heuristics are reasonable choices for systems having tighter energy constraints.

\subsection{Effect of Energy Harvesting Pattern}
\label{subsec:harvesting_pattern}

\begin{figure} [tb]
    \subfloat[Air quality] 
    {
        \includegraphics[width=0.32\textwidth]{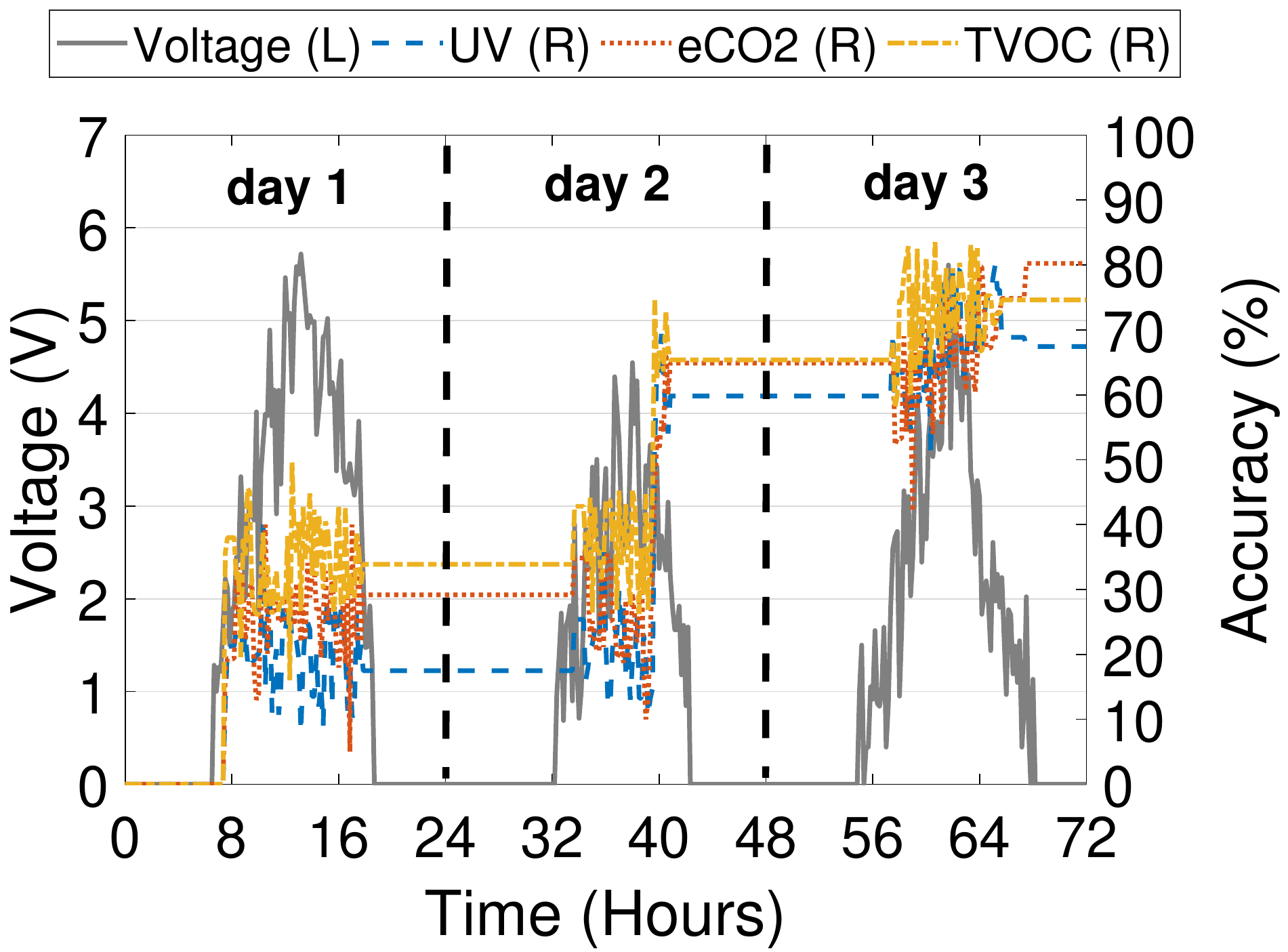}
        \label{fig:air_harvesting_pattern}
    }
    \subfloat[Human presence] 
    {
        \includegraphics[width=0.32\textwidth]{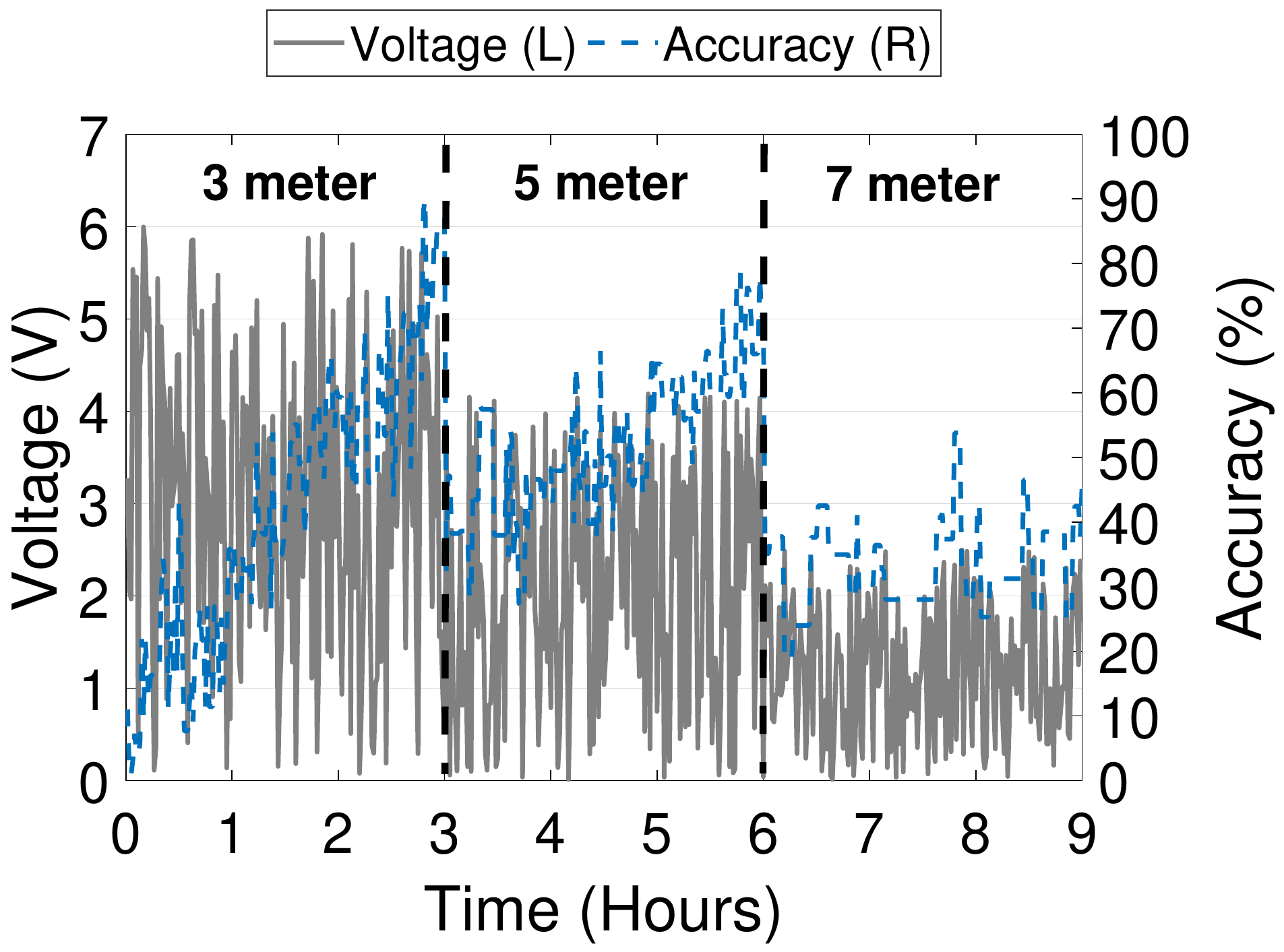}
        \label{fig:rssi_harvesting_pattern}
    }
    \subfloat[Vibration] 
    {
        \includegraphics[width=0.32\textwidth]{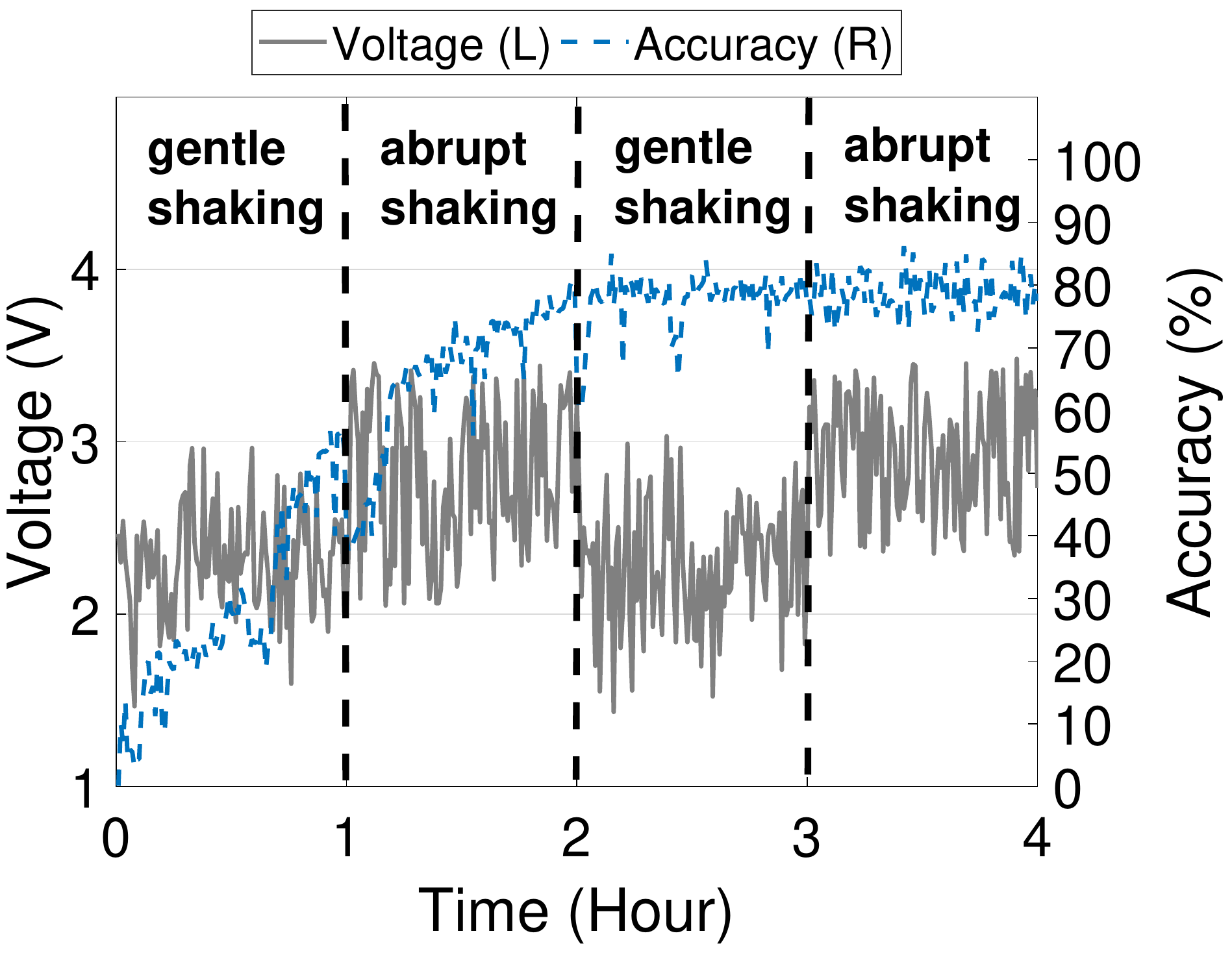}
        \label{fig:vibration_harvesting_pattern}
    }
    \caption{Effect of Energy Harvesting Pattern}
    \label{fig:harvesting_pattern}
\end{figure}

In Figure~\ref{fig:harvesting_pattern}(a)-(c), we plot the energy harvesting patterns (voltage level) of the energy harvesters (solar, RF and piezoelectric harvester) for the three systems and evaluate their accuracy over time. In order to assess the effect of energy harvesting pattern on the detection accuracy, the time period is divided into segments that are expected to have different energy harvesting patterns.

Figure~\ref{fig:harvesting_pattern}(a) shows the solar energy harvesting pattern, along with the accuracy of the air quality learning system for three consecutive days. As shown in the figure, the detection accuracy improves during the daytime (8 am--5 pm) as the system learns new examples using the harvested energy. At night, the system is essentially off, and in the next morning, the system resumes learning new examples, and its accuracy improves over time. We also occasionally observe interruptions in the otherwise continuous energy harvesting pattern during the daytime due to inadequate sunlight. During these periods of inadequate energy supply when a full cycle of learning is not possible, the system senses, selects, and saves the examples that have the potential to improve accuracy. When sufficient energy is harvested again, the system resumes learning the saved examples. Thus, the intermittent learner does not require sensor data to be acquired and processed simultaneously in real-time. Acquired data are buffered by the system in the non-volatile memory, and the CPU processes it when energy is available. This cannot be achieved by the state-of-the-art intermittent computing system~\cite{hester2017timely} that collects sensor data without considering their utility towards learning and discards them when they are stale---which leaves no data to learn when energy becomes available.

Figure~\ref{fig:harvesting_pattern}(b) shows the RF energy harvesting pattern and the detection accuracy of the human presence learning system for nine hours. Every three hours, the system is placed at different distances (3, 5, and 7 meters) from the RF energy source, and the amount of energy harvested at each distance is measured. As expected, less amount of energy gets harvested as the distance increases with an average of 3.1V, 2.2V, and 0.9V at 3, 5, and 7 meters, respectively---which results in a decrease in detection accuracy with the distance, i.e., 86\%, 74\%, and 46\% at hour 3, 6, and 9, respectively. Since a change of location causes changes in the RSSI pattern, the system needs to learn a new RSSI pattern whenever it relocates. However, due to the less harvested energy at a longer distance, it takes more time to harvest energy to execute the \textit{learn} and the \textit{infer} actions, which slows down the execution rate of both learning and inference. The difficulty in learning RSSI patterns from weaker signals at a longer distance is another reason for the decrease in accuracy.

Figure~\ref{fig:harvesting_pattern}(c) shows the piezoelectric energy harvesting pattern and the detection accuracy of the intermittent vibration learning system. The time period is divided into four one-hour segments. To capture different harvesting patterns, the harvester is shaken gently during the first and third hour and abruptly during the second and the fourth hour. The accuracy of the learner increases over time and converges to 80\% at hour 4, irrespective of the shaking type and consequent energy harvesting pattern (2.29V, 2.81V 2.27V, and 2.92V on average at hour 1, 2, 3, and 4, respectively). The energy harvesting pattern, in this case, does not seem to affect the accuracy much since the amount of energy harvested from both gentle and abrupt shakes are above the minimum operation voltage (2V) of the system which allows it to select learning data and execute \textit{learn} action stably.

\subsection{Time and Energy Overhead}
\label{subsec:energy_consumption_and_time}
We measure the energy consumption and execution time of all the action primitives, the dynamic action planner, and the three example selection heuristics to quantify the overhead of the proposed framework. We use an MSP430FR5994 as the experimental platform and measure the energy consumption of each module using the EnergyTrace tool~\cite{EnergyTrace}.

\begin{figure*} [tb]
\centering
\begin{minipage}{.66\textwidth}
    \centering
    \subfloat[Energy consumption ($k$-NN)] 
    {
        \includegraphics[width=0.48\textwidth]{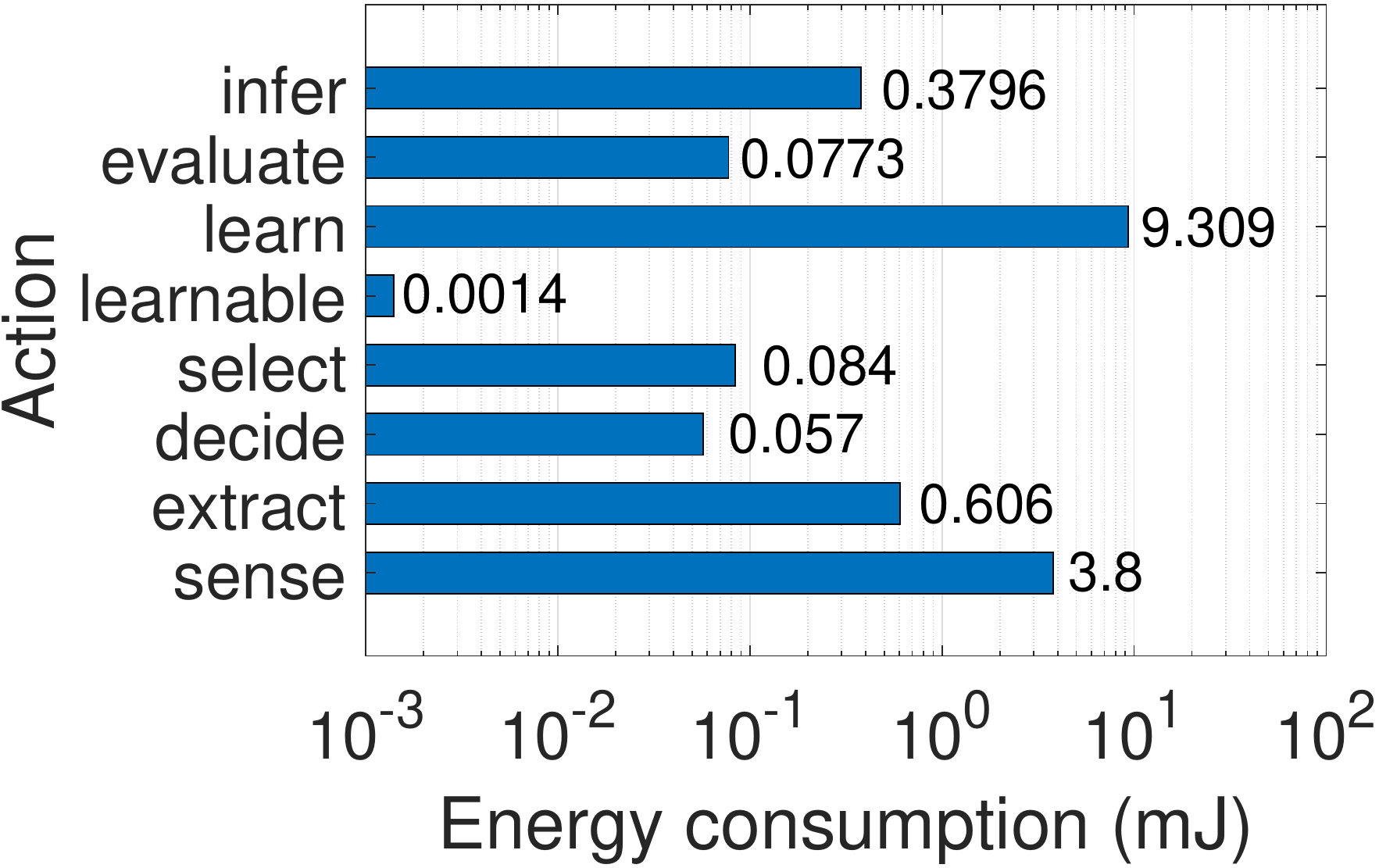}
    }
    \subfloat[Execution time ($k$-NN)] 
    {
        \includegraphics[width=0.48\textwidth]{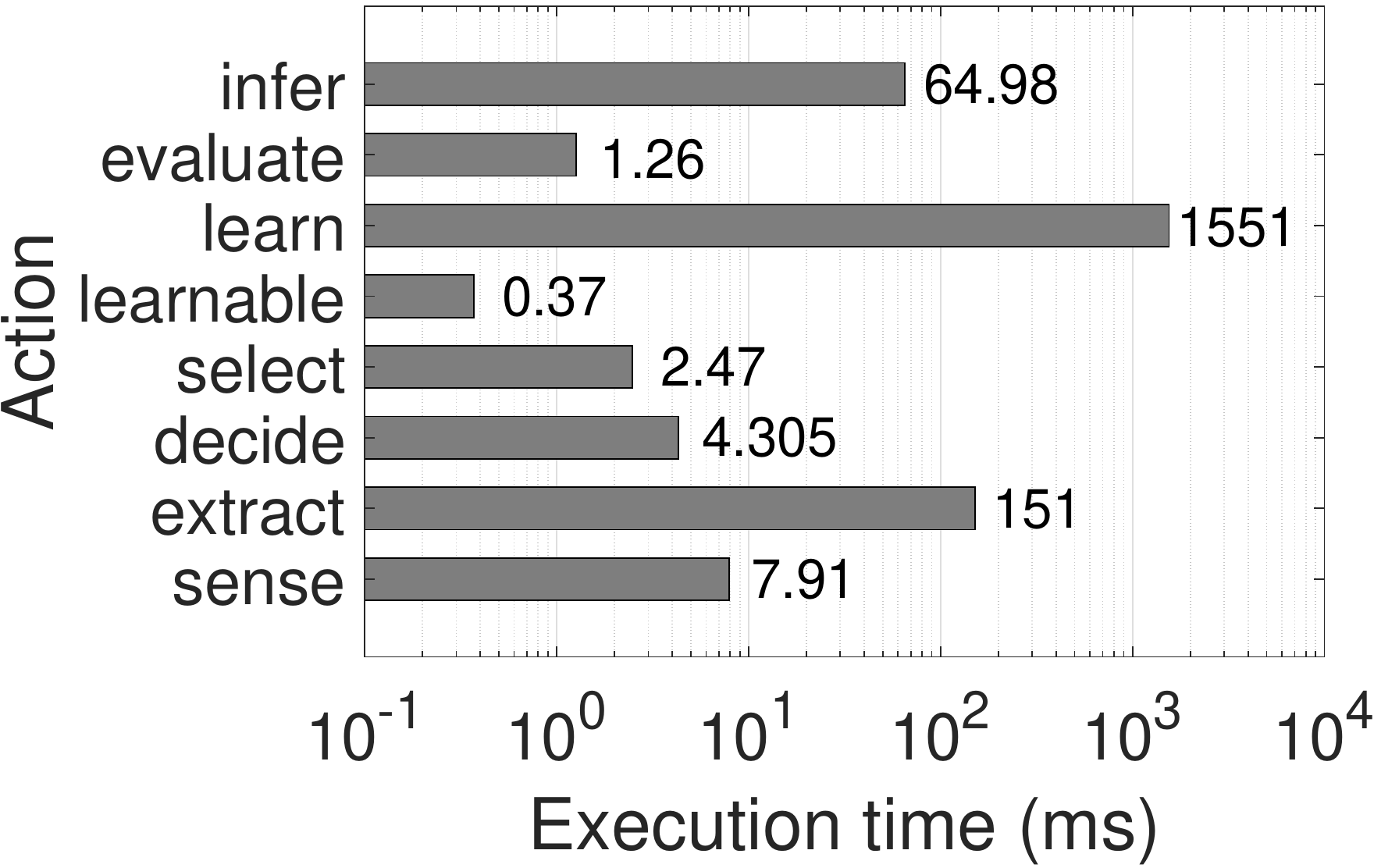}
    }
    \\
    \subfloat[Energy consumption ($k$-means)] 
    {
        \includegraphics[width=0.48\textwidth]{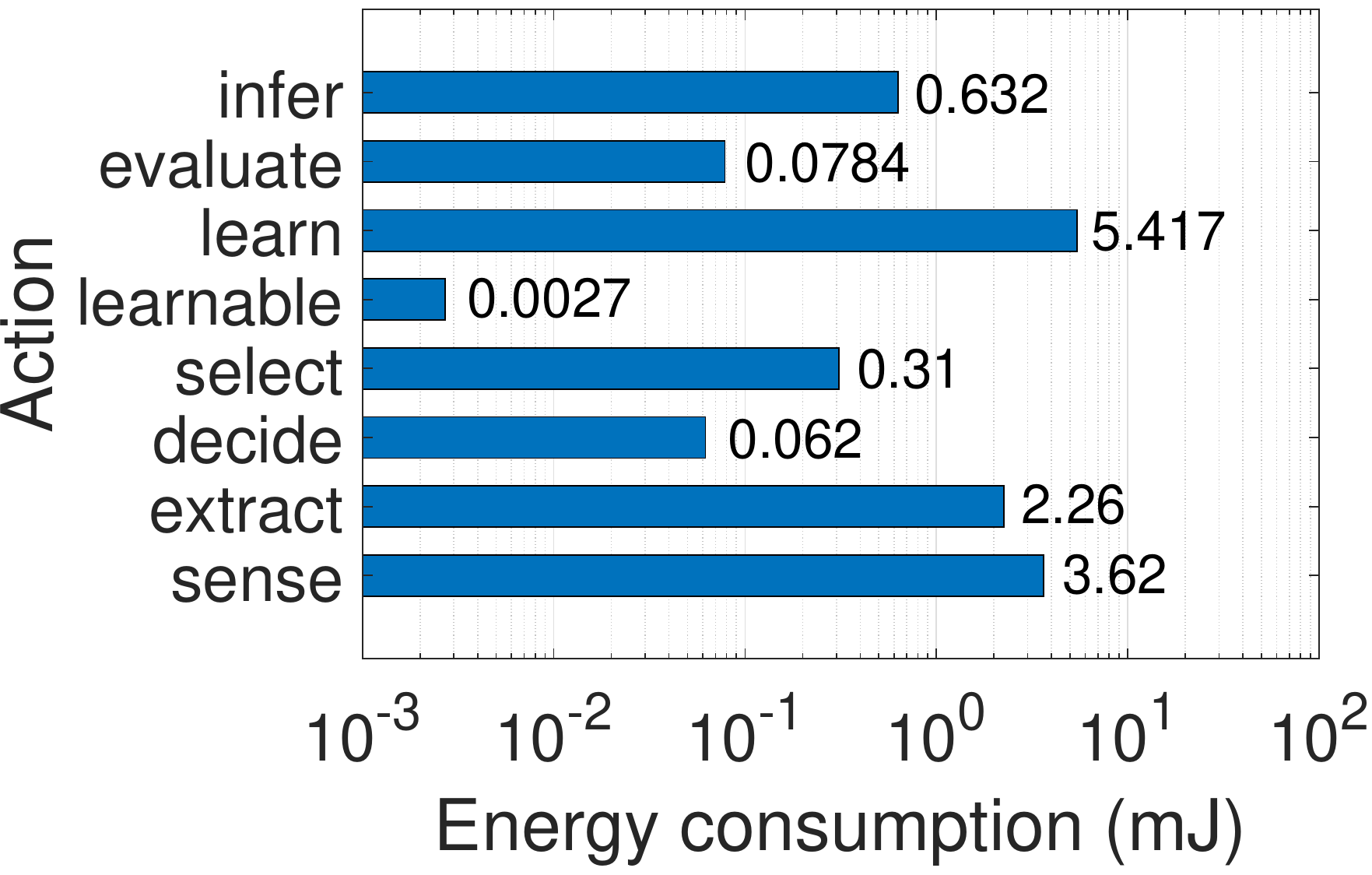}
    }
    \subfloat[Execution time ($k$-means)]
    {
        \includegraphics[width=0.48\textwidth]{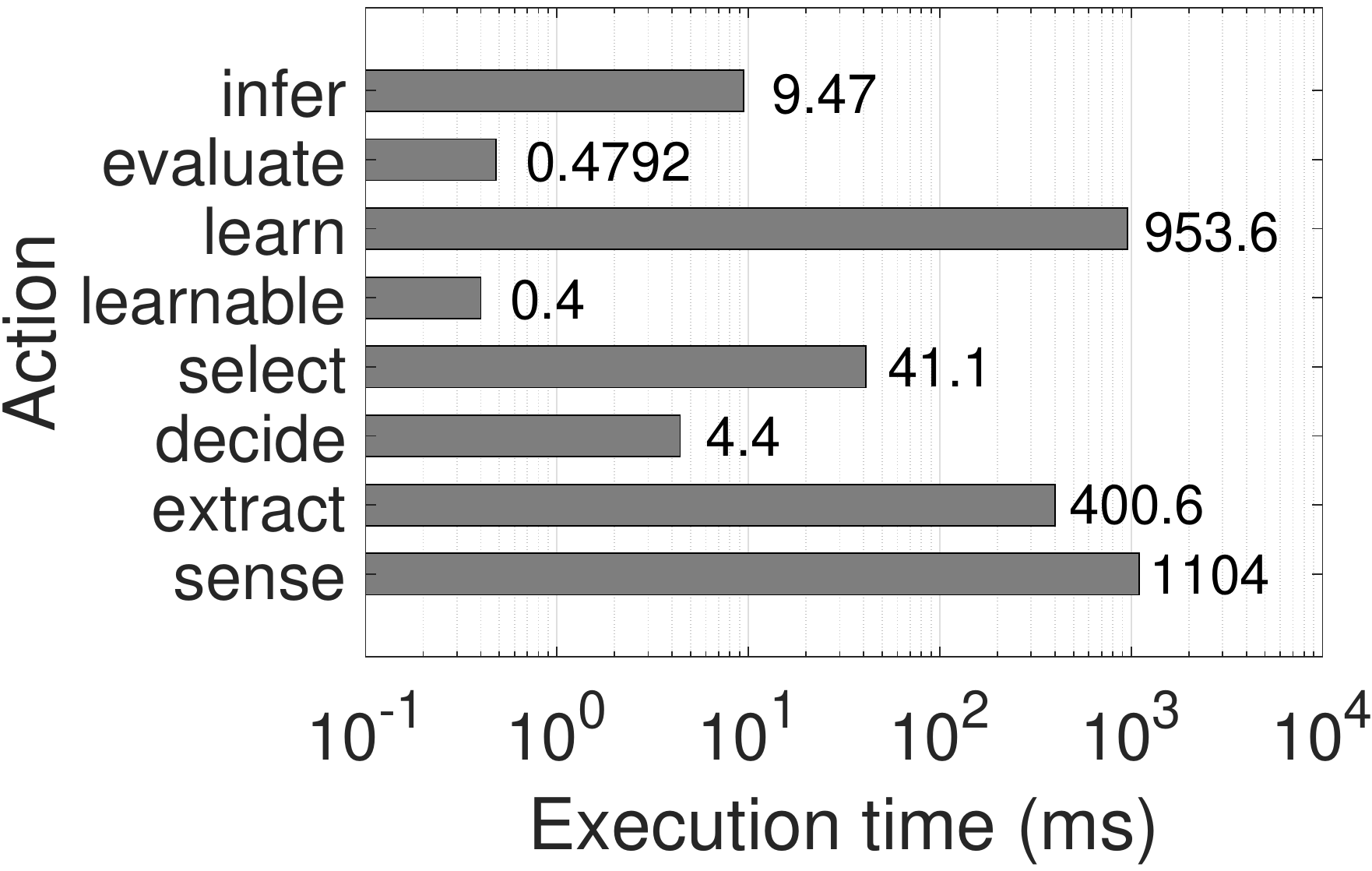}
    }
    \caption{Energy consumption and execution time of actions in two different learning algorithms. (a) and (b): $k$-nearest neighbors ($k$-NN). (c) and (d): neural network-based $k$-means ($k$-means). All plots are in the log-scale.}
    \label{fig:energy_time}
\end{minipage} \hspace{1.5mm}
\begin{minipage}{.32\textwidth}
    \centering
    \subfloat[Energy consumption] 
    {
        \includegraphics[width=0.98\textwidth]{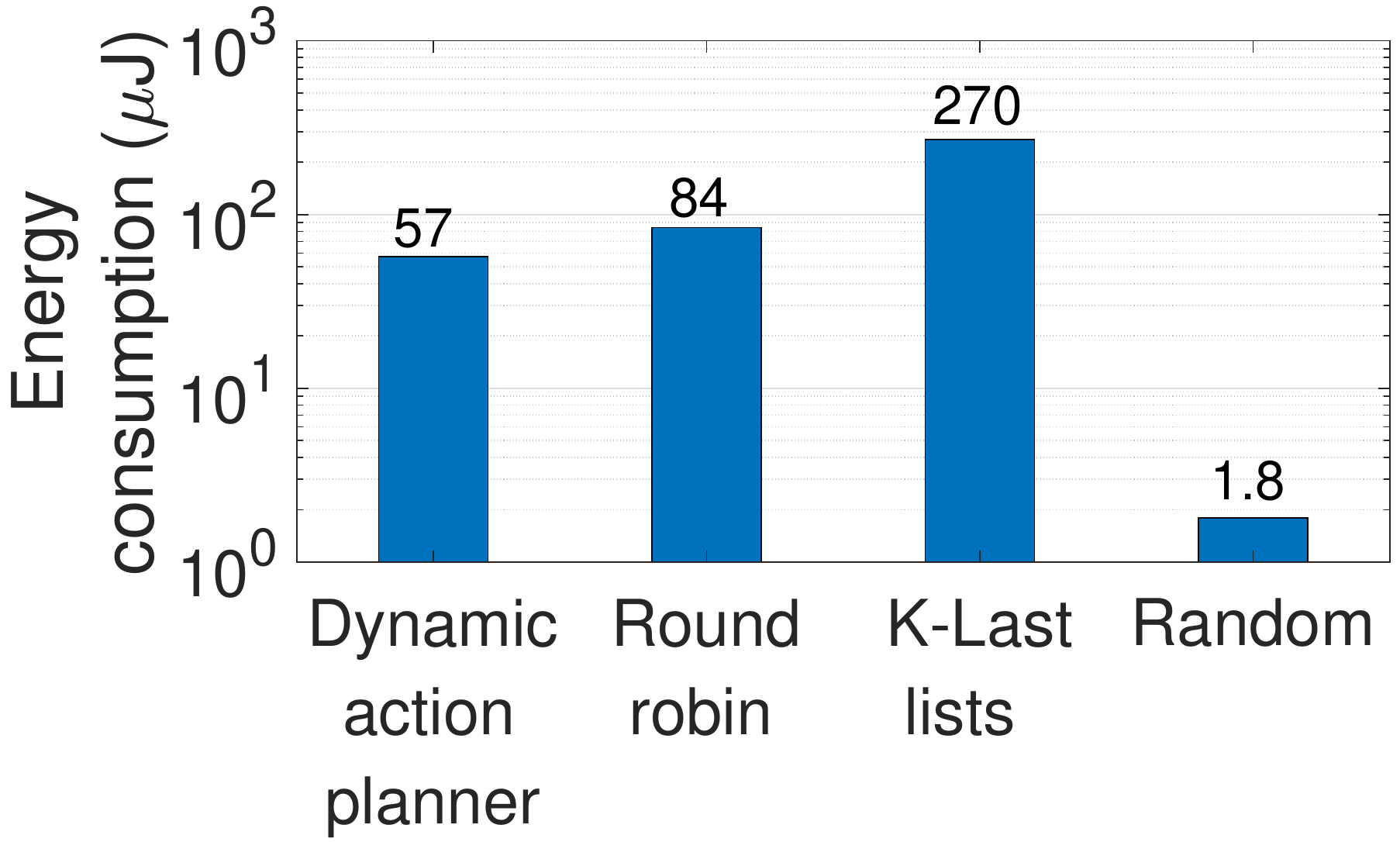}
    }
    \\
    \subfloat[Execution time] 
    {
        \includegraphics[width=0.98\textwidth]{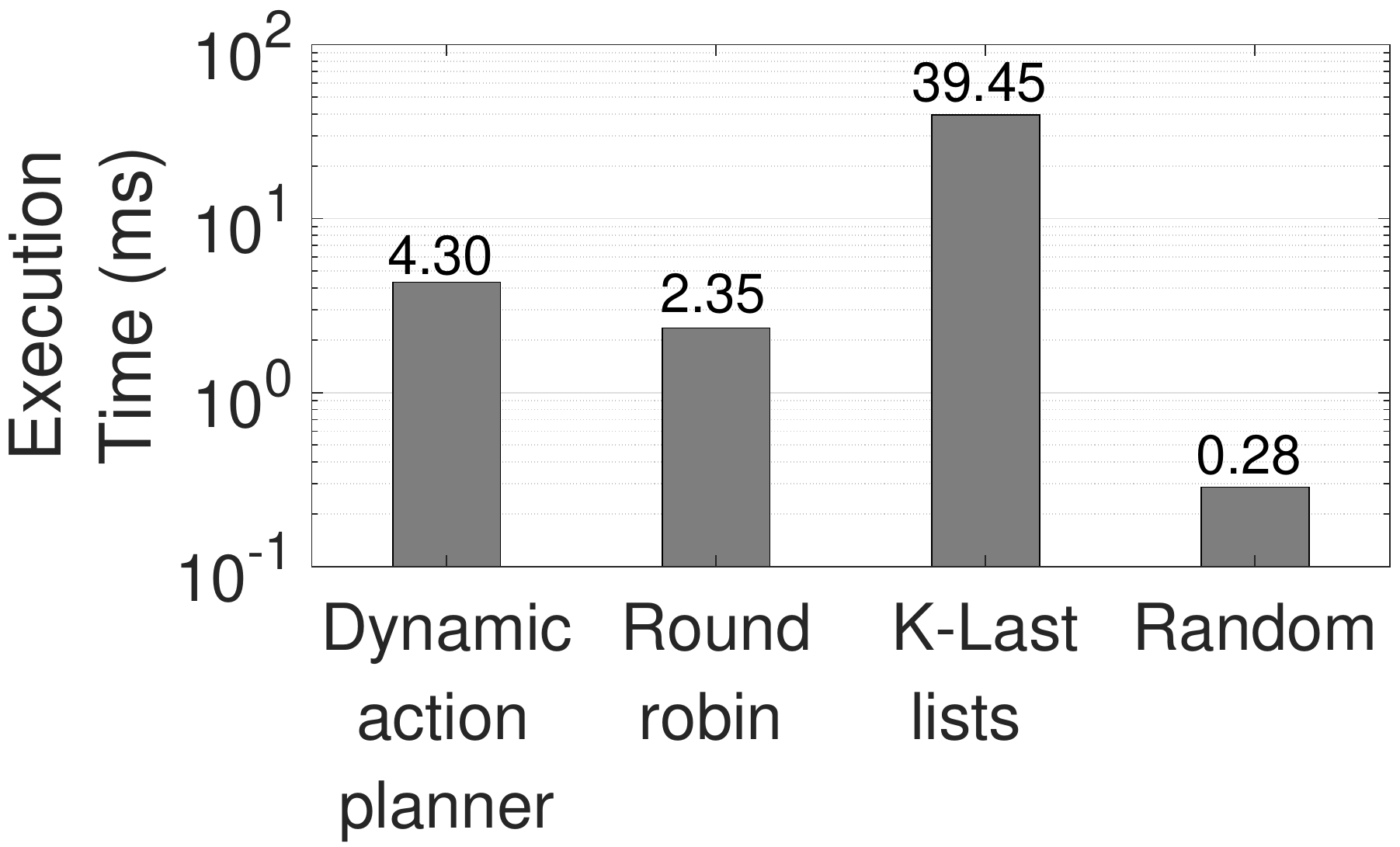}
    }
    \caption{Overhead (energy and execution time) of the dynamic action planner and three example selection algorithms. All plots are in the log-scale.}
    \label{fig:overhead}
\end{minipage}
\end{figure*}

Figures~\ref{fig:energy_time}(a) and \ref{fig:energy_time}(b) show the energy and time required by each action of the $k$-NN algorithm used in the air quality learning system. As expected, \textit{learn} consumes the highest 9.309mJ energy, which is decomposed into three sub-actions for intermittent execution. The energy consumption of \textit{sense} is relatively large (3.8mJ) since it acquires raw data from three sensors (UV, eCO2, and TVOC). Similar to energy consumption, \textit{learn} takes the longest time (1551ms) to execute, followed by \textit{extract} (151ms) and \textit{infer} (64.98ms).

Figures~\ref{fig:energy_time}(c) and \ref{fig:energy_time}(d) show the energy and time required by each action of the neural network-based $k$-means algorithm used in the vibration learning system~\cite{marsland2015machine}. The \textit{sense} and \textit{extract} actions consume the second (3.62mJ) and third (2.26mJ) largest energy after the \textit{learn} (5.417mJ) since they process acceleration sensor data at a high sampling rate. The \textit{learn} and \textit{infer} use the same neural network, but their overheads are different. The overhead of \textit{learn} (5.417mJ and 953.6ms) is about 100X higher than \textit{infer} (0.0632mJ and 9.47ms) since \textit{learn} involves several orders of magnitude more arithmetic operations and more iterations than \textit{infer}.

Figure~\ref{fig:overhead} shows the energy and time overhead of the dynamic action planner and the three example selection heuristics of the vibration learning system. We set the maximum number of admitted examples to two for the dynamic action planner, and the $k$-last lists uses three examples. As shown in Figures~\ref{fig:overhead}(a) and ~\ref{fig:overhead}(b), the dynamic action planner has an energy and time overhead of 57$\mu$J and 4.3ms, respectively. Although the action planner is executed more frequently than any of the actions (once after each action), its total overhead is below $3.5\%$ compared to the end-to-end processing of an example. In more detail, 2.9\% of energy and 1.4\% of time overhead is imposed for learning, while 4.2\% of energy and 4.3\% of time overhead is imposed for inference of an example, compared to the same system that does not run the dynamic action planner.

Figure~\ref{fig:overhead} also compares the example selection heuristics. Among the three heuristics, the $k$-last lists consumes the highest 270$\mu$J energy, whereas the randomized selection consumes the lowest 1.8$\mu$J. This is because the $k$-last lists computes the diversity and the representation scores for $2k$ examples ($\mathcal{O}(k^2)$) while the random heuristic only needs to generate a random number without looking into the acquired data.

%% file: tex/8.LIMITATION.tex
\section{Limitations and Future Work}

This paper is the first step towards intermittent machine learning on embedded devices, which enables a system to adapt its learning capability over a prolonged period of time. Despite the promising results, the proposed framework has several limitations that need further investigations in the future.

$\bullet$ \textit{Usability.} The type and scope of an intermittent learning application are limited by the nature of the energy harvesting sources and their relationship to the sensor data. For instance, energy harvested by the system may never be sufficient to execute a learning algorithm to completion that requires high-resolution sensor data and/or has a high energy demand in general. Furthermore, the occurrence of the events of interest (i.e., sensor events that the system wants to learn or classify) can be uncorrelated to the availability of harvested energy -- resulting in a large number of missed events. This limits the use of the system to applications that sense events that are frequent or always present.





$\bullet$ \textit{Programmability.}
Decomposing a source code into actions that run within the energy constraint is a challenging problem. Although we provide a pre-inspection tool that helps programmers develop such action modules, the tool may overestimate or underestimate the energy. Our approach is based on the assumption that machine learning modules typically follow a standard execution path. However, this iterative and statistical approach does not guarantee that the system experiences all the possible execution scenarios such as different execution paths, data-dependent branching, and events such as system failure, reconfiguration, and deterioration or malfunction of hardware.





$\bullet$ \textit{Learning Algorithm.}
Although the proposed framework supports supervised learning, it is limited by the fact that labels are not usually available to the learner at runtime. Interactions with an Oracle or another high-accuracy learner to obtain the labels, or adopting reinforcement learning principles may solve this problem, but such interactions with the environment increase both the time and energy costs of the system -- which may not be affordable in many intermittent systems. Limited amount of harvesting energy and the sluggishness of low-power microcontroller platforms also limit the execution of complex machine learning applications such as large-sized DNNs.

%% file: tex/9.RELATED_WORK.tex
\section{Related Work}
$\bullet$ \textit{Intermittent Computing Platform.} Several application-specific energy harvesting systems have been proposed that run on harvested RF~\cite{philipose2005battery, sample2008design, buettner2009recognizing, naderiparizi2015wispcam, zhang2011moo} or piezoelectric (kinetic) energy~\cite{huang2016battery, rodriguez2017intermittently}. In general, the goal of general-purpose intermittent computing platforms is to overcome the challenges due to the irregular and scarce power-supply. Mementos~\cite{ransford2012mementos} transforms general-purpose programs into interruptible computations that are protected from frequent power losses by automatic, energy-aware checkpointing. Ratchet~\cite{van2016intermittent} proposes a compiler-based technique that adds lightweight
checkpoints to unmodified programs that allow existing programs to execute across power cycles correctly. To eliminate the need for checkpoint placement heuristics, Hibernus~\cite{balsamo2015hibernus, balsamo2016hibernus++} puts the system to hibernation by monitoring the voltage and saving the system state when power is about to be lost. Chinchilla~\cite{maeng2018adaptive} runs unmodified C programs efficiently by overprovisioning the program with checkpoints to assure that the system makes progress, even with scarce energy. Approaches that are not based on the checkpointing technique have also been proposed. Chain~\cite{colin2016chain} utilizes a set of programmer-defined tasks that compute and exchange data through channels. Alpaca~\cite{maeng2017alpaca} preserves execution progress at the granularity of a task by privatizing the shared data between tasks that are detected using idempotence analysis. Clank~\cite{hicks2017clank} proposes a set of hardware buffers and memory access monitors that dynamically maintain idempotency. Several studies~\cite{hester2015tragedy, colin2018reconfigurable} focus on the power management of batteryless systems. 

Unlike the proposed intermittent learning system, none of these works considers how on-device machine learning can be performed effectively on harvested energy by considering the semantics of machine learning tasks. Several works propose sensing systems~\cite{yerva2012grafting, sudevalayam2011energy, seah2009wireless, kansal2003environmental} whose role is to sense data and forward them to other systems for further processing, but they do not perform on-device learning. Furthermore, these systems neither consider the utility of data nor provide any analysis of a system's expected task completion based on energy. Mayfly~\cite{hester2017timely} considers the timeliness of data, but neither takes into account the usefulness of data nor provides any energy analysis. Some energy prediction models such as ~\cite{kansal2007power} based on \emph{Exponentially Weighted Moving-Average} filter~\cite{cox1961prediction} or \emph{Weather-Conditioned Moving Average} algorithm~\cite{piorno2009prediction} require complex models designed for specific energy (solar) harvesters. Both use conventional time-domain energy analysis and prediction techniques, which is difficult to make in practice while the proposed framework performs energy event-based analysis.


$\bullet$ \textit{Embedded Machine Learning.}
Machine learning algorithms that run on low-performance processors have been studied. Bonsai~\cite{kumar2017resource} develops a tree-based algorithm for efficient inference on IoT devices having limited resources (e.g., 2KB RAM and 32KB read-only flash). ProtoNN~\cite{gupta2017protonn} proposes a compressed and accurate $k$-nearest neighbor algorithm for devices with limited storage. Deep neural networks have been implemented to run on embedded devices by reducing redundancy in their network model~\cite{denil2013predicting}. Neural network compression techniques such as quantization and encoding~\cite{han2015deep}, fixed-point number or binary representation~\cite{courbariaux2015binaryconnect}, HashedNets~\cite{chen2015compressing}, Sparse Neural Networks~\cite{bourely2017sparse}, multiplications using shift and addition~\cite{ding2017lightnn}, vector quantization~\cite{gong2014compressing},  circulant weight matrix~\cite{kotagiri2014memory}, and  structured transform~\cite{sindhwani2015structured} significantly reduce the size of neural network and run them on some high-performance embedded systems such as mobile devices.  

Several hardware architectures have been introduced to surmount the computational limitation of embedded machine learning.~\cite{lee2013low} proposes a custom processor integrating a CPU with configurable accelerators for discriminative machine-learning functions. Mixed-signal circuits such as~\cite{murmann2015mixed} explore a variety of design techniques that are leveraged in the design of embedded ConvNet ASICs. In the computer vision domain, a number of accelerators have been proposed for embedded systems, e.g., NeuFlow (a bio-inspired vision SoC)~\cite{pham2012neuflow}, ShiDianNao (Convolutional Neural Network within an SRAM)~\cite{du2015shidiannao}, and a scalable non-von Neumann architecture~\cite{merolla2014million}.


$\bullet$ \textit{Machine Learning on Harvested Energy.} Recently, an intermittent neural network \emph{inference} system~\cite{gobieski2018intelligence,gobieskiintermittent} has been proposed. But these works are quite different from the proposed framework and is limited in several ways. For instance, they only execute an inference task (i.e., no on-device training), the task pipeline is fixed at compile time (i.e., no dynamic task adjustment), and the evaluation reads pre-loaded in-memory processed data (i.e., no real sensing). Whereas the proposed intermittent learning systems consider all aspects of a machine learning task (including on-device training), portions of these learning tasks (i.e., actions) are dynamically scheduled at run-time by the dynamic action planner, and our evaluation has multiple end-to-end real systems. There exist batteryless systems that are designed for specific applications, such as eye-tracking~\cite{li2018battery} and gesture recognition~\cite{li2018self}, that use a simple threshold-based CFAR algorithm. CapBand~\cite{truong2018capband} combines two energy harvesters (solar and RF) to recognize hand gestures using a Convolutional Neural Network (CNN). Although these systems intermittently classify sensor data, their implementation is application-specific, and they neither consider the data and application-level semantics of machine learning algorithms nor implement on-device training and adaptation.

%% file: tex/A.CONCLUSION.tex
\section{Conclusion}
A new paradigm called the intermittent learning for embedded systems that are powered by harvested energy is introduced. To learn and build up intelligence from harvested energy, a learning task is divided into actions such as sensing, selecting, learning, or inferring, and they are dynamically executed based on an algorithm that chooses the best action to execute that maximizes learning performance under the energy constraints. The proposed system not only optimizes the sequence of actions but also makes a decision which examples should be learned while considering their potential to improve the learning performance as well as the energy level. A programming model and development tools have been proposed, and three applications (i.e., air-quality monitoring, human presence detecting, and vibration learning systems) have been implemented and evaluated. The evaluation results show that the learning tasks are efficiently intermittently executed based on the execution and data selection plans by the dynamic action planner and the example selection heuristics, respectively.

%% file: tex/bibliography.tex
\bibliographystyle{ACM-Reference-Format}
\bibliography{tex/bibliography}